\documentclass[10pt,twoside]{article}

\usepackage[utf8]{inputenc} 
\usepackage[T1]{fontenc}    
\usepackage{hyperref}       
\usepackage{url}            
\usepackage{booktabs}       
\usepackage{amsfonts}       
\usepackage{nicefrac}       
\usepackage{microtype}      
\usepackage{fullpage} 

\newenvironment{quotex}%
  {\list{}{\leftmargin=0.1in\rightmargin=0.1in}\item[]}%
  {\endlist}

\newenvironment{itemizex}%
{\begin{itemize}[leftmargin=0.25in]}%
{\end{itemize}}

\newenvironment{enumeratex}[1][]%
{\begin{enumerate}[#1, leftmargin=0.25in]}%
{\end{enumerate}}

\usepackage[dvipsnames]{xcolor}
\usepackage{xspace, amsmath, amssymb, mathtools, graphicx, bbm, subfig}
\usepackage[shortlabels]{enumitem}
\usepackage[export]{adjustbox}

\usepackage[linesnumbered, algoruled, boxed, lined]{algorithm2e}

\DontPrintSemicolon
\SetArgSty{textnormal}
\SetCommentSty{textnormal}

\newtheorem{theorem}{Theorem}
\newtheorem{lemma}{Lemma}
\newtheorem{definition}{Definition}

\DeclarePairedDelimiter{\abs}{\lvert}{\rvert}
\DeclarePairedDelimiter{\floor}{\lfloor}{\rfloor}

\DeclareMathOperator*{\argmax}{argmax}
\DeclareMathOperator{\median}{median}
\DeclareMathOperator{\mean}{mean}

\newcommand{\Expect}{\mathbb{E}}
\newcommand{\Prob}{\mathbb{P}}
\newcommand{\indicator}{\mathbbm{1}}
\newcommand{\indicatorevent}[1]{\indicator{\{#1\}}}

\newcommand{\union}{\cup}

\newcommand{\defn}{:=}

\newcommand{\grade}{evaluate\xspace} 
\newcommand{\grades}{{\grade}s\xspace} 
\newcommand{\grading}{evaluating\xspace}
\newcommand{\graded}{evaluated\xspace}
\newcommand{\grader}{reviewer\xspace}
\newcommand{\graders}{{\grader}s\xspace}
\newcommand{\Graders}{Reviewers\xspace}

\newcommand{\score}{score\xspace} 
\newcommand{\scores}{{\score}s\xspace} 

\newcommand{\numitems}{n}
\newcommand{\func}{f}
\newcommand{\idxitem}{i}
\newcommand{\val}{x}
\newcommand{\numgraders}{m}
\newcommand{\numgradershalf}{\numgraders/2}
\newcommand{\numgradershalffrac}{\frac{\numgraders}{2}}
\newcommand{\numgradersquarter}{\numgraders/4}

\newcommand{\idxgrader}{j}

\newcommand{\tmpx}{x}
\newcommand{\tmpA}{A} 

\newcommand{\itemsofgrader}{S}
\newcommand{\itemsofgraderunorder}{\widetilde{\itemsofgrader}}

\newcommand{\quantity}{\pi}
\newcommand{\quantitygt}{\quantity^*}
\newcommand{\quantityest}{\widehat{\quantity}}

\newcommand{\loss}{L}
\newcommand{\assignment}{A}

\newcommand{\info}{y}
\newcommand{\infoord}{b}
\newcommand{\greateritem}{\succ}

\newcommand{\textours}{\text{our}}
\newcommand{\capours}{\widetilde}
\newcommand{\capord}{\widehat}

\newcommand{\textcanonical}{\text{can}}
\newcommand{\quantityordcanonical}{\capord{\quantity}_{\textcanonical}}
\newcommand{\quantitycardourscanonical}{\capours{\quantity}_{\textcanonical}^{\textours}}
\newcommand{\funcmonotone}{w}
\newcommand{\topitem}{\widehat{\idxitem}^{(1)}}
\newcommand{\bottomitem}{\widehat{\idxitem}^{(2)}}

\newcommand{\threshenvelope}{Z}
\newcommand{\threshenvelopeobs}{z}
\newcommand{\pdfenvelope}{p}
\newcommand{\funcmonotoneredefine}{\widetilde{\funcmonotone}}

\newcommand{\textabtest}{\text{ab}}
\newcommand{\quantityordabtest}{\capord{\quantity}_{\textabtest}}

\newcommand{\quantitycardoursabtest}{\capours{\quantity}^{\textours}_{\textabtest}}
\newcommand{\estfrompair}{r}

\newcommand{\textsort}{\text{rank}}
\newcommand{\infoordall}{\mathcal{B}}
\newcommand{\infocardall}{\mathcal{Y}}
\newcommand{\textunif}{\text{unif}}

\newcommand{\graph}{\mathcal{G}}

\newcommand{\quantityordsort}{\capord{\quantity}_{\textsort}}
\newcommand{\quantityordsortunif}{\capord{\quantity}_{\textsort\text{-}\textunif}}
\newcommand{\quantitycardourssort}{{\capours{\quantity}}_{\textsort}^{\textours}}

\newcommand{\quantitycardourssortunif}{\capours{\quantity}_{\textsort\text{-}\textunif}^{\textours}}

\newcommand{\quantityestflip}{\quantityest_{\text{flip}}}
\newcommand{\position}{t}

\newcommand{\bias}{b}
\newcommand{\scale}{k}
\newcommand{\noise}{\epsilon}
\newcommand{\improve}[2]{\rho_{#2}(#1)}

\newcommand{\reals}{\mathbb{R}}
\newcommand{\integers}{\mathbb{Z}}

\newcommand{\realsnonneg}{[0,\infty)}
\newcommand{\domX}{\reals}
\newcommand{\domY}{\reals}
\newcommand{\permall}{\Pi}
\newcommand{\permfunc}[1]{\func^{#1}}
\newcommand{\permval}[1]{\val^{#1}}

\newcommand{\symcompare}[2]{\mathrel{\mathop\gtrless\limits^{#1}_{#2}}}

\newcommand{\funcmonotoneparam}{\gamma}

\newcommand{\quantityestdet}{\quantityest_\text{det}}
\newcommand{\quantityestsign}{\quantityest_\text{sign}}
\newcommand{\setassignments}{\mathcal{A}}

\newcommand{\probpairsuccess}{\lambda}
\newcommand{\probpairsuccessmin}{\probpairsuccess_{\text{min}}}

\newcommand{\sumofpairs}{V}
\newcommand{\resultofpair}{V}
\newcommand{\numgradershalfk}{k}

\DeclareMathOperator{\topo}{topo}

\newcommand{\lessitem}{\prec}
\newcommand{\quantityinv}{\sigma}
\newcommand{\quantityinvgt}{\quantityinv^*}
\newcommand{\quantityinvest}{\widehat{\quantityinv}}
\newcommand{\quantityinvestre}{\widehat{\quantityinv}_{\text{re}}}

\newcommand{\quantityinvinit}{\widehat{\quantityinv}_\text{init}}

\newcommand{\event}{E}
\newcommand{\given}{\mid}

\newcommand{\stepone}{\text{(i)}\xspace}
\newcommand{\steptwo}{\text{(ii)}\xspace}
\newcommand{\stepthree}{\text{(iii)}\xspace}

\newcommand{\assignmentobs}{a}

\newcommand{\textmedian}{\text{med}}
\newcommand{\gradersofitem}{R}
\newcommand{\varbinom}{B}
\newcommand{\idxpair}{k}
\newcommand{\compareofpair}{v}

\newcommand{\itemsofgraderunorderall}{\mathcal{Q}}
\newcommand{\itemsofgraderunorderallobs}{q}
\newcommand{\itemsofgraderunorderallavail}{\itemsofgraderunorderall_\text{avail}}

\newcommand{\numorders}{T}
\newcommand{\numflips}{L}

\newcommand{\setposterior}{\Omega}

\newcommand{\varpositions}{\texttt{flippable\_positions}}
\newcommand{\vargraders}{\texttt{{\grader}\_indices}}
\newcommand{\infoordallobs}{\beta}
\newcommand{\numflipsobs}{\ell}
\newcommand{\quantityordsortopt}{\capord{\quantity}_{\textsort}^\text{opt}}

\newcommand{\quantitycardourssortequiv}{{\capours{\quantity}}_{\textsort}^{\text{eq}}}

\title{\bf \Large Your 2 is My 1, Your 3 is My 9: \\
Handling Arbitrary Miscalibrations in Ratings}
\author{\\
  Jingyan Wang and Nihar B. Shah\\~\\
  School of Computer Science \\ 
  Carnegie Mellon University\\
  \texttt{\{jingyanw,nihars\}@cs.cmu.edu} 
}

\date{}

\begin{document}

\maketitle

\begin{abstract}
    Cardinal scores (numeric ratings) collected from people are well known to suffer from miscalibrations. A popular approach to address this issue is to assume simplistic models of miscalibration (such as linear biases) to de-bias the scores. This approach, however, often fares poorly because people's miscalibrations are typically far more complex and not well understood. In the absence of simplifying assumptions on the miscalibration, it is widely believed by the crowdsourcing community that the only useful information in the cardinal scores is the induced ranking.  In this paper, inspired by the framework of Stein's shrinkage, empirical Bayes, and the classic two-envelope problem, we contest this widespread belief. Specifically, we consider cardinal scores with arbitrary (or even adversarially chosen) miscalibrations which are only required to be consistent with the induced  ranking. We design estimators which despite making no assumptions on the miscalibration, strictly and uniformly outperform all possible estimators that rely on only the ranking. Our estimators are flexible in that they can be used as a plug-in for a variety of applications, and we provide a proof-of-concept for A/B testing and ranking. Our results thus provide novel insights in the eternal debate between cardinal and ordinal data. 
\end{abstract}


\section{Introduction}  

\begin{quotex}
\emph{``A raw rating of 7 out of 10 in the
absence of any other information is potentially useless.''}~\cite{mitliagkas2011}

\emph{``The rating scale as well as the individual ratings are often arbitrary and may not be consistent from one user to another.''}~\cite{ammar2012aggregation}
\end{quotex}

Consider two items that need to be evaluated (for example, papers submitted to a conference) and two \graders. Suppose each \grader is assigned one distinct item for evaluation, and this assignment is done uniformly at random. The two \graders provide their evaluations (say, in the range $[0,1]$) for the respective item they \grade, from which the better item must be chosen. However, the \graders' rating scales may be miscalibrated. It might be the case that the first \grader is lenient and always provides scores in $[0.6,1]$ whereas the second \grader is more stringent and provides scores in the range $[0,0.4]$. Or it might be the case that one \grader is moderate whereas the other is extreme -- the first \grader's 0.2 is equivalent to the second \grader's 0.1 whereas the first \grader's 0.3 is equivalent to the second \grader's 0.9. More generally, the miscalibration of the reviewers may be arbitrary and unknown. Then is there any hope of identifying the better of the two items with any non-trivial degree of certainty?

A variety of applications involve collection of human preferences or judgments in terms of cardinal scores (numeric ratings). A perennial problem with eliciting cardinal scores is that of miscalibration -- the systematic errors introduced due to incomparability of cardinal scores provided by different people (see~\cite{griffin2008calibration} and references therein).

This issue of miscalibration is sometimes addressed by making simplifying assumptions about the form of miscalibration, and post-hoc corrections under these assumptions. Such models include  one-parameter-per-reviewer additive biases~\cite{paul1981calibration,baba2013quality,ge13bias,mackay2017calibration}, two-parameters-per-reviewer scale-and-shift biases~\cite{paul1981calibration,roos2011calibrate} and others~\cite{flach2010kdd}. The calibration issues with human-provided scores are often significantly more complex causing significant violations to these simplified assumptions (see~\cite{griffin2008calibration} and references therein).  Moreover, the algorithms for post-hoc correction often try to estimate the individual parameters which may not be feasible due to low sample sizes. For instance, John Langford notes from his experience as the program chair of the ICML 2012 conference~\cite{langford2012icml}:\begin{quotex}
\emph{``We experimented with reviewer normalization and generally found it significantly harmful.''}
\end{quotex}
\noindent This problem of low sample size is exacerbated in a number of applications such as A/B testing where every \grader\ \grades only one item, thereby making the problem underdetermined even under highly restrictive models.

It is commonly believed that when unable or unwilling to make any simplifying assumptions on the bias in cardinal scores, the only useful information is the ranking of the scores~\cite{rokeach1968values,freund2003boosting,harzing2009rating,mitliagkas2011,ammar2012aggregation,negahban2012ranking}. This perception gives rise to a second approach towards handling miscalibrations -- that of using only the induced ranking or otherwise directly eliciting a ranking and not scores from the use. As noted by Freund et al.~\cite{freund2003boosting}:
\begin{quotex}
\emph{``[Using rankings instead of ratings] becomes very important when we combine the rankings of many viewers who often use completely
different ranges of scores to express identical preferences.''}
\end{quotex}
These motivations have spurred a long line of literature on analyzing data that takes the form of partial or total rankings of items~\cite{cook2007proposal,baskin2009recommender,ammar2012aggregation,negahban2012ranking,rajkumar2015ranking,shah2016estimation,shah18simple}.

In this paper, we contest this widely held belief with the following two fundamental questions: 
\begin{itemizex} 
\item In the absence of simplifying modeling assumptions on the miscalibration, is there any estimator (based on the scores) that can outperform estimators based on the induced rankings? 
\item If only one evaluation per \grader is available, and if each \grader may have an arbitrary (possibly adversarially chosen) miscalibration, is there hope of estimation better than random guessing?
\end{itemizex}
We show that the answer to both questions is ``Yes''. One need not make simplifying assumptions about the miscalibration and yet guarantee a performance superior to that of any estimator that uses only the induced rankings. 

In more detail, we consider settings where a number of people provide cardinal scores for one or more from a collection of items. The calibration of each \grader is represented by an unknown monotonic function that maps the space of true values to the scores given by this \grader. These functions are arbitrary and may even be chosen adversarially. We present a class of estimators based on cardinal scores given by the \graders which \emph{uniformly} outperforms any estimator that uses only the induced rankings. A compelling feature of our estimators is that they can be used as a plug-in to improve ranking-based algorithms in a variety of applications, and we provide a proof-of-concept for two applications: A/B testing and ranking. 

The techniques used in our analyses draw inspiration from the framework of Stein's shrinkage~\cite{stein1956inadmissibility,james1961estimation} and empirical Bayes~\cite{robbins1956empirical}. Moreover, our setting with $2$ reviewers and $2$ papers presented subsequently in the paper carries a close connection to the classic two-envelope problem (for a survey of the two-envelope problem, see~\cite{gnedin2016survey}), and our estimator in this setting is similar in spirit to the randomized strategy~\cite{cover1987envelope} proposed by Thomas Cover. We discuss connections with the literature in more detail in  Section~\ref{sec:two_envelope}.

Our work provides a new perspective on the eternal debate between cardinal scores and ordinal rankings. It is often believed that ordinal rankings are a panacea for the miscalibration issues with cardinal scores. Here we show that ordinal estimators are not only inadmissible, they are also strictly and uniformly beaten by our cardinal estimators. Our results thus uncover a new point on the bias-variance tradeoff for this class of problems: Estimators that rely on simplified assumptions about the miscalibration incur biases due to model mismatch, whereas the absence of such assumptions in our work eliminates the modeling bias. Moreover, in this minimal-bias regime, our cardinal estimators incur a strictly smaller variance as compared to estimators based on ordinal data alone. 

Finally, a note qualifying the scope of the problem setting considered here. In applications such as crowdsourced microtasks where workers often spend very little time answering every question, the cardinal scores elicited may not necessarily be consistent with the ordinal rankings, and moreover, ordinal rankings are often easier and faster to provide. These differences cease to exist in a variety of applications such as peer-review or in-person laboratory A/B tests which require the \graders to spend a non-trivial amount of time and effort in the review process, and these applications form the motivation of this work.


\section{Preliminaries}\label{sec:prelim}

Consider a set of $\numitems$ items denoted as $\{1,\ldots,\numitems\}$ or $[\numitems]$ in short.\footnote{We use the standard notation of $[\kappa]$ to denote the set $\{1,\ldots,\kappa\}$ for any positive integer $\kappa$.} Each item $\idxitem \in [\numitems]$ has an unknown value $\val_\idxitem \in \domX$. For ease of exposition, we assume that all items have distinct values.
There are $\numgraders$ \graders $\{1,\ldots,\numgraders\}$ and each \grader\xspace\grades a subset of the items. The calibration of any \grader $\idxgrader \in [\numgraders]$ is given by  an unknown, strictly-increasing function $\func_\idxgrader : \domX \rightarrow \domY$. (More generally, our results hold for any non-singleton intervals on the real line as the domain and range of the calibration functions). When \grader $\idxgrader$ evaluates item $\idxitem$, the reported score is $\func_\idxgrader(\val_\idxitem)$. We make no other assumptions on the calibration functions $\func_1,\ldots,\func_\numgraders$. We use the notation $\greateritem$ to represent a relative order of any items, for instance, we use ``$1 \greateritem 2$'' to say that item $1$ has a larger value (ranked higher) than item $2$. We assume that $\numgraders$ and $\numitems$ are finite.

Every \grader is assigned one or more items to \grade. We denote the assignment of items to \graders as $\assignment = (\itemsofgrader_1, \ldots, \itemsofgrader_\numgraders)$, where $\itemsofgrader_\idxgrader \subseteq [\numitems]$ is the set of items assigned to \grader $\idxgrader \in [\numgraders]$. We use the notation $\permall$ to represent the set of all permutations of $\numitems$ items. We let  $\quantitygt \in \permall$ denote the ranking of the $\numitems$ items induced by their respective values $(\val_1,\ldots,\val_\numitems)$, such that $\val_{\quantitygt(1)} > \val_{\quantitygt(2)} > \cdots > \val_{\quantitygt(\numitems)}$. The goal is to estimate this ranking $\quantitygt$ from the evaluations of the \graders. We consider two types of settings: an ordinal setting where estimation is performed using the rankings induced by each \grader's reported \scores, and a cardinal setting where the estimation is performed using the \graders' \scores (which can have an arbitrary miscalibration and only need to be consistent with the rankings). Formally:
\begin{itemizex}
    \item \textbf{Ordinal:} Each \grader $\idxgrader$ reports a total ranking among the items in $\itemsofgrader_\idxgrader$, that is, the ranking of the items induced by the values $\{\func_{\idxgrader}(\val_\idxitem)\}_{\idxitem \in \itemsofgrader_\idxgrader}$. An ordinal estimator observes the assignment $\assignment$ and the rankings reported by all \graders.
    \item \textbf{Cardinal:} Each \grader $\idxgrader$ reports the \scores for the items in $\itemsofgrader_\idxgrader$, that is, the values of  $\{\func_{\idxgrader}(\val_\idxitem)\}_{\idxitem \in \itemsofgrader_\idxgrader}$. A cardinal estimator observes the assignment $\assignment$ and the \scores reported by all \graders.
\end{itemizex}

Observe that the setting described above considers ``noiseless'' data, where each \grader reports either the \scores $\{\func_\idxgrader(\val_\idxitem)\}$ or the induced rankings. We provide an extension to the noisy setting in Appendix~\ref{app:noisy}.

In order to compare the performance of different estimators, we use the notion of \emph{strict uniform dominance}. Informally, we say that one estimator strictly uniformly dominates another if it incurs a strictly lower risk for all possible choices of the miscalibration functions and the item values. 

In more detail, suppose that you wish to show that an estimator $\quantityest_1$ is superior to estimator $\quantityest_2$ with respect to some metric for estimating $\quantitygt$. However, there is a clever adversary who intends to thwart your attempts. The adversary can choose the miscalibration functions of all \graders and the values of all items, and moreover, can tailor these choices for different realizations of $\quantitygt$. Formally, the adversary specifies a set of values $\{\permfunc{\quantity}_1,\ldots,\permfunc{\quantity}_\numgraders, \permval{\quantity}_{1},\ldots,\permval{\quantity}_{\numitems}\}_{\quantity \in \permall}$. The only constraints in this choice are that the miscalibration functions $\permfunc{\quantity}_1,\ldots,\permfunc{\quantity}_\numgraders$ must be strictly monotonic and that the item values $\permval{\quantity}_{1},\ldots,\permval{\quantity}_{\numitems}$ should induce the ranking $\quantity$. 
In the sequel, we consider two ways of choosing the true ranking $\quantitygt$: In one setting, $\quantitygt$ can be chosen by the adversary, and in the second setting $\quantitygt$ is drawn uniformly at random from $\permall$. Once this ranking $\quantitygt$ is chosen, the actual values of the miscalibration functions and the item values are set as $\permfunc{\quantitygt}_1,\ldots,\permfunc{\quantitygt}_\numgraders$ and $\permval{\quantitygt}_{1},\ldots,\permval{\quantitygt}_{\numitems}$. The items are then assigned to \graders according to the (possibly random) assignment $\assignment$. 
The \graders now provide their ordinal or cardinal evaluations as described earlier, and these evaluations are used to compute and evaluate the two estimators $\quantityest_1$ and $\quantityest_2$. We say that estimator $\quantityest_1$ strictly uniformly dominates $\quantityest_2$, if $\quantityest_1$ is always guaranteed to incur a strictly smaller (expected) error than $\quantityest_2$. Formally:
\begin{definition}[Strict uniform dominance]\label{def:uniformly_better}
Let $\quantityest_1$ and $\quantityest_2$ be two estimators for the true ranking $\quantitygt$. Estimator $\quantityest_1$ is said to strictly uniformly dominate estimator $\quantityest_2$ with respect to a given loss $\loss: \permall\times \permall \rightarrow \reals$ if
\begin{align}
    \Expect[\loss(\quantitygt, \quantityest_1)] < \Expect[\loss(\quantitygt, \quantityest_2)] \qquad  \text{ for all permissible~~} \{\permfunc{\quantity}_1,\ldots,\permfunc{\quantity}_\numgraders, \permval{\quantity}_{1},\ldots,\permval{\quantity}_{\numitems}\}_{\quantity \in \permall}.
\label{eq:uniformly_better}
\end{align}
The expectation is taken over any randomness in the assignment $\assignment$ and the estimators. If the true ranking $\quantitygt$ is drawn at random from a fixed distribution, then the expectation is also taken over this distribution; otherwise, inequality~\eqref{eq:uniformly_better} must hold for all values of $\quantitygt$.
\end{definition}
Note that strict uniform dominance is a stronger notion than comparing estimators in terms of their minimax (worst-case) or average-case risks. Moreover, if an estimator $\quantityest_2$ is strictly uniformly dominated by some estimator $\quantityest_1$, then the estimator $\quantityest_2$ is inadmissible.

Finally, for ease of exposition, we focus on the 0-1 loss in the main text: \begin{align*} 
\loss(\quantity^*, \quantity) = \indicatorevent{\quantity^* \ne \quantity},
\end{align*}
where we use the standard notation $\indicatorevent{\tmpA}$ to denote the indicator function of an event $\tmpA$, where $\indicatorevent{\tmpA} = 1$ if the event $\tmpA$ is true, and $0$ otherwise. Extensions to other metrics of Kendall-tau distance and Spearman's footrule distance are provided in Appendix~\ref{app:other_metrics}.


\section{Main results}

In this section we present our main theoretical results. All proofs are provided in Section~\ref{app:proofs}.


\subsection{A canonical setting}\label{sec:canonical}

We begin with a canonical setting that involves two items and two \graders (that is, $\numitems=2$, $\numgraders = 2$), where each \grader\xspace\grades one of the two items. Our analysis for this setting conveys the key ideas underlying our general results. These ideas are directly applicable towards designing uniformly superior estimators for a variety of applications, and we subsequently demonstrate this general utility with two applications.

In this canonical setting, each of the two \graders\xspace\grades one of the two items chosen uniformly at random without replacement, that is, the assignment $\assignment$ is chosen  uniformly at random from the two possibilities  $(\itemsofgrader_1 = 1, \itemsofgrader_2 = 2)$ and $(\itemsofgrader_1 = 2, \itemsofgrader_2 = 1)$. Since each \grader is assigned only one item, the ordinal data is vacuous. Then the natural ordinal baseline is an estimator which makes a guess uniformly at random:
\begin{align*}
    \quantityordcanonical(\assignment, \{\}) = \begin{cases}
        1 \greateritem 2 & \text{with probability } 0.5\\
        2 \greateritem 1 & \text{with probability } 0.5.
    \end{cases}
\end{align*}

In the cardinal setting, let  $\info_1 $ denote the \score reported for item $1$ by its respective \grader, and let $\info_2$ denote the \score for item $2$ reported by its respective \grader. Since the calibration functions are arbitrary (and may be adversarial), it appears hopeless to obtain information about the relative values of $\val_1$ and $\val_2$ from just this data. Indeed, as we show below, standard estimators such as the sign test --- ranking the items in terms of their \grader-provided scores --- provably fail to achieve this goal. More generally, the following theorem holds for the class of all deterministic estimators, that is, estimators given by deterministic mappings from $\{\assignment,\info_1,\info_2\}$ to the set $\{1 \greateritem 2, 2 \greateritem 1\}$.
\begin{theorem}
\label{thm:canonical_det_fails}
No deterministic (cardinal or ordinal) estimator can strictly uniformly dominate the random-guessing estimator $\quantityordcanonical$.
\end{theorem}

This theorem demonstrates the difficulty of this problem by ruling out all deterministic estimators. Our original question then still remains: is there any estimator that can strictly uniformly outperform the random-guessing ordinal baseline?

We show that the answer is yes, with the construction of a randomized estimator for this canonical setting, denoted as $\quantitycardourscanonical$. This estimator is based on a function $\funcmonotone:\realsnonneg \rightarrow [0,1)$ which may be chosen as any  arbitrary strictly-increasing  function. For instance, one could choose  $\funcmonotone(\tmpx) = \frac{\tmpx}{1+\tmpx}$ or $\funcmonotone$ as the sigmoid function. 
Given the scores $\info_1,\info_2$ reported for the two items, let $\topitem \in \argmax_{\idxitem \in \{1,2\}} \info_\idxitem$ denote the item which receives the higher score, and let $\bottomitem$ denote the remaining item (with ties broken uniformly). 
Then our randomized estimator outputs:
\begin{align}\label{eq:canonical_estimator}
     \quantitycardourscanonical(\assignment,\info_1,\info_2) = & \begin{cases}\topitem \greateritem \bottomitem & \text{with probability } \frac{1 + \funcmonotone(\abs{\info_1-\info_2})}{2}\\
      \bottomitem \greateritem \topitem  & \text{otherwise}.
    \end{cases}
\end{align}
Note that the the output of this estimator is independent of the assignment $\assignment$, so in the remainder of this paper we also denote this estimator as $\quantitycardourscanonical(\info_1, \info_2)$.

The following theorem now proves that our proposed estimator indeed achieves the stated goal.
\begin{theorem}\label{thm:canonical_ours}
The randomized estimator $\quantitycardourscanonical$ strictly uniformly dominates the random-guessing baseline $\quantityordcanonical$.
\end{theorem}

While this result considers a setting with ``noiseless'' observations (that is, where $\info = \func(\val)$), in Appendix~\ref{app:noisy} we show that the guarantee for $\quantitycardourscanonical$ continues to hold when the observations are noisy.

Having established the positive result for this canonical setting, we now discuss some connections and inspirations in the literature.

\subsubsection{Connections to the literature}\label{sec:two_envelope}

The canonical setting has a close connection to the randomized version of the two-envelope problem~\cite{cover1987envelope}. In the two-envelope problem, there are two arbitrary numbers. One of the two numbers is observed uniformly at random, and the other remains unknown. The goal is to estimate which number is larger. This problem can also be viewed from a game-theoretic perspective~\cite{gnedin2016survey} as ours, where one player picks an estimator and the other player picks the two values. Cover~\cite{cover1987envelope} proposed a randomized estimator whose probability of success is strictly larger than $0.5$ uniformly across all arbitrary pairs of numbers. The proposed estimator samples a new random variable $\threshenvelope$ whose distribution has a probability density function $\pdfenvelope$ with $\pdfenvelope(\threshenvelopeobs) > 0$ for all $\threshenvelopeobs \in \reals$. Then if the observed number is smaller than $\threshenvelope$, the estimator decides that the observed number is the smaller number; if the observed number is larger than $\threshenvelope$, the estimator decides that the observed number is the larger number.

Our canonical setting can be reduced to the two-envelope problem as follows. Consider the two values $\func_1(\val_1) - \func_2(\val_2)$ and $\func_1(\val_2) - \func_2(\val_1)$. Since the two items are assigned to the two \graders uniformly at random, we observe one of these two values uniformly at random. By the assumption that $\func_1$ and $\func_2$ are monotonically increasing, we know that these two values are distinct, and furthermore, $\func_1(\val_1) - \func_2(\val_2) > \func_1(\val_2) - \func_2(\val_1)$ if and only if $\val_1 > \val_2$. Hence, the relative ordering of these two values is identical to the relative ordering of $\val_1$ and $\val_2$, reducing our canonical setting to the two-envelope problem. Our estimator $\quantitycardourscanonical$ also carries a close connection to Cover's estimator to the two-envelope problem. Specifically, Cover's estimator can be equivalently viewed as being designated by a ``switching function''~\cite{mcdonnell2009switch}. This switching function specifies the probability to ``switch'' (that is, to guess that the unobserved value is larger), and is a monotonically-decreasing function in the observed value. The use of the monotonic function $\funcmonotone$ in our estimator in~\eqref{eq:canonical_estimator} is similar in spirit.

The two-envelope problem can also be alternatively viewed as a secretary problem with two candidates. Negative results have been shown regarding the effect of cardinal vs. ordinal data when there are more than two candidates~\cite{silverman1992googol,gnedin1994googol}, and positive result has been shown on extensions of the secretary problem to different losses~\cite{gnedin1996exchangeable}.

Our original inspiration for our proposed estimator arose from Stein's phenomenon~\cite{stein1956inadmissibility} and empirical Bayes~\cite{robbins1956empirical}. This inspiration stems for the fact that the two items are not to be estimated in isolation, but in a joint manner. That said, a significant fraction of the work (e.g.,~\cite{robbins1956empirical,stein1956inadmissibility,james1961estimation,baranchik1970minimax,bock1975minimax,tian2017population}) in these areas is based on deterministic estimators. In comparison, our negative result for all deterministic estimators (Theorem~\ref{thm:canonical_det_fails}) and the positive result for our randomized estimator (Theorem~\ref{thm:canonical_ours}) provide interesting insights in this space.


\subsection{A/B testing}\label{sec:abtest}
We now demonstrate how to use the result in the canonical setting as a plug-in for more general scenarios. Specifically, we construct simple extensions to our canonical estimator, as a proof-of-concept for the superiority of cardinal data over ordinal data in A/B testing (this section) and ranking (Section~\ref{sec:ranking}).
A/B testing is concerned with the problem of choosing the better of two given items, based on multiple evaluations of each item, and is used widely for the web and e-commerce (e.g.~\cite{kohavi2009guide}). In many applications of A/B testing, the two items are rated by disjoint sets of individuals (for example, when comparing two web designs, each user sees one and only one design). It is therefore important to take into account the different calibrations of different individuals, and this problem fits in our setting with $\numitems = 2$ items and $\numgraders$ \graders. For simplicity, we assume that $\numgraders$ is even. We consider the assignment obtained by assigning item $1$ to some $\numgradershalf$ \graders chosen uniformly at random (without replacement) from the set of $\numgraders$ \graders, and assigning item $2$ to the remaining $\numgradershalf$ \graders.\footnote{Our results also hold in the following settings: (a) Each \grader is assigned one of the two items independently and uniformly at random. (b) \Graders are grouped (in any arbitrary manner) into $\numgradershalf$ pairs, and within each pair, the two \graders are assigned one distinct item each uniformly at random.}

As in the canonical setting we studied earlier, in the absence of any direct comparison between the two items, a natural ordinal estimator in the A/B testing setting is a random guess:
\begin{align*}
    \quantityordabtest(\assignment, \{\}) =  \begin{cases}
        1 \greateritem 2 & \text{with probability } 0.5\\
        2 \greateritem 1 & \text{with probability } 0.5.
    \end{cases}
\end{align*}

For concreteness, we consider the following method of performing the random assignment of the two items to the $\numgraders$ \graders. We first perform a uniformly random permutation of the $\numgraders$ \graders, and then assign the first $\numgradershalf$ \graders in this permutation to item $1$; we assign the last $\numgradershalf$ \graders in this permutation to item $2$. We let $\info_{1}^{(1)}, \ldots, \info_{1}^{(\numgradershalf)}$ denote the scores given by the $\numgradershalf$ \graders to item $1$, and let  $\info_{2}^{(1)}, \ldots, \info_{2}^{(\numgradershalf)}$ denote the scores given by the $\numgradershalf$ \graders assigned to item $2$. Namely, the \graders (in the permuted order) provide the scores $[\info_1^{(1)}, \ldots, \info_1^{(\numgradershalf)}, \info_2^{(1)}, \ldots, \info_2^{(\numgradershalf)}]$. Now consider the following standard (deterministic) estimators:
\begin{itemizex}
    \item \emph{Sign estimator:} The sign estimator outputs the item which has more pairwise wins:\\ $\sum_{\idxgrader=1}^{\numgradershalf} \indicatorevent{\info_1^{(\idxgrader)} > \info_2^{(\idxgrader)}} \symcompare{1 \succ 2}{2 \succ 1} \sum_{\idxgrader=1}^{\numgradershalf} \indicatorevent{\info_2^{(\idxgrader)} > \info_1^{(\idxgrader)}}$.
    \item \emph{Mean estimator:} The mean estimator outputs the item with the higher mean \score:\\ $\mean(\info_1^{(1)}, \ldots, \info_1^{(\numgradershalf)}) \symcompare{1 \succ 2}{2 \succ 1} \mean(\info_2^{(1)}, \ldots, \info_2^{(\numgradershalf)})$.
    \item \emph{Median estimator:} The median estimator outputs the item with the higher  median \score (upper median if there are multiple medians)\footnote{For values $a_1 \ge \cdots \ge a_n$, we define the median function as the upper median, $\median(a_1, \ldots, a_n) = a_{\floor{(n +1)/ 2}}$. Theorem~\ref{thm:abtest_det_examples_fails} also holds instead for the lower median $a_{\floor{(n+2) / 2}}$, and the median defined as the mean of the two middle values, $(a_{\floor{(n+1) / 2}} + a_{\floor{(n+2) / 2}}) / 2$.}: $\median(\info_1^{(1)}, \ldots, \info_1^{(\numgradershalf)})  \symcompare{1 \succ 2}{2 \succ 1} \median(\info_2^{(1)}, \ldots, \info_2^{(\numgradershalf)})$.
\end{itemizex}
In each estimator, ties are assumed to be broken uniformly at random.

We now show that despite using the \scores given by all $\numgraders$ \graders, where $\numgraders$ can be arbitrarily large, these natural estimators fail to uniformly dominate the na\"ive random-guessing ordinal estimator.

\begin{theorem}\label{thm:abtest_det_examples_fails}
For any (even) number of \graders, none of the sign, mean, and median estimators can strictly uniformly dominate the random-guessing estimator $\quantityordabtest$.
\end{theorem}

The negative result of Theorem~\ref{thm:abtest_det_examples_fails} demonstrates the challenges even when one is allowed to collect an arbitrarily large number of \scores for each item. Intuitively, the more \graders there are, the more miscalibration functions they introduce. Even if the statistics used by these estimators converge as the number of the \graders $\numgraders$ grows large, these values are not guaranteed to be informative towards comparing the values of the items due to the miscalibrations.

The failure of these standard estimators suggests the need of a novel approach towards this problem of A/B testing under arbitrary miscalibrations. To this end, we build on top of our canonical estimator $\quantitycardourscanonical$ from Section~\ref{sec:canonical}, and present a simple randomized estimator $\quantitycardoursabtest$ as follows:

\begin{enumeratex}[(1)]
    \item For every $\idxgrader \in [\numgradershalf]$, use the canonical estimator $\quantitycardourscanonical$ on the $\idxgrader^{th}$ pair of \scores $(\info_1^{(\idxgrader)}, \info_2^{(\idxgrader)})$ and obtain the estimate $\estfrompair_\idxgrader \defn \quantitycardourscanonical(\info_1^{(\idxgrader)}, \info_2^{(\idxgrader)}) \in \{1 \greateritem 2, 2 \greateritem 1\}$.
    \item Set the output $\quantitycardoursabtest$ as the outcome of the majority vote among the estimates $\{\estfrompair_\idxgrader\}_{\idxgrader\in [\numgradershalf]}$ with ties broken uniformly at random.
\end{enumeratex}

The following theorem now shows that the results for the canonical setting from Section~\ref{sec:canonical} translate to this A/B testing application.
\begin{theorem}\label{thm:abtest_ours}
For any (even) number of \graders, the estimator $\quantitycardoursabtest$ strictly uniformly dominates the random guessing estimator $\quantityordabtest$.
\end{theorem}
This result thus illustrates the use of our canonical estimator $\quantitycardourscanonical$ as a plug-in for A/B testing. So far we have considered settings where there are only two items and where each \grader is assigned only one item, thereby making the ordinal information vacuous. We now turn to an application that is free of these restrictions.


\subsection{Ranking}\label{sec:ranking}
It is common in practice to estimate the partial or total ranking for a list of items by soliciting ordinal or cardinal responses from individuals. In conference reviews or peer-grading, each reviewer is asked to rank~\cite{douceur2009paper,shah2013case,shah2017design} or rate~\cite{ge13bias,piech2013tuned,shah2017design} a small subset of the papers, and this information is subsequently used to estimate a partial or total ranking of the papers (or student homework). Other applications for aggregating rankings include voting~\cite{young1988condorcet,procaccia2016vote}, crowdsourcing~\cite{shah2016estimation,shah18simple}, recommendation systems~\cite{freund2003boosting} and meta-search~\cite{dwork2001rank}.

Formally, we let $\numitems > 2$ denote the number of items and $\numgraders$ denote the number of \graders. For simplicity, we focus on a setting where each \grader reports noiseless evaluations of some pair of items, and the goal is to estimate the total ranking of all items. We consider a random design setup where the pairs compared are randomly chosen and randomly assigned to \graders. We assume $1<\numgraders < \binom{\numitems}{2}$ so that the problem does not degenerate. Each \grader\xspace\grades a pair of items, and these pairs are drawn uniformly without replacement from the ${\numitems \choose 2}$ possible pairs of items. We let $\assignment = (\itemsofgrader_1,\ldots,\itemsofgrader_\numgraders)$ denote these $\numgraders$ pairs of items to be \graded by the $\numgraders$ respective \graders, where $\itemsofgrader_\idxgrader\in [\numitems]\times[\numitems]$ denotes the pair of items \graded by \grader $\idxgrader\in [\numgraders]$. 
For each pair $\itemsofgrader_\idxgrader = (\idxitem, \idxitem')$, denote the cardinal evaluation as $\info(\itemsofgrader_\idxgrader) = (\func_{\idxgrader}(\val_\idxitem), \func_{\idxgrader}(\val_{\idxitem'}))$, and the ordinal evaluation as the induced ranking $\infoord(\itemsofgrader_\idxgrader) \in \{\idxitem \greateritem \idxitem', \idxitem' \greateritem \idxitem\}$. Denote the set of ordinal observations as $\infoordall = \{\infoord(\itemsofgrader_\idxgrader)\}_{\idxgrader=1}^\numgraders$, and the set of cardinal observations as $\infocardall = \{\info(\itemsofgrader_\idxgrader)\}_{\idxgrader=1}^\numgraders$.
The input to an ordinal estimator is the ordinal information $\infoordall$. The input to a cardinal estimator is the \grader assignment $\assignment$ and the set of cardinal observations $\infocardall$.
Finally, let $\graph( \infoordall)$ denote a directed acyclic graph (DAG) with nodes comprising the $\numitems$ items and with an edge from any node $\idxitem$ to any other node $\idxitem'$ if and only if $\{\idxitem\greateritem \idxitem'\} \in \infoordall$. A topological ordering on $\graph$ is any total ranking of its vertices which does not violate any pairwise comparisons indicated by $\infoordall$.

\begin{algorithm}[t!]
\DontPrintSemicolon
    Deduce the ordinal observations $\infoordall$ from the cardinal observations $\infocardall$.\;
    Compute a topological ordering $\quantityest$ on the graph $\graph(\infoordall)$, with ties broken in order of the indices of the items.\;
    $\position \leftarrow 1$.\;
    \While{$\position < \numitems$}{
        Let $\quantityest_\text{flip}$ be the ranking obtained by flipping the positions of the $\position^{th}$ and the $(\position+1)^{th}$ items in $\quantityest$.\;
        \eIf{$\quantityest_\text{flip}$ is a topological ordering on $\graph( \infoordall)$, and both the $\position^{th}$ and $(\position+1)^{th}$ items are \graded by at least one \grader each in $\infocardall$\label{line:if_flippable}}{
            From all of the scores of the $\position^{th}$ item in $\infocardall$, sample one uniformly at random and denote it as $\info_{\quantityest(\position)}$. Likewise denote $\info_{\quantityest(\position+1)}$ as a randomly chosen \score of the $(\position + 1)^{th}$ item from $\infocardall$.\;
            Consider the two \graders reporting the \scores $\info_{\quantityest(\position)}$ and $\info_{\quantityest(\position+1)}$. Remove from $\infocardall$ all scores provided by these two \graders.\;
            \If{$\quantitycardourscanonical(\info_{\quantityest(\position)}, \info_{\quantityest(\position+1)})$ outputs $ {\quantityest(\position+1)}\greateritem {\quantityest(\position)}$\label{line:call_canonical}}{
                $\quantityest\leftarrow \quantityest_\text{flip}$.\;
            }
            $\position \leftarrow \position + 2$.\;
        }{
            $\position\leftarrow \position + 1$.\;
        }
    }
    Output $\quantitycardourssort(\assignment, \infocardall) = \quantityest$.\;
    \caption{Our cardinal ranking estimator $\quantitycardourssort(\assignment, \infocardall)$.}\label{alg:sort_cardinal}
\end{algorithm}

We now present our (randomized) cardinal estimator $\quantitycardourssort(\assignment, \infocardall)$ in Algorithm~\ref{alg:sort_cardinal}. In words, this algorithm start from any topological ordering of the items as the initial estimate of the true ranking. Then the algorithm scans one-by-one over the pairs with adjacent items in the initial estimated ranking. If a pair can be flipped (that is, if the ranking after flipping this pair is also a topological ordering), we uniformly sample a pair of \scores for these two items from the cardinal observations $\infocardall$, and use the randomized estimator $\quantitycardourscanonical$ to determine the relative order of the pair. After $\quantitycardourscanonical$ is called, the positions of this pair are finalized. We remove all \scores of these two \graders from future use, and jump to the next pair that does not contain these two items.

The following theorem now presents the main result of this section.
\begin{theorem}\label{thm:sort_ours_uniform_prior}
Suppose that the true ranking $\quantitygt$ is drawn uniformly at random from the collection of all possible rankings, and consider any ordinal estimator $\quantityordsort$ for $\quantitygt$. Then the cardinal estimator $\quantitycardourssort$ strictly uniformly dominates the ordinal estimator $\quantityordsort$.
\end{theorem}
We note that  Algorithm~\ref{alg:sort_cardinal} runs in polynomial time (in the number of items $\numitems$) because the two major operations of this estimator -- finding a topological ordering, and checking if a ranking is a topological ordering on the DAG -- can be implemented in polynomial time~\cite{dasgupta2008algorithms}. Theorem~\ref{thm:sort_ours_uniform_prior} thus demonstrates again the power of the canonical estimator $\quantitycardourscanonical$ as a plug-in component to be used in a variety of applications. An extension of our results to the setting where $\quantitygt$ can be arbitrary (adversarially chosen) is presented in Appendix~\ref{app:ranking_arbit}.

\section{Simulations}
We now experimentally evaluate our proposed estimators for A/B testing and ranking. Since the performance of the ordinal estimators vary significantly in different problem instances, we use the notion of ``relative improvement''. The relative improvement $\improve{\capours{\quantity}}{\capord{\quantity}}$ of an estimator $\capours{\quantity}$ as compared to a baseline estimator $\capord{\quantity}$ is defined as: $
    \improve{\capours{\quantity}}{\capord{\quantity}} = \frac{\Expect[\loss(\quantitygt, \capord{\quantity})] - \Expect[\loss(\quantitygt, \capours{\quantity})]}{\Expect[\loss(\quantitygt, \capord{\quantity})]} \times 100\%.$
A positive value of the relative improvement $\improve{\capours{\quantity}}{\capord{\quantity}}$  indicates the superiority of estimator $\capours{\quantity}$ over the estimator $\capord{\quantity}$. A relative improvement of zero indicates an identical performance of the two estimators. In our proposed estimators, the function $\funcmonotone$ is set as $\funcmonotone(\tmpx) = \frac{\tmpx}{1 + \tmpx}$.


\subsection{A/B testing}\label{sec:exp_abtest}
    We now present simulations to evaluate various points on the bias-variance tradeoff. For A/B testing, we compare our estimator $\quantitycardoursabtest$ with other standard estimators --- the sign, mean and median estimators introduced in Section~\ref{sec:abtest}. The item values $\val_1$ and $\val_2$ are chosen independently and uniformly at random from the interval $[0,1]$. The calibration functions are linear and given by:
    \begin{enumeratex}[(a)]
        \item \emph{One biased \grader:} One \grader gives an abnormally (high or low) \score. Formally,  $\func_\idxgrader(\val) = \val$ for $ \idxgrader \in [\numgraders-1]$, and $\func_\numgraders(\val) = \val + \numgraders$.\label{item:one_biased_grader}
        \item \emph{Incremental biases:} Calibration functions of reviewers are shifted from each other. Formally,  $\func_\idxgrader(\val) = \val + \idxgrader$ for $\idxgrader\in [\numgraders]$.\label{item:incremental biases}
        \item \emph{Incremental biases with one biased \grader:} A combination of setting~\ref{item:one_biased_grader} and setting~\ref{item:incremental biases}. Formally, $\func_\idxgrader(\val) = \val + (\idxgrader-1)$ for $\idxgrader \in [\numgraders - 1]$, and $\func_\numgraders(\val) = \val + \frac{\numgraders(\numgraders-1)}{2}$.
    \end{enumeratex}
    
\begin{figure}
    \centering
    \captionsetup[subfigure]{oneside,margin={1.9cm,0cm}}
    \subfloat[one biased \grader]{\includegraphics[height=3.9cm]{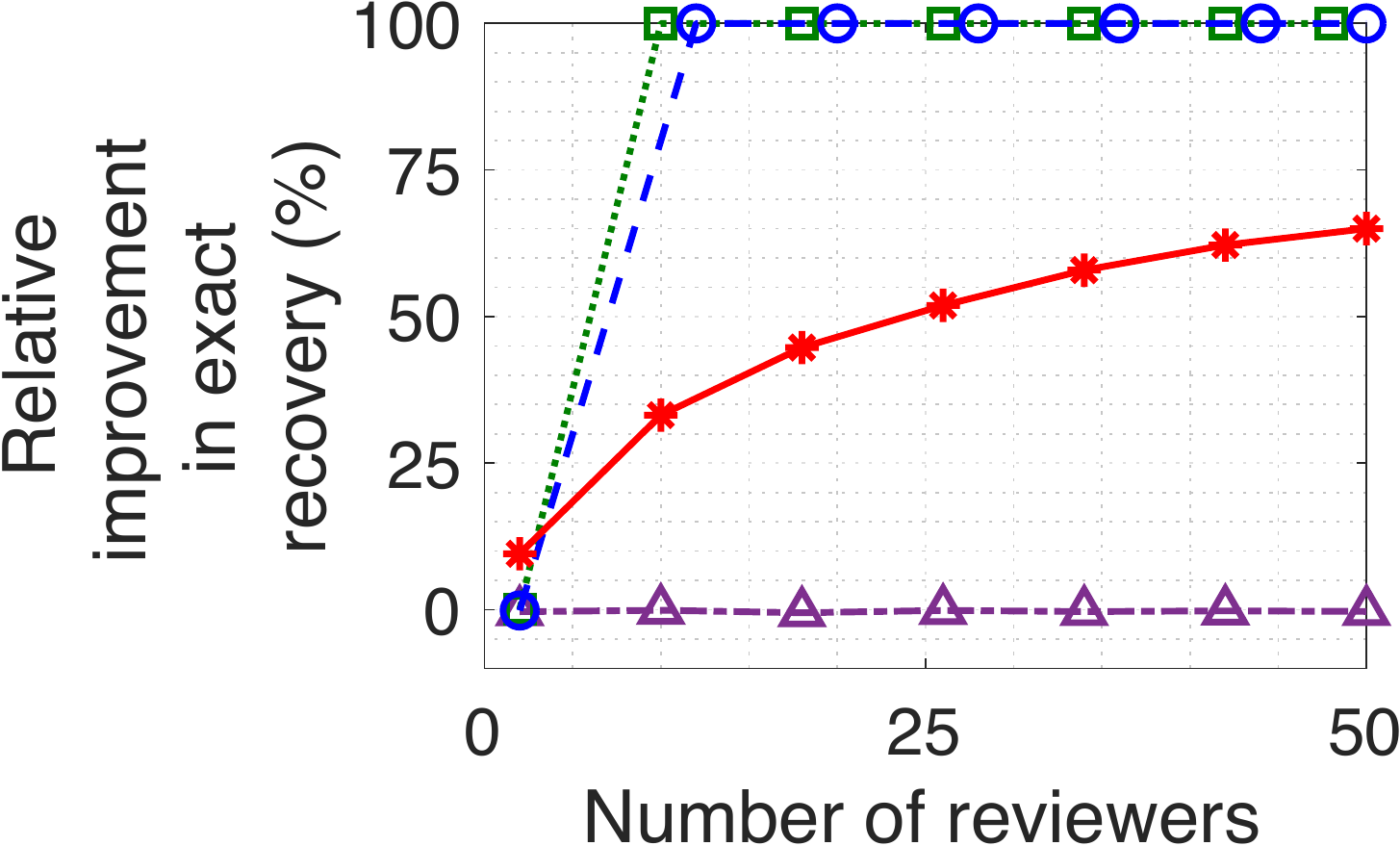}\label{fig:abtest_adversarial_avg}}~~~
    \captionsetup[subfigure]{oneside,margin={0cm,0cm}}
    \subfloat[incremental biases]{\includegraphics[height=3.9cm]{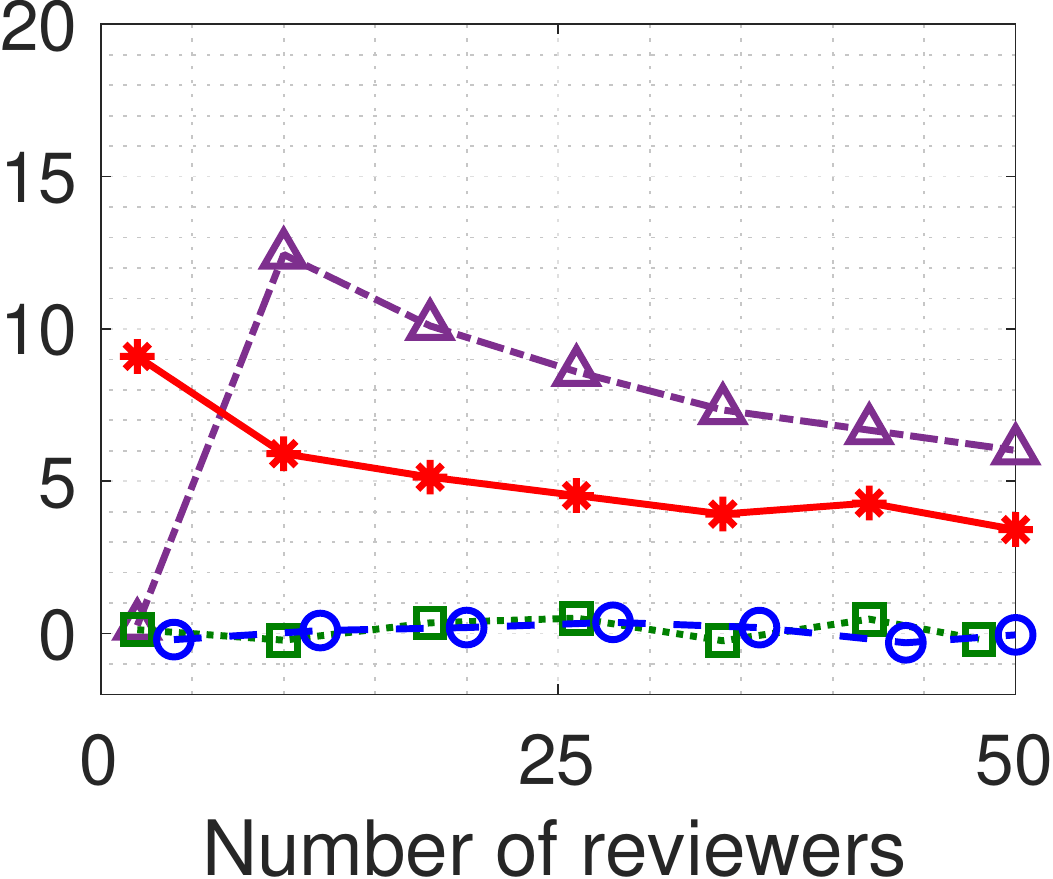}\label{fig:abtest_adversarial_sign}}~~~
    \subfloat[incremental biases with one biased \grader]{\includegraphics[height=3.9cm]{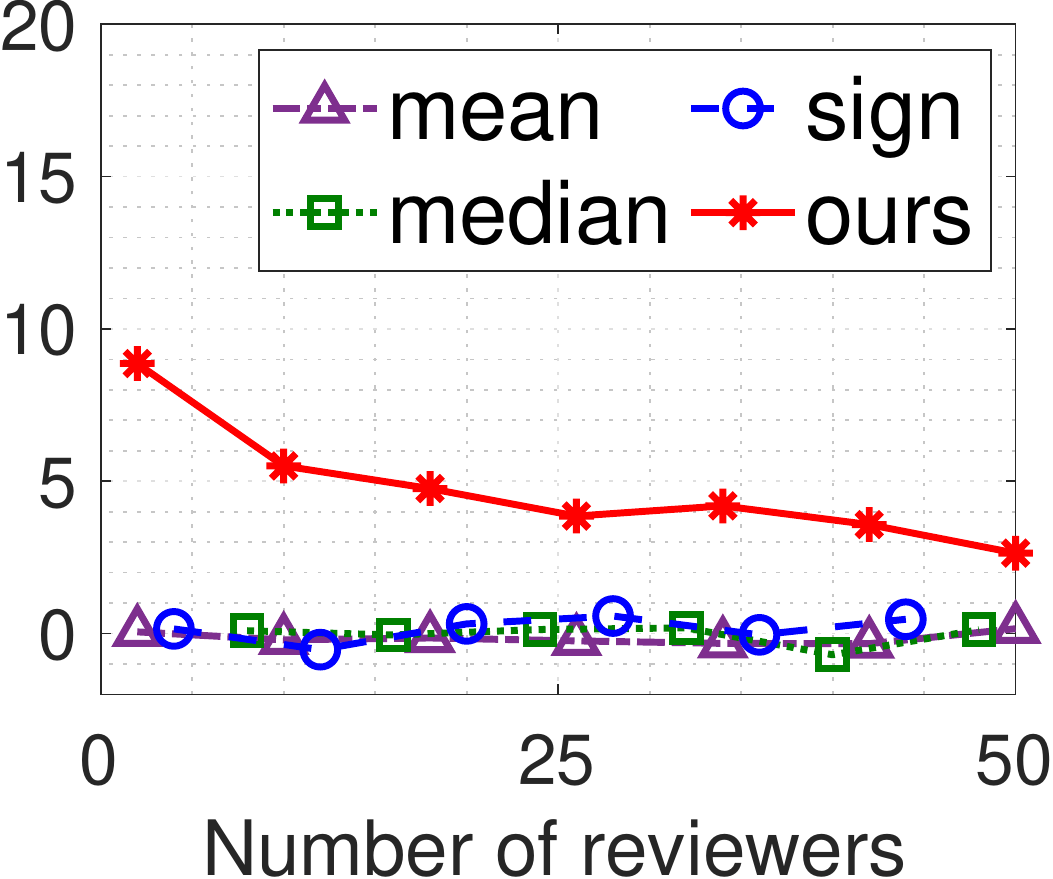}\label{fig:abtest_adversarial_both}}~~~
    \caption{\label{fig:abtest}   
    Relative improvement in exact recovery of various estimators as compared to the random-guessing ordinal estimator $\quantityordabtest$ for A/B testing. Each point is an average over $10,000$ trials. The error bars are too small to display.
    }
\end{figure}

    We simulate and compute the relative improvement of the different estimators as compared to the random-guessing estimator $\quantityordabtest$. The results are shown in Figure~\ref{fig:abtest}. While the performance of the estimators vary with respect to each other, our estimator consistently beats the baseline whereas every other estimator fails. Our estimator thus indeed operates at a unique point on the bias-variance tradeoff with a low (zero) bias and a variance strictly smaller than the ordinal estimators, whereas all other estimators incur a non-zero error due to bias.

\subsection{Ranking}\label{sec:exp_sort}
\begin{figure}
    \centering
    \includegraphics[width=0.8\linewidth]{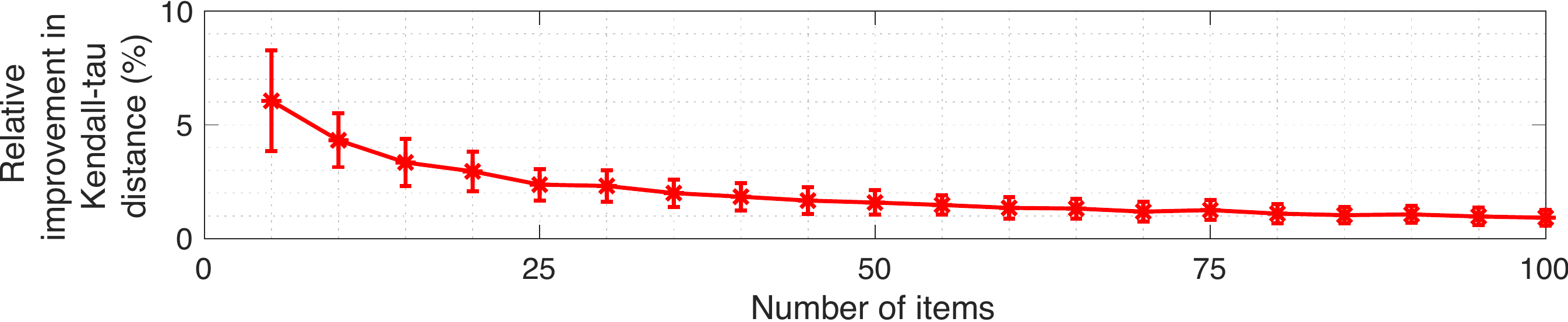}
    \caption{\label{fig:sort}
    Relative improvement in Kendall-tau distance of our ranking estimator $\quantitycardourssort$ as compared to an optimal ordinal estimator $\quantityordsort$ for ranking. Each point is an average over $100$ trials, where in each trial the quantities $\Expect[\loss(\quantitygt, \quantitycardourssort)]$ and $\Expect[\loss(\quantitygt, \quantityordsort)]$ are approximated by an empirical average over $1000$ samples.
    }
\end{figure}

    Next, we evaluate the performance of our ranking estimator $\quantitycardourssort$ when the true ranking $\quantitygt$ is drawn from a uniform prior. We compare this estimator with an optimal ordinal estimator $\quantityordsort$ which outputs a topological ordering with ties broken in order of the indices of the items (this ordinal estimator is optimal regardless of the tie-breaking strategy).
    
    For any number of items $\numitems$, we generate the values $\val_1,\ldots,\val_\numitems$ of the items i.i.d. uniformly from the interval $[0, \numitems]$. We set $\numgraders = \floor{\frac{1}{2}{\numitems \choose 2}}$. We assume that the $\idxgrader^{th}$ \grader has a linear calibration function $\func_\idxgrader(\val) = \scale_\idxgrader\val + \bias_\idxgrader$, where we sample $\scale_\idxgrader$ and $\bias_\idxgrader$ i.i.d. uniformly from the interval $[0, 1]$.
    
    We have previously proved that our estimator $\quantitycardourssort$ based on cardinal data can strictly uniformly outperform the optimal ordinal estimator for the 0-1 loss. We use these simulations to evaluate the efficacy of our approach for a different loss function -- Kendall-tau distance. Specifically, Figure~\ref{fig:sort} compares these two estimators in terms of Kendall-tau distance (Appendix~\ref{app:other_metrics} provides a formal definition of this distance and associated theoretical results). We observe that our estimator $\quantitycardourssort$ is able to consistently yield improvements even for this loss. The reason that the improvement becomes smaller when the number of items is large is that by flipping pairs, our estimator only modifies the ranking in the neighborhood of the initial estimate. We strongly believe that it should be possible to design better estimators for the large $\numitems$ regime using the tools developed in this paper. Having met our stated goal of  outperforming ordinal estimators to handle arbitrary miscalibrations, we leave this interesting problem for future work.
    
\subsection{Tradeoff between estimation under perfect calibration vs. miscalibration}
In this section, we present a preliminary experiment showing the tradeoff between estimation under perfect calibration (all reviewers reporting the true values of the papers) and estimation under miscalibration. For simplicity, we consider the canonical setting from Section~\ref{sec:canonical}. We evaluate the performance of our estimator under two scenarios: (1) perfect calibration, where $\func_\idxgrader(\val) = \val$ for each $\idxgrader\in \{1, 2\}$; (2) miscalibration with one biased reviewer, where $\func_1(\val) = \val$ and $\func_2(\val) = \val + 1$. We consider the function $\funcmonotone$ in our estimator as $\funcmonotone(\val) = \frac{\funcmonotoneparam\val}{1 + \funcmonotoneparam\val}$, where $\funcmonotoneparam\in \{2^k \given -10 \le k \le 10, k \in \integers\}$. We sample $\val_1$ and $\val_2$ uniformly at random from the interval $[0, 1]$.  

\begin{figure}[b!]
    \centering
    \includegraphics[width=0.4\linewidth]{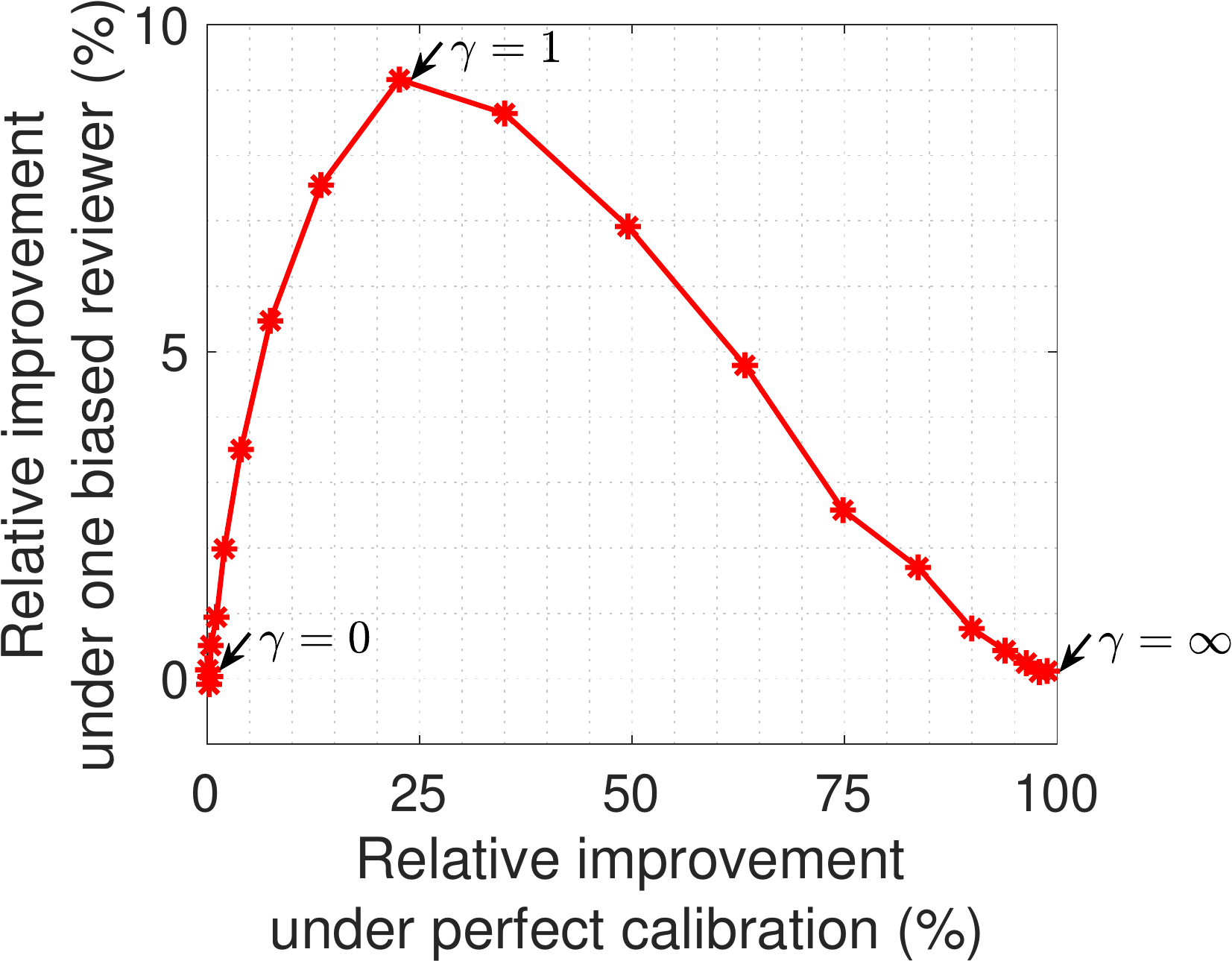}
    \caption{\label{fig:canonical}
        Relative improvement of our canonical estimator $\quantitycardourscanonical$ under perfect calibration and under miscalibration of one biased reviewer, with $\funcmonotone(\val) = \frac{\funcmonotoneparam\val}{1 + \funcmonotoneparam\val}$ and $\funcmonotoneparam\in \{2^k \given -10 \le k \le 10, k \in \integers\}$, where $\funcmonotoneparam$ increases from left to right in the plot. Each point is an average over $5\times 10^5$ trials. The error bars are too small to display.}
\end{figure}

Figure~\ref{fig:canonical} shows the relative improvement of our estimator over the random-guessing baseline under perfect calibration and under miscalibration, where $\funcmonotoneparam$ increases from left to right. Let us focus on a few regimes in this plot. First, on the left end of the curve, when $\funcmonotoneparam$ is close to $0$, we have $\funcmonotone(\val)$ close to $0$. The estimator is close to random-guessing. At the other extreme, on the right end of the curve, when $\funcmonotoneparam$ goes to infinity, we have $\funcmonotone(\val)$ close to $1$. The estimator always outputs the item with the higher score, and hence gives perfect estimation under perfect calibration. Under miscalibration, this estimator always chooses the biased reviewer giving the higher score and hence performs the same as random guess. Past the maximum point of the function at approximately $(25\%, 9\%)$ when $\funcmonotoneparam = 1$, the value of the curve starts decreasing, suggesting a tradeoff of estimation accuracy under perfect calibration and under miscalibration. It is clear that points to the left of the maximum point are not Pareto-efficient, since there exist other points with the same accuracy under miscalibration but improved accuracy under perfect calibration. 

We thus see that robustness under arbitrary miscalibration comes at a cost of lower accuracy under perfect calibration. Establishing a formal understanding of this tradeoff and designing estimators that are provably Pareto-efficient are important open problems.

\section{Proofs}\label{app:proofs}
In this section, we present the proofs of our theoretical results.

For notational simplicity, we use ``$1 \lessitem 2$'' to denote that item $1$ has a smaller value than item $2$. Since the items have distinct values, we have $1\lessitem 2$ if and only if $2 \greateritem 1$. For the 0-1 loss $\loss(\quantitygt, \quantityest) = \indicatorevent{\quantityest \ne \quantitygt}$, we call the expected loss $\Expect[\loss(\quantitygt, \quantityest)] = \Prob(\quantityest \ne \quantitygt)$ as the ``probability of error'' of any estimator $\quantityest$, and $\Prob(\quantityest = \quantitygt)$ as the ``probability of success''. For the canonical setting and A/B testing, the probability of success of random guessing is $0.5$. To show that some estimator $\quantityest$ strictly uniformly dominates random guessing for the canonical setting or A/B testing, we only need to show that the probability of success of this estimator is strictly higher than $0.5$, or equivalently, the probability of error of of this estimator is strictly lower than $0.5$. 

\subsection{Proof of Theorem~\ref{thm:canonical_det_fails}}
We prove that no deterministic cardinal estimator can strictly uniformly dominate the random-guessing estimator $\quantityordcanonical$, which implies the negative result for any deterministic ordinal estimator.

Recall the notation $\topitem = \argmax_{\idxitem\in \{1, 2\}} \info_\idxitem$ as the item receiving the higher \score (with ties broken uniformly at random), and the notation $\bottomitem$ as the remaining item. First, we consider a deterministic estimator that always outputs $\topitem$ as the item whose value is greater. We call this estimator the ``sign estimator'', denoted $\quantityestsign$:
\begin{align*}
    \quantityestsign(\assignment, \info_1, \info_2) = (\topitem\greateritem\bottomitem).
\end{align*}

The proof consists of two steps. (1) We show that the sign estimator does not strictly uniformly dominate random guess. (2) Building on top of (1), we show that more generally, no deterministic estimator strictly uniformly dominates random guess.

\noindent\textbf{Step 1: } The sign estimator does not strictly uniformly dominate random guess.

We construct the following counterexample such that the probability of error of the sign estimator is $0.5$. We construct \grader calibration functions such that their ranges are disjoint, that is, one \grader always gives a higher \score than the other \grader, regardless of the items they are assigned. Then the relative ordering of the two \scores does not convey any information about the relative ordering of the two items, and we show that in this case, the sign estimator has a probability of error of $0.5$. Concretely, let the item values be bounded as $\val_1, \val_2\in (0, 1)$, and let the calibration functions be $\func_1(\val) = \val$ and $\func_2(\val) = \val + 1$. Then the \score given by \grader $2$ is higher than the \score given by \grader $1$ regardless of the item values they are assigned. The sign estimator always observes $\info_1 < \info_2$, and outputs the item assigned to \grader $2$ as the larger item. The assignment is either $\assignment = (\itemsofgrader_1 = 1, \itemsofgrader_2 = 2)$ or $(\itemsofgrader_1 = 2, \itemsofgrader_2 = 1)$ with probability $0.5$ each. Under assignment $(\itemsofgrader_1 = 1, \itemsofgrader_2 = 2)$, the sign estimator outputs $1 \lessitem 2$. Under assignment $(\itemsofgrader_1 = 2, \itemsofgrader_2 = 1)$, the sign estimator outputs $1\greateritem 2$. Under one (and exactly one) of the two assignments, the output of the sign estimator is correct. Hence, the probability of error of the sign estimator is $0.5$.

\noindent\textbf{Step 2: } No deterministic estimator strictly uniformly dominates random guess.

Let $\setassignments$ be the set of the two assignments, $\setassignments = \{(\itemsofgrader_1 = 1, \itemsofgrader_2 = 2), (\itemsofgrader_1 = 2, \itemsofgrader_2 = 1)\}$. A deterministic estimator $\quantityestdet: \setassignments \times \reals \times \reals \rightarrow \{1 \greateritem2, 1\lessitem 2\}$ is a deterministic function that takes as input the assignment and the \scores for the two items, and outputs the relative ordering between the two items. Step 1 has shown that the sign estimator does not strictly uniformly dominate random guess. Hence, we only need to prove that any deterministic estimator $\quantityestdet$ that is different from the sign estimator does not strictly uniformly dominate random guess. For this deterministic estimator $\quantityestdet$, there exist some input values $(\assignmentobs, \widetilde{\info}_1, \widetilde{\info}_2)$ such that the output of this deterministic estimator differs from the sign estimator. If the two estimators $\quantityestsign$ and $\quantityestdet$ only differ at points where $\widetilde{\info}_1 = \widetilde{\info}_2$, then we can use the same counterexample in Step 1 to show that the probability of error of this deterministic estimator is $0.5$. It remains to consider the case when $\widetilde{\info}_1 \ne \widetilde{\info}_2$. Without loss of generality, assume $\widetilde{\info}_1 > \widetilde{\info}_2$. Then consider the following counterexample. Let $\val_1 > \val_2$. Let $\func_1, \func_2$ be strictly-increasing functions such that $\func_1(\val_1) = \func_2(\val_1) = \widetilde{\info}_1, \func_1(\val_2) = \func_2(\val_2) = \widetilde{\info}_2$. Regardless of the \grader assignment, the \score $\info_1$ for item $1$ is $\widetilde{\info}_1$, and the \score $\info_2$ for item $2$ is $\widetilde{\info}_2$. The item receiving a higher \score is always $\topitem = \argmax_{\idxitem\in\{1, 2\}}\info_\idxitem = 1$, so the sign estimator $\quantityestsign$ always outputs $1\greateritem 2$. Under assignment $\assignmentobs$, the deterministic estimator differs from the sign estimator, so the deterministic estimator gives the incorrect output $(1 \lessitem 2)$. The assignment $\assignmentobs$ happens with probability $0.5$, so the probability of error of this deterministic estimator is at least $0.5$.\newline

The two steps above complete the proof that there exists no deterministic estimator that strictly uniformly dominates random guess.


\subsection{Proof of Theorem~\ref{thm:canonical_ours}}
In what follows, we prove that the probability of success of our estimator is strictly greater than $0.5$ under arbitrary item values $\val_1, \val_2$ and arbitrary calibration functions $\func_1, \func_2$. We start with re-writing our estimator in~\eqref{eq:canonical_estimator} into an alternative and equivalent expression, and then prove the result on this new expression of our estimator.

Recall that $\topitem = \argmax_{\idxitem\in \{1, 2\}} \info_\idxitem$ denotes the item receiving the higher \score, and $\bottomitem$ denotes the remaining item (with ties broken uniformly). Depending on the relative ordering of $\info_1$ and $\info_2$, we can split~\eqref{eq:canonical_estimator} into the following three cases:

\begin{subequations}\label{eq:canonical_rewrite_three_cases}
\begin{align}
    \quantitycardourscanonical(\assignment, \info_1, \info_2 \given \info_1 > \info_2) & = 
        \begin{cases}
        1 \greateritem 2 & \text{with probability $\frac{1 + \funcmonotone(\info_1 - \info_2)}{2}$}\\
        2 \greateritem 1 &\text{otherwise,}
        \end{cases}\\
    \quantitycardourscanonical(\assignment, \info_1, \info_2 \given \info_1 < \info_2) & = 
        \begin{cases}
        1 \greateritem 2 & \text{with probability $\frac{1 - \funcmonotone(\info_2 - \info_1)}{2}$}\\
        2 \greateritem 1 &\text{otherwise,}
        \end{cases}\\
    \quantitycardourscanonical(\assignment, \info_1, \info_2 \given \info_1 = \info_2) & = 
        \begin{cases}
        1 \greateritem 2 & \text{with probability $\frac{1}{2}$}\\
        2 \greateritem 1 &\text{otherwise.}
        \end{cases}
\end{align}
\end{subequations}

Recall that the function $\funcmonotone$ is from $\realsnonneg$ to $[0,1)$. Now we define the following auxiliary function $\funcmonotoneredefine: \reals \rightarrow (0, 1)$:
\begin{align}\label{eq:envelope_monotonic_function}
    \funcmonotoneredefine(\val) = \begin{cases}
    \frac{1 + \funcmonotone(\val)}{2} & \text{if } \val > 0\\
    \frac{1}{2} & \text{if }\val = 0\\
    \frac{1 - \funcmonotone(-\val)}{2} & \text{otherwise.}
    \end{cases}
\end{align}

Combining~\eqref{eq:canonical_rewrite_three_cases} and~\eqref{eq:envelope_monotonic_function}, we have
\begin{align}\label{eq:canonical_rewrite}
    \quantitycardourscanonical(\assignment, \info_1, \info_2) = 
    \begin{cases}
        1 \greateritem 2 & \text{with probability } \funcmonotoneredefine(\info_1 - \info_2)\\
        2 \greateritem 1 & \text{otherwise.}
    \end{cases}
\end{align}

Without loss of generality, assume $\val_1 > \val_2$. The assignment is either $\assignmentobs \defn (\itemsofgrader_1 = 1, \itemsofgrader_2 = 2)$ or $\assignmentobs' \defn (\itemsofgrader_1 = 2, \itemsofgrader_2 = 1)$ with probability $0.5$ each. Thus, the estimator observes $\{\info_1 = \func_1(\val_1), \info_2 = \func_2(\val_2)\}$ under assignment $\assignmentobs$, or $\{\info_1 = \func_2(\val_1), \info_2 = \func_1(\val_2)\}$ under assignment $\assignmentobs'$. The probability of success of our estimator $\quantitycardourscanonical$ is:

\begin{align}
    \Prob(\quantitycardourscanonical = \quantitygt) = & \sum_{\widetilde{\assignmentobs}\in \{\assignmentobs, \assignmentobs'\}} \Prob(\quantitycardourscanonical =\quantitygt \given \assignment = \widetilde{\assignmentobs}) \Prob(\assignment = \widetilde{\assignmentobs})\nonumber\\
    \stackrel{\stepone}{=} &  \frac{1}{2}\funcmonotoneredefine(\func_1(\val_1) - \func_2(\val_2))  + \frac{1}{2} \funcmonotoneredefine(\func_2(\val_1) - \func_1(\val_2)) \nonumber\\
    = & \frac{1}{2}\left[\funcmonotoneredefine(\func_1(\val_1) - \func_2(\val_2))  + \funcmonotoneredefine(\func_2(\val_1) - \func_1(\val_2))\right]\nonumber\\
    \stackrel{\steptwo}{=} & \frac{1}{2}\left[1 + \funcmonotoneredefine(\func_1(\val_1) - \func_2(\val_2))  - \funcmonotoneredefine( \func_1(\val_2) - \func_2(\val_1))\right],\label{eq:canonical_probability_success_intermediate}
\end{align}
where step \stepone is true by plugging in~\eqref{eq:canonical_rewrite}, and step \steptwo is true because $\funcmonotoneredefine(\val) + \funcmonotoneredefine(-\val) = 1$ by the definition of the function $\funcmonotoneredefine$ in~\eqref{eq:envelope_monotonic_function}.

By the monotonicity of the functions $\func_1$ and $\func_2$, and by the assumption that $\val_1 > \val_2$, we have $\func_1(\val_1) + \func_2(\val_1) > \func_1(\val_2) + \func_2(\val_2)$, and therefore $\func_1(\val_1) - \func_2(\val_2) > \func_1(\val_2) - \func_2(\val_1)$. Since $\funcmonotone(0) \ge 0$ and $\funcmonotone$ is monotonically increasing on $\realsnonneg$, it is straightforward to verify that $\funcmonotoneredefine$ is monotonically increasing on $\reals$. Hence, we have
\begin{align}\label{eq:canonical_func_redefine_relation}
    \funcmonotoneredefine(\func_1(\val_1) - \func_2(\val_2)) > \funcmonotoneredefine(\func_1(\val_2) - \func_2(\val_1)).
\end{align}

Combining~\eqref{eq:canonical_probability_success_intermediate} and~\eqref{eq:canonical_func_redefine_relation}, we have
\begin{align*}
    \Prob(\quantitycardourscanonical = \quantitygt) > \frac{1}{2}.
\end{align*}

\subsection{Proof of Theorem~\ref{thm:abtest_det_examples_fails}}

We construct a counterexample on which the mean, median and sign estimators have a probability of error of $0.5$. In this counterexample, let the item values be bounded as $\val_1, \val_2\in (0, 1)$, and let the $\numgraders$ \grader calibration functions be as follows:

\begin{align}
    \func_\idxgrader(\val) = 
    \begin{dcases}
        \val + (\idxgrader-1) & \text{if } 1\le \idxgrader \le \numgraders-1\\
         \val + \frac{\numgraders(\numgraders-1)}{2} & \text{if } \idxgrader=\numgraders.
    \end{dcases}\label{eq:counterexample_calibration}
\end{align}

In these calibration functions, the \score provided by each \grader is the sum of the true value of the item assigned to this \grader, and a bias term specific to this \grader. The analysis is performed separately for the three estimators. At a high level, the analysis for the mean estimator uses the fact that one \grader (specifically, \grader $\numgraders)$ has a significantly greater bias than the rest of the \graders. The analysis for the median and the sign estimators uses the fact that the ranges of these calibration functions are disjoint.

\noindent\textbf{Mean estimator:}
Recall that each \grader is assigned one of the two items. Given any assignment, consider the item assigned to \grader $\numgraders$. Trivially, the sum of the \scores for this item must be strictly greater than $\func_\numgraders(0) = \frac{\numgraders(\numgraders-1)}{2}$. Now consider the remaining item (not assigned to \grader $\numgraders$). The sum of the \scores for the remaining item can be at most $\sum_{\idxgrader=1}^{\numgraders-1} \func_\idxgrader(1) = \sum_{\idxgrader=1}^{\numgraders-1}  \idxgrader = \frac{\numgraders(\numgraders-1)}{2}$.

From these two bounds on the sum of the \scores, an item has a greater sum of \scores if and only if \grader $\numgraders$ is assigned to this item. By symmetry of the assignment, \grader $\numgraders$ is assigned to either item with probability $0.5$. With the true ranking being either $1 \greateritem 2$ or $1\lessitem 2$, the mean estimator makes an error in one of the two assignments, and this assignment happens with probability $0.5$. Hence, the mean estimator makes an error with probability $0.5$.\\

\noindent\textbf{Median estimator and sign estimator:}
For the median estimator and the sign estimator, we first present an alternative view on the assignment, which is used for the analysis of both estimators. Recall that the assignment specifies $\numgradershalf$ \graders to \grade item $1$, drawn uniformly at random without replacement, and the remaining $\numgradershalf$ \graders to item $2$. Equivalently, we can view this assignment as comprising the following two steps. (1) We sample uniformly at random a permutation of the $\numgraders$ \graders, denoted as a list $(\idxgrader_1, \ldots, \idxgrader_\numgraders)$. Define $\gradersofitem$ and $\gradersofitem'$ as the first half and second half of the \graders in the list, $\gradersofitem = (\idxgrader_1, \ldots, \idxgrader_{\numgradershalffrac})$ and $\gradersofitem' = (\idxgrader_{\numgradershalffrac + 1}, \ldots, \idxgrader_\numgraders)$. (2) We draw uniformly at random one of the two items, and assign the list $\gradersofitem$ of \graders to this item. Then assign the list $\gradersofitem'$ of \graders to the remaining item. For each $\idxpair\in [\numgradershalf]$ , call reviewers $\{\idxgrader_\idxpair, \idxgrader_{\numgradershalffrac+\idxpair}\}$ as the $\idxpair^{th}$ pair of \graders.

For the median estimator and the sign estimator, we prove that given any arbitrary lists of \graders $\gradersofitem$ and $\gradersofitem'$ in Step (1) of the assignment, the randomness in Step (2) yields the probability of error of the two estimators as $0.5$.

Recall that the item values are bounded as $\val_1, \val_2 \in (0, 1)$. Since the biases of any two \graders differ by at least $1$ in Eq.~\eqref{eq:counterexample_calibration}, any \grader $\idxgrader$ gives a higher \score than any other \grader $\idxgrader'$ if and only if $\idxgrader < \idxgrader'$, independent of the item values and the assignment. Formally, for any $\val, \val'\in (0, 1)$, and any $\idxgrader, \idxgrader' \in [\numgraders]$, we have
\begin{align}
    \func_\idxgrader(\val) < \func_{\idxgrader'}(\val')\quad \text{ if and only if } \quad \idxgrader< \idxgrader'.\label{eq:median_odd}
\end{align}

The remaining analysis is performed separately for the median estimator and the sign estimator.

\noindent\emph{Median estimator:}
Denote $\idxgrader_1^\textmedian$ and $\idxgrader_2^\textmedian$ as the indices of the \graders providing the (upper) median \scores in the set $\gradersofitem_1$ and $\gradersofitem_2$, respectively. From~\eqref{eq:median_odd}, we have
\begin{align}\label{eq:median_indices}
\begin{split}
    \idxgrader_1^\textmedian = & \median(\idxgrader_1, \ldots, \idxgrader_\numgradershalffrac)\\ \idxgrader_2^\textmedian = & \median(\idxgrader_{\numgradershalffrac + 1}, \ldots, \idxgrader_\numgraders).
\end{split}
\end{align}
Also from~\eqref{eq:median_odd}, the higher \score in the two \scores given by \grader $\idxgrader_1^\textmedian$ and $\idxgrader_2^\textmedian$ is the \grader with the larger index, $\max\{\idxgrader_1^\textmedian, \idxgrader_2^\textmedian\}$. In Step (2) of the assignment, \grader $\idxgrader_1^\textmedian$ is assigned to item $1$ or item $2$ with equal probability. Hence, the probability of error of the median estimator is $0.5$. This proves the claim that the (upper) median estimator does not strictly uniformly dominates random guess.\\

We now comment on using the median function defined as the lower median, or as the mean of the two middle values. For the lower median, the same argument as above applies. Now consider the median defined as the mean of the two middle values. When $\numgradershalf$ is odd, Eq.~\eqref{eq:median_indices} still holds, and the argument as above still applies. When $\numgradershalf$ is even, the median value may not be equal to any \scores from the \graders. We construct a counterexample where the item values are still bounded as $\val_1, \val_2\in (0, 1)$, and the calibration functions as follows:
\begin{align*}
    \func_\idxgrader(\val) =
        \val + 2^{\idxgrader}\qquad  \text{for every } \idxgrader\in [\numgraders].
\end{align*}

With these calibration functions, for any $\val, \val', \val'', \val'''\in (0, 1)$, and any $\idxgrader, \idxgrader', \idxgrader'', \idxgrader'''\in [\numgraders]$, we have
\begin{align*}
    \func_\idxgrader(\val) + \func_{\idxgrader'}(\val') < \func_{\idxgrader''}(\val'')+ \func_{\idxgrader'''}(\val''') \quad\text{ if and only if } \quad \max\{\idxgrader, \idxgrader'\} < \max\{\idxgrader'', \idxgrader'''\}.
\end{align*} 

Using this fact, we can
show that the output of this median estimator only depends on \grader indices and the realization of Step (2), independent of the item values. The probability of error of this median estimator is also $0.5$.\\

\noindent\emph{Sign estimator:}
Denote $\assignmentobs$ as the assignment that \graders in $\gradersofitem$ are assigned to item $1$, and denote $\assignmentobs'$ as the assignment that \graders in $\gradersofitem$ are assigned to item $2$. For each $\idxpair\in [\numgradershalf]$, define $\compareofpair_\idxpair\in\{0, 1\}$ as the binary value of whether the higher \score in the $\idxpair^{th}$ pair of \scores comes from item $1$, under assignment $\assignmentobs$. Set $\compareofpair_\idxpair = 1$ if the higher \score comes from item $1$ and $\compareofpair_\idxpair=0$ otherwise. Define $\compareofpair'_\idxpair\in \{0, 1\}$ similarly under assignment $\assignmentobs'$. Set $\compareofpair'_\idxpair = 1$ if the higher \score comes from item $1$, and $\compareofpair'_\idxpair=0$ otherwise. Inequality~\eqref{eq:median_odd} implies that $\compareofpair_\idxpair + \compareofpair'_\idxpair = 1$ for any $\idxpair\in [\numgradershalf]$. Define $\compareofpair = \sum_{\idxpair=1}^{\numgradershalf} \compareofpair_\idxpair$ as the count of pairwise wins for item $1$ under assignment $\assignmentobs$, and define $\compareofpair'$ similarly. Then we have
\begin{align}
    \compareofpair + \compareofpair' = \numgradershalffrac.\label{eq:sign_estimator_distinct_output}
\end{align} 

The sign estimator outputs the item with more pairwise wins. That is, the sign estimator outputs item 1 under assignment $\assignmentobs$ if $\compareofpair > \numgraders/4$, outputs item 1 under assignment $\assignmentobs'$ if $\compareofpair' > \numgraders/4$, and outputs one of the two items uniformly at random if $\compareofpair =\numgraders/4$ or $\compareofpair' = \numgraders/4$. When $\compareofpair =\compareofpair' = \numgradersquarter$, then under either assignment, the sign estimator has a tie, and hence outputs one of the two items uniformly at random. The probability of error of the sign estimator is $0.5$. Otherwise, we have $\compareofpair \ne \numgradersquarter$. By~\eqref{eq:sign_estimator_distinct_output}, we have either $\compareofpair > \numgraders / 4 > \compareofpair'$ or $\compareofpair' > \numgraders / 4 > \compareofpair$. The sign estimator gives different outputs under the two assignments, out of which one and only one output is correct. The probability of error of the sign estimator is $0.5$.

\subsection{Proof of Theorem~\ref{thm:abtest_ours}}
Recall that a subset of $\numgradershalf$ \graders, drawn uniformly at random without replacement, are assigned to item $1$, and the remaining $\numgradershalf$ \graders are assigned to item $2$. We provide an alternative and equivalent view of the assignment as the following two steps:
\begin{enumeratex}[(1)]
    \item  We sample two \graders, uniformly at random without replacement, as the first pair of \graders for the two items, and call them $\{\idxgrader_1, \idxgrader_1'\}$. Then sample two \graders, uniformly at random without replacement, from the remaining $(\numgraders-2)$ \graders as the second pair of \graders for the two items, and call them $\{\idxgrader_2, \idxgrader'_2\}$. Continue until all $\numgraders$ \graders are exhausted, and call the subsequent pairs of \graders $\{\idxgrader_3, \idxgrader'_3\}, \ldots, \{\idxgrader_{\numgradershalf}, \idxgrader'_{\numgradershalf}\}$.
    \item  Within each pair, assign the pair of \graders to the two items uniformly at random. That is, for each $\idxpair \in [\numgradershalf]$, assign \grader $\idxgrader_\idxpair$ to one of the two items uniformly at random, and assign \grader $\idxgrader_\idxpair'$ to the remaining item. The assignments are independent across pairs.
\end{enumeratex}

Consider any arbitrary values of items $\val_1, \val_2\in \reals$. Given any arbitrary realization of Step (1) of the assignment procedure described above, we apply Theorem~\ref{thm:canonical_ours} and show that on each pair of \graders, the canonical estimator gives the correct output with probability strictly greater than $0.5$. Then we show that combining the $\numgradershalf$ pairs by majority-voting yields probability of success strictly greater than $0.5$.

Denote $\probpairsuccess(\val_1, \val_2, \{\func, \func'\})$ as the probability that our canonical estimator in Eq.~\eqref{eq:canonical_estimator} gives the correct output comparing items of values $\val_1, \val_2$ under \grader calibration functions $\func, \func'$. In Step (2) of the assignment procedure described above, for any $\idxpair\in [\numgradershalf]$, consider the $\idxpair^{th}$ pair of \graders, $\{\idxgrader_\idxpair, \idxgrader_\idxpair'\}$. Suppose that the calibration functions of these two \graders are denoted as $\{\func, \func'\}$. By Theorem~\ref{thm:canonical_ours}, since the two \graders are assigned to the two items uniformly at random, we have
\begin{align}\label{eq:abtest_ours_pair_success}
    \probpairsuccess\left(\val_1, \val_2, \{\func, \func'\}\right) > \frac{1}{2}\qquad \text{for all permissible } \func, \func'.
\end{align}

Let $\probpairsuccessmin$ denote the probability of success of our canonical estimator when run on the worst pair of calibration functions among all pairs of \graders
\begin{align*}
    \probpairsuccessmin = \min_{\func, \func' \in \{\func_1, \ldots, \func_\numgraders\}} \probpairsuccess(\val_1, \val_2, \{\func, \func'\}) \stackrel{\stepone}{>} \frac{1}{2},
\end{align*}
where inequality \stepone is true because of Eq.~\eqref{eq:abtest_ours_pair_success}, and because the number of \graders $\numgraders$ is finite.

Now assume that we are given any arbitrary realization of Step (1) of the assignment. For each $\idxpair\in [\numgradershalf]$, define $\resultofpair_\idxpair\in \{0, 1\}$ as the indicator variable of the correctness of our canonical estimator on the $\idxpair^{th}$ pair of \scores. We set $\resultofpair_\idxpair = 1$ if the canonical estimator gives the correct output on the $\idxpair^{th}$ pair, and $0$ otherwise. Then $\resultofpair_\idxpair$ is a Bernoulli random variable with mean $\probpairsuccess(\val_1, \val_2, \{\func_{\idxgrader_\idxpair}, \func_{\idxgrader_\idxpair'}\}) \ge \probpairsuccessmin$. Moreover, since Step (2) of the assignment is performed independently across all pairs, the variables $\{\resultofpair_\idxgrader\}_{\idxgrader=1}^\numgradershalfk$ are independent given the item values and Step (1) of the assignment.

Let $\sumofpairs = \sum_{\idxgrader=1}^{\numgradershalf} \resultofpair_\idxgrader$ be the number of pairs for which the canonical estimator $\quantitycardourscanonical$ gives the correct output. Define a binomial random variable $\varbinom$ with $\numgradershalfk$ trials and the success probability parameter $\probpairsuccessmin$.
Then the random variable $\sumofpairs$ stochastically dominates the random variable $\varbinom$. Recall that our estimator breaks ties uniformly at random. The probability of success of our estimator with the majority-voting procedure is then bounded as
\begin{align*}
     \Prob[\sumofpairs > \frac{\numgradershalfk}{2}] + \frac{1}{2}\Prob[\sumofpairs = \frac{\numgradershalfk}{2}] = & \frac{1}{2}\left(\Prob[\sumofpairs > \frac{\numgradershalfk}{2}] + \Prob[\sumofpairs\ge \frac{\numgradershalfk}{2}]\right)\\
    \ge & \frac{1}{2}\left(\Prob[\varbinom > \frac{\numgradershalfk}{2}] + \Prob[\varbinom\ge \frac{\numgradershalfk}{2}]\right)\\
    = & \Prob[\varbinom > \frac{\numgradershalfk}{2}] + \frac{1}{2}\Prob[\varbinom = \frac{\numgradershalfk}{2}]\\
    \stackrel{\stepone}{>} & \frac{1}{2},
\end{align*}
where inequality \stepone is true because the success probability parameter $\probpairsuccessmin$ of the binomial variable is strictly greater than $\frac{1}{2}$.

We complete the proof that the probability of success of our estimator is strictly greater than $0.5$ uniformly on any item values $\val_1, \val_2$ and any permissible calibration functions $\{\func_\idxgrader\}_{\idxgrader=1}^\numgraders$.

\subsection{Proof of Theorem~\ref{thm:sort_ours_uniform_prior}}
We first provide a high-level description of the proof. We call a pair of items ``flippable'', if Algorithm~\ref{alg:sort_cardinal} uses the canonical estimator to decide the relative ordering of this pair (that is, the if-condition in Line~\ref{line:if_flippable} in Algorithm~\ref{alg:sort_cardinal} is true). Note that a ``flippable'' pair may or may not be flipped by the algorithm, as the outcome depends on the output of the canonical estimator. In Theorem~\ref{thm:canonical_ours}, we show that our canonical estimator $\quantitycardourscanonical$ predicts the relative ordering of a pair of items correctly with probability strictly greater than $0.5$. The main idea of the proof is to apply Theorem~\ref{thm:canonical_ours} to each flippable pair. Then we show that an improvement on the probability of correctness on these flippable pairs translates to an improvement on the probability of success of exact recovery.

Theorem~\ref{thm:canonical_ours} requires that within each pair, the two \graders are assigned the two items uniformly at random. To be able to apply this theorem, we separate the different sources of randomness in the joint procedure of the assignment and the algorithm. We derive an equivalent algorithm by re-ordering the steps of Algorithm~\ref{alg:sort_cardinal}, so that in this equivalent algorithm, given any flippable pair of items and two \graders evaluating this pair, the last sources of randomness comes from the random assignment of the two \graders to the two items within this pair.

We introduce some additional notation for our re-ordered algorithm. Recall the notation of $\assignment = (\itemsofgrader_1, \ldots, \itemsofgrader_\numgraders)$ for the \grader assignment, where $\itemsofgrader
_\idxgrader$ is a pair of items assigned to \grader $\idxgrader$ for each $\idxgrader\in [\numgraders]$. Denote $\itemsofgraderunorderall = \{\itemsofgraderunorder_\idxgrader\}_{\idxgrader=1}^\numgraders$ as the same $\numgraders$ pairs of items, but the \grader assigned to each pair $\itemsofgraderunorder_\idxgrader$ is unspecified. Now we present an equivalent joint procedure of the assignment and the cardinal estimator $\quantitycardourssort$ in Algorithm~\ref{alg:sort_cardinal_alternative}. In what follows, we provide a high-level summary of Algorithm~\ref{alg:sort_cardinal_alternative}:

\begin{algorithm}[t!]
\DontPrintSemicolon
    Sample pairwise comparisons $\itemsofgraderunorderall = \{\itemsofgraderunorder_\idxgrader\}_{\idxgrader=1}^\numgraders$ uniformly at random from all ${\numitems \choose 2}$ pairs. Obtain the ordinal comparisons $\infoordall$.\label{line_alt:assignment_sample_pairs}\;
    Compute a topological ordering $\quantityest$ on the graph $\graph(\infoordall)$, with ties broken in order of the indices of the items.\label{line_alt:initial_guess}\;
    $\position \leftarrow 1$.\label{line_alt:estimator_store_start}\;
    $\itemsofgraderunorderallavail \leftarrow \itemsofgraderunorderall$.\;
    $\varpositions \leftarrow [\,]$.\;
    $\vargraders  \leftarrow [\,]$.\;
    \While{$\position < \numitems$}{
        Let $\quantityest_\text{flip}$ be the ranking obtained by flipping the positions of the $\position^{th}$ and the $(\position+1)^{th}$ items in $\quantityest$.\;
        \eIf{$\quantityest_\text{flip}$ is a topological ordering on $\graph( \infoordall)$, and both the $\position^{th}$ and $(\position+1)^{th}$ items are each included in at least one pairwise comparison in $\itemsofgraderunorderallavail$\label{line_alt:if_flippable}}{
            From all of the pairwise comparisons in $\itemsofgraderunorderallavail$ including the $\position^{th}$ item, sample one uniformly at random and denote it as $\itemsofgraderunorder_{\position}$. Likewise denote $\itemsofgraderunorder_{\position + 1}$ as a randomly chosen pairwise comparison including the $(\position + 1)^{th}$ item from $\itemsofgraderunorderallavail$.\;
            Append $\position$ to \varpositions.\;
            Append the pair $[\itemsofgraderunorder_{\position}, \itemsofgraderunorder_{\position + 1}]$ to \vargraders.\;
            Remove $\itemsofgraderunorder_{\position}$ and $\itemsofgraderunorder_{\position + 1}$ from $\itemsofgraderunorderallavail$.\;
            $\position \leftarrow \position + 2$.\;
            }{
            $\position\leftarrow \position + 1$.\;
        } 
    } \label{line_alt:estimator_store_end}
    For each pair $[\itemsofgraderunorder_{\position}, \itemsofgraderunorder_{\position + 1}]$ in \vargraders, sample uniformly at random without replacement a pair of \graders $\{\idxgrader_\position, \idxgrader_{\position+1}\}$.\label{line_alt:assignment_assign_graders}\;
    \ForEach{$\position \in$\varpositions}{ \label{line_alt:estimator_flip_start}
        Assign \grader $\idxgrader_\position$ to one of the two pairs $\itemsofgraderunorder_{\position}$ or $\itemsofgraderunorder_{\position + 1}$, uniformly at random. Assign \grader  $\idxgrader_{\position+1}$ to the remaining pair. Obtain the \scores from these two \graders for their corresponding pair.\label{line_alt:assignment_within_pair}\;
        Denote $\info_{\quantityest(\position)}$ as the \score for the $\position^{th}$ item in $\itemsofgraderunorder_{\position}$. Likewise denote $\info_{\quantityest(\position+1)}$ as the \score for the $(\position+1)^{th}$ item in $\itemsofgraderunorder_{\position+1}$.\label{line_alt:estimator_flip_substart}\;
        \If{$\quantitycardourscanonical(\info_{\quantityest(\position)}, \info_{\quantityest(\position+1)})$ outputs $ {\quantityest(\position+1)}\greateritem {\quantityest(\position)}$\label{line_alt:canonical_call}}{
            Let $\quantityestflip$ be the ranking obtained by flipping the positions of the $\position^{th}$ and the $(\position+1)^{th}$ items in $\quantityest$.\;
            $\quantityest\leftarrow \quantityestflip$.\;
        } \label{line_alt:canonical_call_end}
    }
    Output $\quantityest$.\label{line_alt:estimator_flip_end}\;
    \caption{An equivalent joint procedure of the assignment $\assignment$, the evaluation $\infocardall$, and the execution of our cardinal ranking estimator  $\quantitycardourssort(\assignment, \infocardall)$ in Algorithm~\ref{alg:sort_cardinal}.}\label{alg:sort_cardinal_alternative}
\end{algorithm}    

\begin{enumeratex}[(1)]
    \item \emph{Line~\ref{line_alt:assignment_sample_pairs}-\ref{line_alt:initial_guess}:} We sample $\numgraders$ pairwise comparisons of the items, drawn uniformly at random without replacement from the ${\numitems \choose 2}$ pairs. Obtain an initial estimate $\quantityest$ of the ranking, by computing a topological ordering on the graph $\graph(\infoordall)$. \label{enum:assignment_sample_pairs}
    \item \emph{Line~\ref{line_alt:estimator_store_start}-\ref{line_alt:estimator_store_end}: } Store the positions of all flippable pairs (if any) determined by Algorithm~\ref{alg:sort_cardinal}. If an item is included in some flippable pair, then this item is matched to a distinct pairwise comparison in $\itemsofgraderunorderall$. Store the matching between the items in flippable pairs and the pairwise comparisons.\label{enum:estimator_store}
    \item \emph{Line~\ref{line_alt:assignment_assign_graders}: } For the two pairwise comparisons associated with each pair of flippable items, sample two \graders uniformly at random without replacement to \grade the two comparisons.\label{enum:assignment_assign_graders}
    \stepcounter{enumi}
        \begin{enumerate}[(4a), leftmargin=0.06in]
            \item \emph{Line~\ref{line_alt:estimator_flip_start}-\ref{line_alt:assignment_within_pair}: } Within each flippable pair, assign the two \graders to the two items uniformly at random.\label{enum:assignment_within_pair}
            \item \emph{Line~\ref{line_alt:estimator_flip_substart}-\ref{line_alt:estimator_flip_end}: } Run the canonical estimator on each flippable pair, and flip the pair if the canonical estimator decides to do so (Line~\ref{line_alt:canonical_call}-\ref{line_alt:canonical_call_end}). After all flippable pairs are examined, output the final ranking $\quantityest$.\label{enum:estimator:flip}
        \end{enumerate}
\end{enumeratex}

We now briefly discuss the equivalence of Algorithm~\ref{alg:sort_cardinal_alternative} to Algorithm~\ref{alg:sort_cardinal}. We first discuss the equivalence of the assignment procedures in the two algorithms, and then the estimation aspect in the next paragraph. The assignment consists of Steps~\ref{enum:assignment_sample_pairs},~\ref{enum:assignment_assign_graders} and~\ref{enum:assignment_within_pair}. Recall that the assignment in Algorithm~\ref{alg:sort_cardinal} samples $\numgraders$ pairwise comparisons, uniformly at random without replacement, to assign to the $\numgraders$ \graders. In Algorithm~\ref{alg:sort_cardinal_alternative}, this assignment is decomposed into the choice of pairwise comparisons, the choice of a pair of \graders to two pairwise comparisons in each flippable pair, and the assignment within each flippable pair, corresponding to Steps~\ref{enum:assignment_sample_pairs},~\ref{enum:assignment_assign_graders} and~\ref{enum:assignment_within_pair}, respectively. Note that only the selected pairwise comparison for each item within some flippable pair is used for Algorithm~\ref{alg:sort_cardinal_alternative}, so we do not need to specify the assignment of the \graders for the rest of the comparisons. This re-ordering of the assignment is equivalent to Algorithm~\ref{alg:sort_cardinal}.

The cardinal ranking estimator consists of the rest of the steps, namely Steps~\ref{enum:estimator_store} and~\ref{enum:estimator:flip}. In the original presentation of the estimator in Algorithm~\ref{alg:sort_cardinal}, the estimator scans through the items, identifies flippable pairs, calls the canonical estimator on each flippable pair, and flips the pairs accordingly. Note that the identification of flippable pairs does not need the assignment of \graders or the \scores from the \graders, so Algorithm~\ref{alg:sort_cardinal_alternative} first scans through the items and identifies all flippable pairs, without using the choice of the \graders in the assignment or using the \scores from the \graders. Then Algorithm~\ref{alg:sort_cardinal_alternative} calls the canonical estimator on each flippable pair once the choice of the \graders and the \scores are determined, and flips each pair based on the corresponding output from the canonical estimator. Note that when checking for a flippable pair (the if-condition in Line~\ref{line:if_flippable} in Algorithm~\ref{alg:sort_cardinal} and Line~\ref{line_alt:if_flippable} in Algorithm~\ref{alg:sort_cardinal_alternative}), Algorithm~\ref{alg:sort_cardinal} checks whether the flipped ranking $\quantityestflip$ is a topological ordering, where the previous flippable pairs in $\quantityestflip$ may have already been flipped. In Algorithm~\ref{alg:sort_cardinal_alternative}, the previous flippable pairs are identified but are not flipped. However, whether the flipped ranking $\quantityestflip$ is a topological ordering is independent of whether the previous flippable pairs in $\quantityestflip$ are flipped. Hence, the identification of the flippalbe pairs is equivalent in the two algorithms. The re-ordering of the steps of the cardinal estimator $\quantitycardourssort$ is valid.

Having now established the equivalence of Algorithm~\ref{alg:sort_cardinal_alternative} to Algorithm~\ref{alg:sort_cardinal}, we now prove Theorem~\ref{thm:sort_ours_uniform_prior} with respect to Algorithm~\ref{alg:sort_cardinal_alternative}. Let us denote $\quantitycardourssortequiv$ as the cardinal estimator in Algorithm~\ref{alg:sort_cardinal_alternative}. Denote $\topo(\infoordall)$ as the set of all topological orderings on the directed graph $\graph(\infoordall)$ induced by the set of ordinal observations $\infoordall$. We denote a random variable $\numorders(\infoordall) \defn \abs{\topo(\infoordall)}$ as the number of such topological orderings. Note that the definition of flippable pairs carries over from Algorithm~\ref{alg:sort_cardinal} to Algorithm~\ref{alg:sort_cardinal_alternative}. We denote a random variable $\numflips$ as the number of flippable pairs in Algorithm~\ref{alg:sort_cardinal_alternative}.

Let us first consider the probability of success of the ordinal estimator. The following lemma describes the posterior distribution of the true ranking conditioned on the set of ordinal observations $\infoordall$. Using this posterior distribution, the optimal ordinal estimators and their probability of success are derived.

\begin{lemma}\label{lem:ranking}
(a) Given any possible set of ordinal observations $\infoordallobs$, the posterior distribution of the true ranking $\quantitygt$ is uniformly distributed over the $\numorders(\infoordallobs)$ topological orderings:
\begin{align}
    \Prob(\quantitygt =\quantity\given \infoordall = \infoordallobs) = \begin{cases}
        \frac{1}{\numorders(\infoordallobs)} & \text{if } \quantity \in \topo(\infoordallobs)\\
        0 & \text{otherwise}.\label{eq:posterior}
    \end{cases}
\end{align}

(b) Any ordinal estimator $\quantityordsortopt$ is optimal for the 0-1 loss, if and only if given any set of ordinal observations $\infoordallobs$, the output of this ordinal estimator belongs to the $\numorders(\infoordallobs)$ topological orderings with probability $1$, that is, if and only if
\begin{align}
    \Prob(\quantityordsortopt(\infoordallobs)\in \topo(\infoordallobs) \given \infoordall = \infoordallobs) = 1 \qquad \text{for all possible set of ordinal observations }\infoordallobs.\label{eq:ranking_ordinal_optimality_condition}
\end{align}

Moreover, conditioned on the set of ordinal observations $\infoordallobs$, the probability of success of any optimal ordinal estimator $\quantityordsortopt$ is 
\begin{align}
    \Prob(\quantityordsortopt = \quantitygt \given \infoordall = \infoordallobs) = \frac{1}{\numorders(\infoordallobs)}.\label{eq:ranking_success_optimal_ordinal}
\end{align}
\end{lemma}
See Section~\ref{app:proof_lemma_ranking} for the proof of the lemma.

Now consider the probability of success of our cardinal estimator $\quantitycardourssortequiv$ from Algorithm~\ref{alg:sort_cardinal_alternative}. We write the probability of success of our cardinal estimator as
\begin{align}
    \Prob(\quantitycardourssortequiv=\quantitygt) = & \sum_\infoordallobs\sum_\numflipsobs \Prob(\quantitycardourssort = \quantitygt \given \infoordall=\infoordallobs, \numflips = \numflipsobs)\Prob(\infoordall=\infoordallobs, \numflips = \numflipsobs),\label{eq:ranking_cardinal_success}
\end{align}
where $\infoordallobs$ is summed over all possible sets of ordinal observations, and $\numflipsobs$ is summed from $0$ to $\floor{\numitems/2}$.

We consider each term $\Prob(\quantitycardourssortequiv = \quantitygt \given \infoordall=\infoordallobs, \numflips = \numflipsobs)$ separately for each $\infoordallobs$ and $\numflipsobs$. We prove that for any $\infoordallobs$ and any $\numflipsobs$, the probability of success of our cardinal estimator is greater than or equal to the probability of success of any optimal ordinal estimator $\quantityordsortopt$. We also show that the probability of success is strictly greater for some $\infoordallobs$ and $\numflipsobs$. We discuss the following two cases separately, depending on the number of flippable pairs.

\noindent\textbf{Case 1:} $\numflipsobs = 0$.

We have the number of flippable pairs $\numflips = 0$ either if there is a unique topological ordering on the graph $\graph(\infoordall)$, or if in each pair of adjacent items that can be flipped without violating pairwise comparisons, at least one item in this pair does not have any \score. Note that these two conditions are fully determined by the set of ordinal observations. Hence, conditioned on the set of ordinal observations $\infoordall = \infoordallobs$, the event of $\numflips = 0$ is fully determined, and  is independent of everything else given $\infoordall$.

The initial estimated ranking of the cardinal estimator is a topological ordering (Line~\ref{line_alt:initial_guess} of Algorithm~\ref{alg:sort_cardinal_alternative}). Since there is no flippable pair, the cardinal estimator simply outputs this topological ordering. For any set of ordinal observations $\infoordallobs$ such that $\Prob(\infoordall = \infoordallobs, \numflips = 0) > 0$, we have
\begin{align}
    \Prob(\quantitycardourssortequiv = \quantitygt \given \infoordall=\infoordallobs, \numflips = 0) \stackrel{\stepone}{=} & \Prob(\quantitycardourssortequiv = \quantitygt \given \infoordall=\infoordallobs)\nonumber\\
    \stackrel{\steptwo}{=} & \Prob(\quantityordsortopt = \quantitygt \given \infoordall = \infoordallobs),\label{eq:ranking_cardinal_flippable_pair_zero}
\end{align}
where $\quantityordsortopt$ denotes any optimal ordinal estimator. Here in~\eqref{eq:ranking_cardinal_flippable_pair_zero}, equality \stepone is true because the event $\numflips = 0$ is fully determined by $\infoordall$, and equality \steptwo is true because this cardinal estimator that simply outputs a topological ordering is equivalent to an ordinal estimator that outputs the same topological ordering. From~\eqref{eq:ranking_ordinal_optimality_condition}, this ordinal estimator is one optimal ordinal estimator.

\noindent\textbf{Case 2:} $\numflipsobs > 0$.

In this case, Algorithm~\ref{alg:sort_cardinal_alternative} identifies at least one flippable pair. The probability of success of our cardinal estimator is
\begin{align}
    \Prob(\quantitycardourssortequiv = \quantitygt \given \infoordall = \infoordallobs, \numflips = \numflipsobs) = & \sum_{\quantity\in \permall}\Prob(\quantitycardourssortequiv = \quantity \given \quantitygt = \quantity, \infoordall = \infoordallobs, \numflips = \numflipsobs) \Prob(\quantitygt = \quantity \given \infoordall = \infoordallobs, \numflips = \numflipsobs) \nonumber\\
    \stackrel{\stepone}{=} & \sum_{\quantity\in \permall}\Prob(\quantitycardourssortequiv = \quantity \given \quantitygt = \quantity, \infoordall = \infoordallobs, \numflips = \numflipsobs) \Prob(\quantitygt = \quantity \given \infoordall = \infoordallobs) \nonumber\\
    \stackrel{\steptwo}{=} & \frac{1}{\numorders(\infoordallobs)}\sum_{\quantity\in \topo(\infoordallobs)}\Prob(\quantitycardourssortequiv = \quantity \given \quantitygt = \quantity, \infoordall = \infoordallobs, \numflips = \numflipsobs),\label{eq:ranking_cardinal_flippable_pair_at_least_one_intermediate}
\end{align}
where equality \stepone is true because $\numflips$ is independent of $\quantitygt$ conditioned on $\infoordall$. Equality \steptwo is true by plugging in~\eqref{eq:posterior}.

In Algorithm~\ref{alg:sort_cardinal_alternative}, Lines~\ref{line_alt:assignment_sample_pairs}-\ref{line_alt:assignment_assign_graders} fully determine the number of the flippable pairs, their positions, and the two \graders \grading each flippable pair. In Lines~\ref{line_alt:estimator_flip_start}-\ref{line_alt:estimator_flip_end}, within each flippable pair, the algorithm first assigns uniformly at random one \grader to one item and the remaining \grader to the remaining item, and then calls the canonical estimator to output the relative ordering of this pair. Conditioned on the randomness in Lines~\ref{line_alt:assignment_sample_pairs}-\ref{line_alt:assignment_assign_graders} of Algorithm~\ref{alg:sort_cardinal_alternative}, we now apply Theorem~\ref{thm:canonical_ours} to each flippable pair. Since the \grader assignment within each flippable pair (Line~\ref{line_alt:assignment_within_pair}) is uniformly at random, by Thoerem~\ref{thm:canonical_ours}, the probability that the canonical estimator outputs the correct relative ordering of each flippable pair is strictly greater than $\frac{1}{2}$. Since the assignment within each flippable pair is independent across pairs, the probability that the canonical estimator outputs the correct relative ordering of all $\numflipsobs$ flippable pairs is strictly greater than $\frac{1}{2^\numflipsobs}$.

Recall that the initial estimated ranking of our cardinal estimator is a topological ordering. Consider all total rankings that are identical to this initial ranking, except for (possibly) the relative ordering of the $\numflipsobs$ flippable pairs. Since the items in the flippable pairs are disjoint, there are $2^\numflipsobs$ such total rankings. By definition, a pair is flippable only if the total ranking after this pair is flipped is still a topological ordering. Hence, all these $2^\numflipsobs$ total rankings are topological orderings on the graph $\graph(\infoordall)$. In~\eqref{eq:ranking_cardinal_flippable_pair_at_least_one_intermediate}, the summation of $\quantity$ is over all topological orderings. In particular, this summation includes these $2^\numflipsobs$ total rankings. On each of these $2^\numflipsobs$ total rankings, the output of our ranking estimator $\quantitycardourssortequiv$ is correct if and only if the canonical estimator outputs the correct relative orderings for the $\numflipsobs$ flippable pairs, which happens with probability strictly greater than $\frac{1}{2^\numflipsobs}$. Hence, we bound~\eqref{eq:ranking_cardinal_flippable_pair_at_least_one_intermediate} as

\begin{align}
    \Prob(\quantitycardourssortequiv = \quantitygt \given \infoordall = \infoordallobs, \numflips = \numflipsobs) >  \frac{1}{\numorders(\infoordallobs)} \cdot 2^\numflipsobs \cdot \frac{1}{2^\numflipsobs} = & \frac{1}{\numorders(\infoordallobs)}\nonumber\\
    \stackrel{\stepone}{=} &  \Prob(\quantityordsortopt = \quantitygt \given \infoordall = \infoordallobs),\label{eq:ranking_cardinal_flippable_pair_at_least_one}
\end{align}
where $\quantityordsortopt$ denotes any optimal ordinal estimator. Equality \stepone is true because of~\eqref{eq:ranking_success_optimal_ordinal} in Lemma~\ref{lem:ranking}.

Plugging~\eqref{eq:ranking_cardinal_flippable_pair_zero} and~\eqref{eq:ranking_cardinal_flippable_pair_at_least_one} into~\eqref{eq:ranking_cardinal_success}, we have
\begin{align}
    \Prob(\quantitycardourssortequiv = \quantitygt) \ge \Prob(\quantityordsortopt = \quantitygt) \qquad \text{for any optimal ordinal estimator } \quantityordsortopt,\label{eq:ranking_cardinal_uniform_dominance_equal_to}
\end{align}
and a strict inequality holds in~\eqref{eq:ranking_cardinal_uniform_dominance_equal_to} if there exists some $\infoordallobs$ and some $\numflipsobs > 0$, such that
\begin{align}
    \Prob(\infoordall = \infoordallobs, \numflips = \numflipsobs) > 0.\label{eq:ranking_flippable_nonzero_probability}
\end{align}

It remains to find some $\infoordallobs$ and some $\numflipsobs > 0$ such that~\eqref{eq:ranking_flippable_nonzero_probability} is true. We construct such $\infoordallobs$ and $\numflipsobs > 0$ as follows. Consider the true ranking $1\greateritem2 \greateritem\cdots \greateritem\numitems$, which happens with a strictly positive probability as the prior distribution of the true ranking is uniform. Conditioned on this true ranking, consider the event that the sampled pairwise comparisons in $\itemsofgraderunorderall$ do not include a direct comparison between items $1$ and $2$, but both item $1$ and item $2$ have at least one \score each (from comparisons with at least one of the remaining $(\numitems - 2)$ items). Recall that the number of pairs satisfies $1 < \numgraders < {\numitems \choose 2}$, so such a set $\itemsofgraderunorderall$ of pairwise comparisons arises with a strictly positive probability. Let $\infoordallobs$ be the set of ordinal observations derived from the true ranking and the set $\itemsofgraderunorderall$ of pairwise comparisons described as above. With this $\infoordallobs$, item $1$ and item $2$ are the first two items in the initial ranking of the topological ordering, they can be flipped, and they both have some \scores. Hence, item $1$ and item $2$ form a flippable pair, and we have $\numflips > 0$.  Hence, with this construction of $\infoordallobs$, we have
\begin{align*}
    \sum_{\numflipsobs = 1}^{\floor{\numitems / 2}} \Prob(\infoordall = \infoordallobs, \numflips = \numflipsobs) > 0.
\end{align*}
Thus there exists some $\numflipsobs > 0$ such that $\Prob(\infoordall = \infoordallobs, \numflips = \numflipsobs) > 0$. Hence, Equation~\eqref{eq:ranking_flippable_nonzero_probability} is true, implying the strictly inequality in~\eqref{eq:ranking_cardinal_uniform_dominance_equal_to}. Consequently, the probability of success of our cardinal ranking estimator $\quantitycardourssortequiv$ is strictly uniformly greater than the probability of success of any optimal ordinal estimator. By the equivalence of Algorithm~\ref{alg:sort_cardinal_alternative} and Algorithm~\ref{alg:sort_cardinal}, this result also holds for the original cardinal estimator $\quantitycardourssort$, completing the proof.

\subsubsection{Proof of Lemma~\ref{lem:ranking}}\label{app:proof_lemma_ranking}

We first prove part (a) of the lemma. By Bayes rule, for any ranking $\quantity\in \permall$ and any possible set of ordinal observations $\infoordallobs$, we have
\begin{align}
    \Prob(\quantitygt =\quantity \given \infoordall = \infoordallobs) = \frac{\Prob( \infoordall = \infoordallobs\given \quantitygt = \quantity)\Prob(\quantitygt=\quantity)}{\Prob(\infoordall=\infoordallobs)}.\label{eq:lem_bayes_rule}
\end{align}

Given the set of ordinal observations $\infoordallobs$, the denominator in~\eqref{eq:lem_bayes_rule} is independent of $\quantity$. Since the prior of the true ranking is uniform, in the numerator we have $\Prob(\quantitygt = \quantity) =\frac{1}{\numitems!}$, independent of $\quantity$. Now it remains to consider the term $\Prob(\infoordall = \infoordallobs \given \quantitygt = \quantity)$ in the numerator. Recall the notation of the random variable $\itemsofgraderunorderall$ as the set of pairwise comparisons in the ordinal observations (but $\itemsofgraderunorderall$ does not include the results of the relative orderings of these pairs). We write the term $\Prob(\infoordall = \infoordallobs \given \quantitygt = \quantity)$ as
\begin{align}
    \Prob(\infoordall = \infoordallobs \given \quantitygt = \quantity) = & \sum_\itemsofgraderunorderallobs \Prob(\infoordall = \infoordallobs \given \itemsofgraderunorderall = \itemsofgraderunorderallobs, \quantitygt = \quantity)\Prob(\itemsofgraderunorderall = \itemsofgraderunorderallobs \given \quantitygt = \quantity)\nonumber\\
    \stackrel{\stepone}{=} & \sum_\itemsofgraderunorderallobs \Prob(\infoordall = \infoordallobs \given \itemsofgraderunorderall = \itemsofgraderunorderallobs, \quantitygt = \quantity)\Prob(\itemsofgraderunorderall = \itemsofgraderunorderallobs),\label{eq:lem_bayes_rule_conditional}
\end{align}
where $\itemsofgraderunorderallobs$ is summed over all possible sets of $\numgraders$ pairwise comparisons. Equality \stepone is true because the sampling of the set of pairwise comparisons $\itemsofgraderunorderall$ is independent of the true ranking $\quantitygt$.

Recall that the set of ordinal observations $\infoordallobs$ includes the pairwise comparisons and results of the relative orderings of these pairwise comparisons, whereas $\itemsofgraderunorderallobs$ only includes the pairwise comparisons themselves, so $\infoordallobs$ fully determines $\itemsofgraderunorderallobs$. For this term to be non-zero, the set of pairwise comparisons indicated by $\infoordallobs$ and the set of pairwise comparisons indicated by $\itemsofgraderunorderallobs$ need to be identical. Hence, there is only one $\itemsofgraderunorderallobs$ in the summation of~\eqref{eq:lem_bayes_rule_conditional} consistent with $\infoordallobs$, and we denote $\widetilde{\itemsofgraderunorderallobs}$ as the set of pairs consistent with $\infoordallobs$. Then~\eqref{eq:lem_bayes_rule_conditional} reduces to
\begin{align}
        \Prob(\infoordall = \infoordallobs \given \quantitygt = \quantity) = & \Prob(\infoordall = \infoordallobs \given \itemsofgraderunorderall = \widetilde{\itemsofgraderunorderallobs}, \quantitygt = \quantity)\Prob(\itemsofgraderunorderall = \widetilde{\itemsofgraderunorderallobs}),\label{eq:lem_bayes_rule_conditional_intermediate}
\end{align}

In~\eqref{eq:lem_bayes_rule_conditional_intermediate}, the second term $\Prob(\itemsofgraderunorderall = \widetilde{\itemsofgraderunorderallobs})$ is independent of $\quantity$. Now consider the first term $\Prob(\infoordall = \infoordallobs \given \itemsofgraderunorderall = \widetilde{\itemsofgraderunorderallobs}, \quantitygt = \quantity)$. If $\quantity$ is a topological ordering on $\graph(\infoordallobs)$, then by definition, the relative orderings on the pairs $\widetilde{\itemsofgraderunorderallobs}$ induced by the ranking $\quantity$ is the set of ordinal observations $\infoordallobs$. If $\quantity$ is not a topological ordering, then by definition, the relative orderings induced by the ranking $\quantity$ violates at least one relative ordering in $\infoordallobs$. Hence,

\begin{align}
    \Prob(\infoordall = \infoordallobs \given \itemsofgraderunorderall = \widetilde{\itemsofgraderunorderallobs}, \quantitygt = \quantity) = 
    \begin{cases}
        1 & \text{if } \quantity \in \topo(\infoordallobs)\\
        0 & \text{otherwise}.
    \end{cases}\label{eq:lem_bayes_rule_topological_ordering}
\end{align}

Combining the law of total probability with~\eqref{eq:lem_bayes_rule},~\eqref{eq:lem_bayes_rule_conditional_intermediate} and~\eqref{eq:lem_bayes_rule_topological_ordering}, the posterior distribution of the true ranking is
\begin{align}\label{eq:lemma_posterior}
    \Prob(\quantitygt = \quantity \given \infoordall=\infoordallobs) = \begin{cases}
        \frac{1}{\numorders(\infoordallobs)} & \text{if }\quantity \in \topo(\infoordallobs)\\
        0 & \text{otherwise.}
    \end{cases}
\end{align}

Conditioned on the set of ordinal observations $\infoordallobs$, the posterior distribution of the true ranking is uniform over all topological ordering on the graph $\graph(\infoordallobs)$. This completes the proof for part (a) of the lemma.\\

For part (b) of the lemma, we condition on any possible set of ordinal observations $\infoordallobs$. On the input $\infoordallobs$, the probability of success of any (possibly-randomized) ordinal estimator $\quantityordsort$ is:
\begin{align}
    \Prob(\quantityordsort(\infoordallobs) = \quantitygt \given \infoordall = \infoordallobs)
    = & \sum_{\quantity\in \permall} \Prob(\quantityordsort(\infoordallobs) = \quantity \given \quantitygt = \quantity, \infoordall = \infoordallobs) \Prob(\quantitygt = \quantity\given \infoordall = \infoordallobs)\nonumber\\
    \stackrel{\stepone}{=} &  \frac{1}{\numorders(\infoordallobs)}\sum_{\quantity\in \topo(\infoordallobs)} \Prob(\quantityordsort(\infoordallobs) = \quantity \given \quantitygt = \quantity, \infoordall= \infoordallobs)\nonumber\\
    \stackrel{\steptwo}{=} & \frac{1}{\numorders(\infoordallobs)} \sum_{\quantity\in \topo(\infoordallobs)}\Prob(\quantityordsort(\infoordallobs) = \quantity)\nonumber\\
    \stackrel{\stepthree}{\le} & \frac{1}{T(\infoordallobs)},\label{eq:lem_success_probability_inequality}
\end{align}
where equality \stepone is true by plugging in~\eqref{eq:lemma_posterior}. Equality \steptwo is true because the output of the ordinal estimator $\quantityordsort(\infoordallobs)$ on the input $\infoordallobs$ only depends on its internal randomness, and hence independent of the $\quantitygt$ and $\infoordall$. Inequality \stepthree is true by the law of total probability. In particular, the equality sign in \stepthree holds if and only if the output of the ordinal estimator is always a topological ordering consistent with $\infoordallobs$, that is, if and only if
\begin{align}
    \Prob(\quantityordsort(\infoordallobs)\in \topo(\infoordallobs) \given \infoordall= \infoordallobs) = 1. \label{eq:lemma_optimal_condition}
\end{align}
Taking an expectation over all possible ordinal observations $\infoordallobs$, we have
\begin{align}
    \Prob(\quantityordsort(\infoordall) = \quantitygt) = \sum_{\infoordallobs} \Prob(\quantityordsort(\infoordallobs) = \quantitygt\given \infoordall = \infoordallobs) \Prob(\infoordall=\infoordallobs).\label{eq:lem_success_probability_conditional}
\end{align}

Combining~\eqref{eq:lem_success_probability_conditional} with the condition~\eqref{eq:lemma_optimal_condition} for the equality sign in~\eqref{eq:lem_success_probability_inequality}, an ordinal estimator is optimal if and only if Eq.~\eqref{eq:lemma_optimal_condition} holds on all possible ordinal observations $\infoordallobs$ where $\Prob(\infoordall = \infoordallobs) > 0$. This completes the proof for part (b) of the lemma.

\section{Discussion}

Breaking the barrier of using only ranking data in the presence of arbitrary (and potentially adversarial) miscalibrations, we show that cardinal scores can yield strict and uniform improvements over rankings. This result uncovers a novel, strictly-superior point on the tradeoff between cardinal scores and ordinal rankings, and provides a new perspective on this eternally-debated tradeoff. Our estimator allows for easily plugging into a variety of algorithms, thereby yielding it a wide applicability.

The results of this paper lead to several useful open problems. First, while our estimators indeed uniformly outperform ordinal estimators, in the future, a more careful design in our estimators (e.g. how to choose the function $\funcmonotone$ in the canonical estimator, and how to design better estimators for A/B testing and ranking) may yield even better results. Second, it is of interest to obtain statistical bounds on the relative errors of the cardinal and ordinal estimators in terms of the unknown miscalibration functions. Third, a promising direction of future research is to design estimators that achieve the guarantees of our proposed estimator under arbitrary/adversarial miscalibrations while simultaneously being able to adapt and yield stronger guarantees when the calibration functions follow one of the popular simpler models of miscalibration (\`a la ``win-win'' models and estimators in prior work~\cite[Part I]{shah2017learning}~\cite{heckel2016active,shah2016permutation,shah2017stochastically,shah2018low,shah18simple}). Fourth, although we consider the rating scales as continuous intervals, it is not hard to see that our results extend to discrete scales (but with the strict inequality in Equation~\eqref{eq:uniformly_better} sometimes replaced by a non-strict inequality to account for ties). Using our results to guide the choice of the scale used for elicitation is an open problem of interest. And finally, practical applications such as peer-review do not suffer from the problem of miscalibration in isolation. It is a useful and challenging open problem to address miscalibration simultaneously with other issues such as noise~\cite{stelmakh2018forall}, subjectivity~\cite{noothigattu2018choosing}, strategic behavior~\cite{xu2018strategyproof} and others.


\section*{Acknowledgments}
This work was supported in parts by NSF grants CRII: CIF: 1755656 and CCF: 1763734. The authors thank Bryan Parno for very useful discussions on biases in conference peer review. The authors thank Pieter Abbeel for pointing out the related work on the two-envelope problem.

{
\small
\bibliographystyle{alpha}
\bibliography{references}
}

\appendix

\newcommand{\gap}{G}
\newcommand{\gaptrue}{g}
\newcommand{\gapnoisy}{\widetilde{\gap}}
\newcommand{\noisegap}{\Delta}
\newcommand{\funcmonotoneplus}{\funcmonotone_+}
\newcommand{\noisedist}{D}
\newcommand{\sample}{\sim}

\newcommand{\textkt}{\text{KT}}
\newcommand{\textspearman}{\text{SF}}
\newcommand{\quantityinvordsortkt}{\capord{\quantityinv}_{\textsort}}
\newcommand{\quantityinvcardourssortkt}{{\capours{\quantityinv}}_{\textsort\text{-metric}}^{\textours}}

\newcommand{\losskt}{\loss_\textkt}
\newcommand{\losssf}{\loss_\textspearman}

\newcommand{\topoidentical}{topologically-identical\xspace}

\newcommand{\ktcontribution}{\alpha}

\newcommand{\idxitemcontrib}{\ell}
\newcommand{\idxitemcontribmid}{\idxitemcontrib_{\text{mid}}}

\newcommand{\setcompare}{V}
\newcommand{\setcomparelarger}{\setcompare^+}
\newcommand{\setcomparesmaller}{\setcompare^-}

\newcommand{\varpositionskt}{\texttt{positions}}
\newcommand{\eventexist}{\event}

\newcommand{\tmpa}{a}
\newcommand{\tmpb}{b}

\newcommand{\quantityinit}{\quantityest_{\text{init}}}
\newcommand{\quantitycardourssortunifequiv}{\capours{\quantity}_{\textsort\text{-}\textunif}^{\text{eq}}}

~\\~\\
\noindent{\bf \Large Appendix: Extensions}\\

\noindent We now present three extensions of our problem setting and results from the main text. 

\section{Noisy data}\label{app:noisy}

In this section, we show that even when the \scores given by the reviewers are noisy, our estimator in~\eqref{eq:canonical_estimator} continues to strictly uniformly dominate random guessing in the canonical setting (Section~\ref{sec:canonical}). We focus on the canonical estimator.

In the noisy setting, when \grader $\idxgrader\in [\numgraders]$ \grades item $\idxitem\in [\numitems]$, the reported \score is \begin{align*} 
\func_\idxgrader(\val_\idxitem) + \noise_{\idxitem\idxgrader},
\end{align*}
where $\noise_{\idxitem\idxgrader}$ is a noise term. We assume that the noise terms $\{\noise_{\idxitem\idxgrader}\}_{\idxitem\in [\numitems], \idxgrader\in [\numgraders]}$ are drawn i.i.d. from an unknown distribution. In this setting of noisy reported \scores, we modify Definition~\ref{def:uniformly_better} of strict uniform dominance, and let the expectation include the randomness in the noise.

The following theorem establishes the strict uniform dominance in the noisy setting for the cardinal estimator $\quantitycardourscanonical$ in~\eqref{eq:canonical_estimator} (cf. Theorem~\ref{thm:canonical_ours} for the noiseless setting).

\begin{theorem}
\label{cor:noisy}
The canonical estimator $\quantitycardourscanonical$ strictly uniformly dominates the random-guessing estimator $\quantityordcanonical$ in the presence of noise.
\end{theorem}

Observe that this result is quite general, since the noise distribution can be arbitrary and unknown. The remainder of this section is devoted to the proof of Theorem~\ref{cor:noisy}. 

\subsection{Proof of Theorem~\ref{cor:noisy}}\label{app:proof_noisy}
The proof is a slight modification to the proof of Theorem~\ref{thm:canonical_ours}, so we only highlight the difference.

Recall that $\noise_{\idxitem\idxgrader}$ denotes the noise in the reported \score of \grader $\idxgrader\in \{1, 2\}$ for item $\idxitem\in \{1, 2\}$. In Eq.~\eqref{eq:canonical_probability_success_intermediate} from the proof of Theorem~\ref{thm:canonical_ours}, we replace all the noiseless terms $\func_\idxgrader(\val_\idxitem)$ by the noisy terms $\func_\idxgrader(\val_\idxitem) + \noise_{\idxitem\idxgrader}$ for each $\idxitem\in \{1, 2\}$ and $\idxgrader\in \{1, 2\}$. Using the fact that the noise terms are independent of everything else, and taking an expectation over all the noise terms, we have
\begin{align}
    \Prob(\quantitycardourscanonical = \quantitygt) = & \frac{1}{2}\Expect_{\noise_{11}, \noise_{12}, \noise_{21}, \noise_{22}} \left[1 + \funcmonotoneredefine((\func_1(\val_1) + \noise_{11}) - (\func_2(\val_2) + \noise_{22}))  - \funcmonotoneredefine( (\func_1(\val_2) + \noise_{21}) - (\func_2(\val_1) + \noise_{12}))\right]\nonumber\\
    = & \frac{1}{2}\Expect_{\noise_{11}, \noise_{12}, \noise_{21}, \noise_{22}} \left[1 + \funcmonotoneredefine(\func_1(\val_1) - \func_2(\val_2) + \noise_{11} - \noise_{22})  - \funcmonotoneredefine( \func_1(\val_2) - \func_2(\val_1) +\noise_{21} - \noise_{12})\right]\nonumber\\
    \stackrel{\stepone}{=} & \frac{1}{2}\Expect_{\noise_{1}, \noise_{2}}\left[1 + \funcmonotoneredefine(\func_1(\val_1) - \func_2(\val_2) + \noise_{1} - \noise_{2})  - \funcmonotoneredefine( \func_1(\val_2) - \func_2(\val_1) +\noise_{1} - \noise_{2})\right],\label{eq:noisy_probability_success_intermediate}
\end{align}
where step \stepone uses linearity of expectation with a change of variable names, as the noise terms $\noise_{ij}$ are i.i.d.

Without loss of generality, assume $\val_1 > \val_2$. Recall from the proof of Theorem~\ref{thm:canonical_ours} that $\func_1(\val_1) - \func_2(\val_2) > \func_1(\val_2) - \func_2(\val_1)$, and therefore we have the deterministic inequality
\begin{align*}
   \func_1(\val_1) - \func_2(\val_2) + \noise_1 - \noise_2 > \func_1(\val_2) - \func_2(\val_1) + \noise_1 - \noise_2,\quad \text{for any $\noise_1,\noise_2\in \reals$.}
\end{align*}

Using the monotonicity of $\funcmonotoneredefine$, we have
\begin{align}\label{eq:noisy_func_redefine_relation}
    \funcmonotoneredefine(\func_1(\val_1) - \func_2(\val_2) + \noise_1 - \noise_2)) > \funcmonotoneredefine(\func_1(\val_2) - \func_2(\val_1) + \noise_1 - \noise_2).
\end{align}

Taking an expectation over $\noise_1$ and $\noise_2$ in~\eqref{eq:noisy_func_redefine_relation} and combining with~\eqref{eq:noisy_probability_success_intermediate} gives \begin{align*}
    \Prob(\quantitycardourscanonical = \quantitygt) > \frac{1}{2}.
\end{align*}


\section{Ranking under Kendall-tau and Spearman's footrule distance}\label{app:other_metrics}

In addition to the 0-1 exact recovery loss considered in Theorem~\ref{thm:sort_ours_uniform_prior}, Kendall-tau distance and Spearman's footrule distance are also common metrics for ranking. Recall that a ranking of $\numitems$ items is defined by a function $\quantity:[\numitems]\rightarrow [\numitems]$, such that $\quantity(\position)$ is the index of the $\position^{th}$ ranked item for each $\position \in [\numitems]$. Equivalently, we can define a ranking by the function $\quantityinv:[\numitems] \rightarrow [\numitems]$, such that $\quantityinv(\idxitem)$ is the rank of each item $\idxitem\in [\numitems]$. With this notation, we have the relation $\quantityinv = \quantity^{-1}$. 

The Kendall-tau distance and the Spearman's footrule distance are usually defined in terms of the ranking $\quantityinv$. Hence for consistency with these definitions, throughout this section we focus on the rankings as defined by $\quantityinv$ (instead of $\quantity$ as done throughout the remainder of the paper). Kendall-tau distance and Spearman's footrule distance between any two rankings $\quantityinv_1$ and $\quantityinv_2$ of $\numitems$ items are defined as:
\begin{align*}
    \text{Kendall-tau distance: } \qquad & \losskt(\quantityinv_1, \quantityinv_2) = \sum_{\substack{ \idxitem\in [\numitems], \idxitem' \in [\numitems]:\\ \quantityinv_1(\idxitem) < \quantityinv_1(\idxitem')}} \indicatorevent{\quantityinv_2(\idxitem) > \quantityinv_2(\idxitem')}\\
    \text{Spearman's footrule distance: }\qquad & \losssf(\quantityinv_1, \quantityinv_2) = \sum_{\idxitem\in [\numitems]}\abs{\quantityinv_1(\idxitem) - \quantityinv_2(\idxitem)}.
\end{align*}

The following theorem states that given any arbitrary ordinal estimator, there exists a cardinal estimator that performs strictly uniformly better than this ordinal estimator, simultaneously on Kendall-tau distance and Spearman's footrule distance (cf. Theorem~\ref{thm:sort_ours_uniform_prior} for 0-1 loss).

\begin{theorem}\label{thm:other_metrics}
Suppose that the true ranking $\quantityinvgt$ is drawn uniformly at random from the collection of all possible rankings. For any arbitrary ordinal estimator $\quantityinvordsortkt$, there exists a cardinal estimator with access to one call to the ordinal estimator $\quantityinvordsortkt$ that strictly uniformly dominates the ordinal estimator $\quantityinvordsortkt$ with respect to Kendall-tau distance and Spearman's footrule distance. The computatinal complexity of this cardinal estimator is polynomial in the number of items $\numitems$, in addition to the time taken by one call to the ordinal estimator $\quantityinvordsortkt$.
\end{theorem}
This result demonstrates the generality of our results in the main text with respect to various (not only 0-1) loss functions. The remainder of this section is devoted to the proof of Theorem~\ref{thm:other_metrics}.

\subsection{Proof of Theorem~\ref{thm:other_metrics}}\label{app:proof_other_metrics}

We first present the construction of a cardinal estimator $\quantityinvcardourssortkt$, which has access to one call to any arbitrary ordinal estimator $\quantityinvordsortkt$. For any $\idxitem, \idxitem' \in [\numitems]$ with $\idxitem \ne \idxitem'$, we call the pair of items $(\idxitem, \idxitem')$ ``\topoidentical'' under the set of ordinal observations $\infoordall$, if the following conditions are met. There is no direct comparison between item $\idxitem$ and item $\idxitem'$ in  $\infoordall$. For any item $\idxitemcontrib \not\in \{\idxitem, \idxitem'\}$, the set $\infoordall$ includes a comparison between item $\idxitem$ and item $\idxitemcontrib$, if and only if the set $\infoordall$ includes a comparison between item $\idxitem'$ and item $\idxitemcontrib$. Moreover, if two comparisons $(\idxitem
, \idxitemcontrib)$ and $(\idxitem', \idxitemcontrib)$ are in the set $\infoordall$, their comparison results are the same, that is, $\indicatorevent{\idxitem\greateritem\idxitemcontrib} = \indicatorevent{\idxitem'\greateritem\idxitemcontrib}$. Note that it is possible that item $\idxitem$ is compared to no item in $\infoordall$ (and hence item $\idxitem'$ is also compared to no item).

For any item $\idxitem\in [\numitems]$ and any possible set of ordinal observations $\infoordall$, we define the following sets:
\begin{align*}
    \setcomparelarger(\idxitem, \infoordall) = & \{\idxitemcontrib\in [\numitems], \idxitemcontrib \ne \idxitem \mid \text{there exists a directed path from item $\idxitemcontrib$ to item $\idxitem$ in the graph $\graph(\infoordall)$}\}\\
    \setcomparesmaller(\idxitem, \infoordall) = & \{\idxitemcontrib\in [\numitems], \idxitemcontrib \ne \idxitem \mid \text{there exists a directed path from item $\idxitem$ to item $\idxitemcontrib$ in the graph $\graph(\infoordall)$}\}.
\end{align*}
In words, $\setcomparelarger(\idxitem, \infoordall)$ is the set of items that are ranked higher than item $\idxitem$ according to the set of ordinal observations $\infoordall$, and $\setcomparesmaller(\idxitem, \infoordall)$ is the set of items that are  ranked lower than item $\idxitem$. For any \topoidentical pair $(\idxitem, \idxitem')$, we have $\setcomparelarger(\idxitem, \infoordall)= \setcomparelarger(\idxitem', \infoordall)$ and  $\setcomparesmaller(\idxitem, \infoordall)= \setcomparesmaller(\idxitem', \infoordall)$, so we denote $\setcomparelarger(\idxitem, \idxitem', \infoordall) \defn \setcomparelarger(\idxitem, \infoordall)$ and $\setcomparesmaller(\idxitem, \idxitem', \infoordall) \defn \setcomparesmaller(\idxitem, \infoordall)$. Now we present a cardinal estimator $\quantityinvcardourssortkt$ in Algorithm~\ref{alg:sort_kendall_tau}.

In words, Algorithm~\ref{alg:sort_kendall_tau} obtains an initial estimated ranking $\quantityinvinit$ by making one call to the given ordinal estimator $\quantityinvordsortkt$. Then Algorithm~\ref{alg:sort_kendall_tau} identifies two items that are \topoidentical. If such a \topoidentical pair $(\idxitem, \idxitem')$ exists, we perform the following two steps on this \topoidentical pair: 

\begin{enumeratex}[(1)]
    \item \emph{Line~\ref{line:kt_rearrange_start}-\ref{line:kt_rearrange_end}: } Using the set of ordinal observations $\infoordall$, we obtain a new ranking $\quantityinvest$ by re-arranging the items in the initial estimated ranking $\quantityinvinit$. In this new ranking $\quantityinvest$, we keep all items outside $\setcomparelarger\union \setcomparesmaller\union \{\idxitem, \idxitem'\}$ in the same positions as they are in $\quantityinvinit$. We re-arrange the items in $\setcomparelarger\union \setcomparesmaller\union \{\idxitem, \idxitem'\}$, so that they occupy the remaining positions; the set $\setcomparelarger$ is ranked higher than items $\{\idxitem, \idxitem'\}$, and the set $\setcomparesmaller$ is ranked lower than items $\{\idxitem, \idxitem'\}$. Moreover, the relative ranking of items within each set ($\setcomparelarger$, $\setcomparesmaller$ or $\{\idxitem, \idxitem'\}$) is preserved. That is, if $\idxitemcontrib, \idxitemcontrib'\in \setcompare$ with some $\setcompare \in  \{\setcomparelarger, \setcomparesmaller, \{\idxitem, \idxitem'\}\}$, we have $\quantityinvest(\idxitemcontrib) < \quantityinvest(\idxitemcontrib')$ if and only if $\quantityinvinit(\idxitemcontrib) < \quantityinvinit(\idxitemcontrib')$.
    \item \emph{Line~\ref{line:kt_canonical_start}-\ref{line:kt_canonical_end}: } We sample uniformly at random a \score for each item in the \topoidentical pair $(\idxitem, \idxitem')$. Based on this pair of \scores, we call the canonical estimator to decide the relative ordering of these two items. Depending on the outcome of the canonical estimator, we keep the relative ordering of these two items unchanged, or flip the two items accordingly.
\end{enumeratex}

This completes the description of the cardinal estimator $\quantityinvcardourssortkt$. 

\begin{algorithm}[t!]
    \SetKw{Break}{break from both for-loops and go to Line~\ref{line:kt_output}.}

    Deduce the ordinal observations $\infoordall$ from the cardinal observations $\infocardall$. Compute an initial estimated ranking $\quantityinvinit=\quantityinvordsortkt(\infoordall)$.\label{line_kt:call_ordinal}\;
    \For{$\idxitem=1, \ldots, \numitems$}{
        \For{$\idxitem' = (\idxitem + 1), \ldots, \numitems$}{
            \If{the pair $(\idxitem, \idxitem')$ is \topoidentical, and both items $\idxitem$ and $\idxitem'$ have at least one \score each from $\infocardall$}{
                Compute $\setcomparelarger \defn \setcomparelarger(\idxitem, \idxitem', \infoordall)$. Denote the items in $\setcomparelarger$ as $\idxitem^+_1 \greateritem \cdots \greateritem \idxitem^+_{\abs{\setcomparelarger}}$ under the ranking $\quantityinit$.\label{line:kt_rearrange_start}\;
                Compute $\setcomparesmaller \defn \setcomparesmaller(\idxitem, \idxitem', \infoordall)$. Denote the items in $\setcomparesmaller$ as $\idxitem^-_1 \greateritem \cdots \greateritem \idxitem^-_{\abs{\setcomparesmaller}}$ under the ranking $\quantityinit$.\;
                $\varpositionskt = \{\idxitemcontrib\in \setcomparelarger \union \setcomparesmaller \union \{\idxitem, \idxitem'\} \mid \quantityinvinit(\idxitemcontrib) \}$. \label{line:kt_rearrange_placing_start}\;
                $\quantityinvest \leftarrow \quantityinvinit$.\;
                \eIf{$\idxitem \greateritem \idxitem' \text{ under }\quantityinvinit$}{
                    Re-arrange items in $\setcomparelarger \union \setcomparesmaller \union \{\idxitem, \idxitem'\}$ in $\quantityinvest$, such that they still occupy \varpositionskt, and $\idxitem^+_1 \greateritem \cdots \greateritem \idxitem^+_{\abs{\setcomparelarger}} \greateritem\idxitem\greateritem \idxitem' \greateritem \idxitem^-_1 \greateritem \cdots \greateritem \idxitem^-_{\abs{\setcomparesmaller}}$.\label{line:kt_rearrange_placement_one}\;
                }{ 
                    Re-arrange items in $\setcomparelarger \union \setcomparesmaller \union \{\idxitem, \idxitem'\}$ in $\quantityinvest$, such that they still occupy \varpositionskt, and $\idxitem^+_1 \greateritem \cdots \greateritem \idxitem^+_{\abs{\setcomparelarger}} \greateritem\idxitem'\greateritem \idxitem \greateritem \idxitem^-_1 \greateritem \cdots \greateritem \idxitem^-_{\abs{\setcomparesmaller}}$.\label{line:kt_rearrange_placement_two}\;
                }\label{line:kt_rearrange_end}
                From all of the scores of item $\idxitem$ in $\infocardall$, sample one uniformly at random and denote it as $\info_{\idxitem}$. Likewise denote $\info_{\idxitem'}$ as a randomly chosen score of item $\idxitem'$ from $\infocardall$.\label{line:kt_canonical_start}\;
                \If{$\quantitycardourscanonical(\info_{\idxitem}, \info_{\idxitem'})$ indicates a relative ordering of the pair $(\idxitem, \idxitem')$ different from $\quantityinvest$ }{
                    Let $\quantityinvest_\text{flip}$ be the ranking obtained by flipping items $\idxitem$ and $\idxitem'$ in $\quantityinvest$.\;
                    $\quantityinvest\leftarrow \quantityinvest_\text{flip}$.\;
                }\label{line:kt_canonical_end} 
                \Break\label{line:kt_break}
            } 
        } 
    } 
    Output $\quantityinvcardourssortkt(\assignment, \infocardall) = \quantityinvest$.\label{line:kt_output}\;

    \caption{Our cardinal ranking estimator $\quantityinvcardourssortkt(\assignment, \infocardall)$ concerning Kendall-tau distance and Spearman's footrule distance.}\label{alg:sort_kendall_tau}
\end{algorithm}

We now show that the cardinal estimator $\quantityinvcardourssortkt$ takes polynomial time in the number of items $\numitems$, in addition to the time taken by one call to the given ordinal estimator $\quantityinvordsortkt$. To check if a pair of items $(\idxitem, \idxitem')$ is \topoidentical, it takes polynomial time to go through the pairwise comparisons in $\infoordall$. Hence, it takes polynomial time to identify a \topoidentical pair (or determine that such a pair does not exist). For any \topoidentical pair, in the re-arranging step, the set $\setcomparesmaller(\idxitem, \idxitem', \infoordall)$ can be found by a graph traversal from node $\idxitem$. The set  $\setcomparelarger(\idxitem, \idxitem', \infoordall)$  can be found by a graph traversal from node $\idxitem$ on the graph $\graph(\infoordall)$ but with all edges reversed. Both traversals take polynomial time. Hence, Algorithm~\ref{alg:sort_kendall_tau} takes polynomial time, in addition to one call to the ordinal estimator $\quantityinvordsortkt$.\\

We now present the proof for the uniform strict dominance of the cardinal estimator $\quantityinvcardourssortkt$ over the given ordinal estimator $\quantityinvordsortkt$. Given any two rankings $\quantityinv_1, \quantityinv_2$ and any two items $(\idxitem, \idxitem')$, we denote $\ktcontribution(\quantityinv_1, \quantityinv_2, \idxitem, \idxitem') \defn \indicatorevent{\indicatorevent{\quantityinv_1(\idxitem) > \quantityinv_1(\idxitem')} \ne \indicatorevent{\quantityinv_2(\idxitem) > \quantityinv_2(\idxitem')}}$ as Kendall-tau distance between $\quantityinv_1$ and $\quantityinv_2$ contributed by the pair of items $(\idxitem, \idxitem')$. Then we can write Kendall-tau distance between $\quantityinv_1, \quantityinv_2$ as
\begin{align}
    \losskt(\quantityinv_1, \quantityinv_2) = & \sum_{\substack{ \idxitem\in[\numitems], \idxitem'\in [\numitems] :\\ \quantityinv_1(\idxitem) < \quantityinv_2(\idxitem')}} \indicatorevent{\quantityinv_2(\idxitem) > \quantityinv_2(\idxitem')}\nonumber\\
    = & \sum_{1\le \idxitem < \idxitem'\le \numitems} \indicatorevent{\indicatorevent{\quantityinv_1(\idxitem) > \quantityinv_1(\idxitem')} \ne \indicatorevent{\quantityinv_2(\idxitem) > \quantityinv_2(\idxitem')}}\nonumber\\
    = & \sum_{1 \le \idxitem < \idxitem' \le \numitems}\ktcontribution(\quantityinv_1, \quantityinv_2, \idxitem, \idxitem').\label{eq:kt_rewrite}
\end{align}

For Spearman's footrule dsitance, for each item $\idxitem\in [\numitems]$, we call the term $\abs{\quantityinv_1(\idxitem) - \quantityinv_2(\idxitem)}$ as Spearman's footrule distance between $\quantityinv_1$ and $\quantityinv_2$ contributed by item $\idxitem$.

We analyze Step (1) of re-arranging the items and Step (2) of evoking the canonical estimator separately. 
The following rearrangement inequality is used for analyzing both steps. For any $\tmpa_1, \tmpa_2, \tmpb_1, \tmpb_2\in \reals$ where $\tmpa_1 < \tmpa_2$ and $\tmpb_1 < \tmpb_2$, it is straightforward to verify that
\begin{align}
    \abs{\tmpa_1 - \tmpb_2} + \abs{\tmpa_2 - \tmpb_1} \ge \abs{\tmpa_1 - \tmpb_1} + \abs{\tmpa_2 - \tmpb_2}.\label{eq:rearrangement}
\end{align}

We first analyze the re-arranging step in Line~\ref{line:kt_rearrange_start}-\ref{line:kt_rearrange_end} of Algorithm~\ref{alg:sort_kendall_tau}. We denote the random variable  $\quantityinvestre$ as the estimated ranking after the re-arranging step (that is, the value of the quantity $\quantityinvest$ after Line~\ref{line:kt_rearrange_end} of Algorithm~\ref{alg:sort_kendall_tau}). The re-arranged ranking $\quantityinvestre$ is a deterministic function of the initial ranking $\quantityinvinit$. The following lemma proves a deterministic result about this re-arranging step.

\begin{lemma}\label{lemma_rearrange}
For any true ranking $\quantityinvgt$, any set of ordinal observations $\infoordall$ consistent with the true ranking, and any initial estimated ranking $\quantityinvinit$, the re-arranged ranking $\quantityinvestre$ yields smaller or equal loss compared to the initial ranking $\quantityinvinit$, regarding Kendall-tau distance and Spearman's footrule distance. That is,
\begin{subequations}
\begin{align}
    \losskt(\quantityinvestre, \quantityinvgt) & \le \losskt(\quantityinvinit, \quantityinvgt)\label{eq:kt_kt_step_rearranging}\\
    \losssf(\quantityinvestre, \quantityinvgt) & \le \losssf(\quantityinvinit, \quantityinvgt).\label{eq:kt_sf_step_rearranging}
\end{align}
\end{subequations}
\end{lemma}

The lemma is proved at the end of this section.\\

Now we turn to analyze the second step of calling the canonical estimator on the \topoidentical pair. This step starts from the re-arranged ranking $\quantityinvestre$. Denote $\eventexist$ as the event that Algorithm~\ref{alg:sort_kendall_tau} identifies some \topoidentical pair (that is, Line~\ref{line:kt_rearrange_start}-\ref{line:kt_break} of Algorithm~\ref{alg:sort_kendall_tau} is executed). Then $\eventexist^c$ denotes the event that no \topoidentical pair is found. If there exists no \topoidentical pairs, then the second step in Line~\ref{line:kt_canonical_start}-\ref{line:kt_canonical_end} of Algorithm~\ref{alg:sort_kendall_tau} is never executed. Trivially, the final output $\quantityinvcardourssortkt$ is identical to the re-arranged ranking $\quantityinvestre$. We have
\begin{subequations}
\begin{align}
    \Expect[\losskt(\quantityinvcardourssortkt, \quantityinvgt) \given  \eventexist^c] & = \Expect[\losskt(\quantityinvestre, \quantityinvgt) \given \eventexist^c]\label{eq:kt_kt_conditioned_no_twin}\\
    \Expect[\losssf(\quantityinvcardourssortkt, \quantityinvgt) \given \eventexist^c] & = \Expect[\losssf(\quantityinvestre, \quantityinvgt) \given \eventexist^c]\label{eq:kt_sf_conditioned_no_twin}.
\end{align}
\end{subequations}

It remains to consider the case when the event $\eventexist$ is true. We start by showing that the event $\eventexist$ happens with non-zero probability. Consider any arbitrary true ranking $\quantityinvgt$. Under this true ranking, denote the top item as $\idxitem^{(1)}$, and denote the second-ranked item as $\idxitem^{(2)}$. Conditioned on this true ranking, consider the set of pairwise comparisons $\itemsofgraderunorderall$ such that the set $\itemsofgraderunorderall$ includes comparisons between item $\idxitem^{(1)}$ and a subset of $\min\{\floor{\numgraders / 2}, \numitems - 2\}$ items from $[\numitems]\setminus \{\idxitem^{(1)}, \idxitem^{(2)}\}$. Assume that $\itemsofgraderunorderall$ also includes comparisons between item $\idxitem^{(2)}$ and the identical subset of items from $[\numitems] \setminus \{\idxitem^{(1)}, \idxitem^{(2)}\}$. The rest of the comparisons can be arbitrary between the $(\numitems - 2)$ items in $[\numitems] \setminus \{\idxitem^{(1)}, \idxitem^{(2)}\}$. Recall that $1 < \numgraders < {\numitems \choose 2}$, so such a set $\itemsofgraderunorderall$ arises with non-zero probability. Hence, the event $\eventexist$ happens with non-zero probability.

Note that the set of ordinal observations $\infoordall$ fully determines the \topoidentical pair (if any) selected by Algorithm~\ref{alg:sort_kendall_tau}. Since the event $\eventexist$ happens with non-zero probability, there exists $\infoordallobs$ such that $\Prob(\infoordall = \infoordallobs, \eventexist) > 0$. We condition on the event $\eventexist$ and any set of ordinal observations $\infoordallobs$ such that $\Prob(\infoordall = \infoordallobs, \eventexist) > 0$. We denote the two items in the \topoidentical pair selected by the algorithm as items $(\idxitem(\infoordallobs), \idxitem'(\infoordallobs))$ (or items ($\idxitem,\idxitem')$ in short). In what follows, we consider Kendall-tau distance and Spearman's footrule separately.\\

\noindent\textbf{Kendall-tau distance: } For each $\idxitemcontrib, \idxitemcontrib'\in [\numitems]$ with $\idxitemcontrib \ne \idxitemcontrib'$, we consider Kendall-tau distance contributed by the pair $(\idxitemcontrib, \idxitemcontrib')$ . Recall that conditioned on the event event $\event$ and the set of ordinal observations $\infoordallobs$, the only pair that can be flipped by Algorithm~\ref{alg:sort_kendall_tau} is $(\idxitem(\infoordallobs), \idxitem'(\infoordallobs))$. We only need to consider the pairs $(\idxitemcontrib, \idxitemcontrib')$ such that the relative ordering of $(\idxitemcontrib, \idxitemcontrib')$ can be potentially changed by flipping the pair $(\idxitem, \idxitem')$. We consider the following two cases separately.

\noindent\textit{Case 1: We consider Kendall-tau distance contributed by the pair $(\idxitem, \idxitem')$ itself. That is, $\{\idxitemcontrib, \idxitemcontrib'\} = \{\idxitem ,\idxitem'\}$.}

Consider the ranking $\quantityinvestre$ from the re-arranging step. We have
\begin{align}
    \Expect[\ktcontribution(\quantityinvestre, \quantityinvgt, \idxitem, \idxitem') \given \infoordall = \infoordallobs, \eventexist] = & \sum_\quantityinv \Expect[\ktcontribution(\quantityinvestre, \quantityinvgt, \idxitem, \idxitem') \given \quantityinvgt = \quantityinv, \infoordall = \infoordallobs, \eventexist]\cdot \Prob(\quantityinvgt = \quantityinv \given \infoordall = \infoordallobs, \eventexist)\nonumber\\
    \stackrel{\stepone}{=} & \sum_\quantityinv \Expect[\ktcontribution(\quantityinvestre, \quantityinv, \idxitem, \idxitem') \given \quantityinvgt = \quantityinv, \infoordall = \infoordallobs, \eventexist]\cdot\Prob(\quantityinvgt = \quantityinv \given \infoordall = \infoordallobs)\nonumber\\
    \stackrel{\steptwo}{=} & \frac{1}{\numorders(\infoordallobs)}\sum_{\quantityinv\in \topo(\infoordallobs)} \Expect[\ktcontribution(\quantityinvestre, \quantityinv, \idxitem, \idxitem') \given \quantityinvgt = \quantityinv, \infoordall = \infoordallobs, \eventexist].\label{eq:kt_kt_case_one_intermediate}
\end{align}
where equality \stepone is true because $\quantityinvgt$ is independent of $\eventexist$ conditioned on $\infoordall$. Equality \steptwo is true because of~\eqref{eq:posterior} in Lemma~\ref{lem:ranking}.

Recall that the initial ranking $\quantityinvinit$ is obtained by calling the (possibly randomized) ordinal estimator $\quantityinvordsortkt$ taking input $\infoordall$, and the re-arranged ranking $\quantityinvestre$ is fully determined by $\quantityinvinit$. Hence, we further write~\eqref{eq:kt_kt_case_one_intermediate} as
\begin{align}
\Expect[\ktcontribution (\quantityinvestre, \quantityinvgt, & \idxitem, \idxitem')  \given \infoordall = \infoordallobs, \event] \nonumber\\  
= & \frac{1}{\numorders(\infoordallobs)}\sum_{\quantityinvest} \sum_{\quantityinv\in \topo(\infoordallobs)} \Expect[\ktcontribution(\quantityinvest, \quantityinv, \idxitem, \idxitem') \given \quantityinvestre = \quantityinvest, \quantityinvgt = \quantityinv, \infoordall = \infoordallobs, \eventexist] \cdot  \Prob(\quantityinvestre = \quantityinvest \given \quantityinvgt = \quantityinv, \infoordall = \infoordallobs, \eventexist)\nonumber\\
\stackrel{\stepone}{=} & \frac{1}{\numorders(\infoordallobs)}\sum_{\quantityinvest} \sum_{\quantityinv\in \topo(\infoordallobs)} \Expect[\ktcontribution(\quantityinvest, \quantityinv, \idxitem, \idxitem') \given \quantityinvestre = \quantityinvest, \quantityinvgt = \quantityinv, \infoordall = \infoordallobs, \eventexist] \cdot \Prob(\quantityinvestre = \quantityinvest \given \infoordall = \infoordallobs),\label{eq:kt_kt_half_expression}
\end{align}
where equality \stepone is true, because $\quantityinvordsortkt$ is independent of the true ranking $\quantityinvgt$ and the event $\eventexist$ conditioned on $\infoordall$. Hence, $\quantityinvestre$ is independent of the true ranking $\quantitygt$ and the event $\eventexist$ conditioned on $\infoordall$.

Define the set $\setposterior_{\idxitem \greateritem\idxitem'}\subseteq\topo(\infoordallobs)$ as the collection of topological orderings where $\idxitem$ is ranked higher than $\idxitem'$. Define the set $\setposterior_{\idxitem \lessitem\idxitem'}\subseteq \topo(\infoordallobs)$ as the collection of topological orderings where $\idxitem$ is ranked lower than $\idxitem$. Then $\{\setposterior_{\idxitem \greateritem\idxitem'}, \setposterior_{\idxitem \lessitem\idxitem'}\}$ is a partition of the collection of all topological orderings, $\topo(\infoordallobs)$. Given that the pair $(\idxitem, \idxitem')$ is \topoidentical, for any ranking $\quantityinv\in \topo(\infoordallobs)$, we can flip items $(\idxitem, \idxitem')$, and the flipped ranking is still a topological ordering.  Flipping the items $(\idxitem, \idxitem')$ defines a bijection between the set $\setposterior_{\idxitem \greateritem\idxitem'}, \setposterior_{\idxitem \lessitem\idxitem'}$, so we have $\abs{\setposterior_{\idxitem \lessitem\idxitem'}} = \abs{ \setposterior_{\idxitem \lessitem\idxitem'}}$. Any ranking $\quantityinvestre$ is correct on one and only one of the sets $\setposterior_{\idxitem \greateritem\idxitem'}$ and $\setposterior_{\idxitem \lessitem\idxitem'}$, and hence the re-arranged ranking $\quantityinvestre$ is correct on exactly half of the topological orderings. For any $\quantityinvest$, we have
\begin{align}
    \sum_{\quantityinv\in \topo(\infoordallobs)} \Expect[\ktcontribution(\quantityinv, \quantity, \idxitem, \idxitem') \given \quantityinvestre = \quantityinvest, \quantityinvgt = \quantityinv, \infoordall = \infoordallobs, \eventexist] = \frac{1}{2}.\label{eq:kt_kt_half}
\end{align}

Combining~\eqref{eq:kt_kt_half} with~\eqref{eq:kt_kt_half_expression} yields 
\begin{align*}
    \Expect[\ktcontribution(\quantityinvestre, \quantityinvgt, \idxitem, \idxitem') \given \infoordall = \infoordallobs, \eventexist] = \frac{1}{2}.
\end{align*}

Now consider the cardinal estimator. Similar to the proof of Theorem~\ref{thm:sort_ours_uniform_prior}, we have
\begin{align*}
    \Expect[\ktcontribution(\quantityinvcardourssortkt, \quantityinvgt, \idxitem, \idxitem') \given \infoordall = \infoordallobs, \eventexist] < \frac{1}{2}.
\end{align*}

Consequently, in Case 1, we have
\begin{align}
        \Expect[\ktcontribution(\quantityinvcardourssortkt, \quantityinvgt, \idxitem, \idxitem') \given \infoordall = \infoordallobs, \eventexist] <     \Expect[\ktcontribution(\quantityinvestre, \quantityinvgt, \idxitem, \idxitem') \given \infoordall = \infoordallobs, \eventexist].\label{eq:kt_kt_case_one}
\end{align}

\noindent\textit{Case 2: Consider any pair $(\idxitemcontrib, \idxitemcontrib')$ that is not identical to the pair $(\idxitem, \idxitem')$. Since the relative ordering of the pair $(\idxitemcontrib, \idxitemcontrib')$ is changed by flipping the pair $(\idxitem, \idxitem')$, then one item has to be either $\idxitem$ or $\idxitem'$. Without loss of generality, assume $\idxitemcontrib\not\in \{\idxitem, \idxitem'\}$ and $\idxitemcontrib'\in \{\idxitem, \idxitem'\}$. We consider pairs in the form of $(\idxitemcontrib, \idxitem)$ and $(\idxitemcontrib, \idxitem')$.}

If the position of $\idxitemcontrib$ is not in between item $\idxitem$ and item $\idxitem'$ in the ranking $\quantityinvestre$ (that is, if $\quantityinvestre(\idxitemcontrib) < \min\{\quantityinvestre(\idxitem), \quantityinvestre(\idxitem')\}$ or $\quantityinvestre(\idxitemcontrib) > \max\{\quantityinvestre(\idxitem), \quantityinvestre(\idxitem')\}$), then flipping the pair $(\idxitem, \idxitem')$ does not change the relative ordering of the pair $(\idxitemcontrib, \idxitem)$ or $(\idxitemcontrib, \idxitem')$. Now we restrict our attention to item $\idxitemcontrib$ ranked in between item $\idxitem$ and item $\idxitem'$ in the ranking $\quantityinvestre$. Moreover, if the position of $\idxitemcontrib$ is not in between the positions of item $\idxitem$ and item $\idxitem'$ in the true ranking (that is, if $\quantityinvgt(\idxitemcontrib) < \min\{\quantityinvgt(\idxitem), \quantityinvgt(\idxitem')\}$ or $\quantityinvgt(\idxitemcontrib) > \max\{\quantityinvgt(\idxitem), \quantityinvgt(\idxitem')\}$), then whether flipping the pair $(\idxitem, \idxitem')$ or not, one and only one of the two comparisons $(\idxitemcontrib, \idxitem)$ and $(\idxitemcontrib', \idxitem)$ is correct. Hence, we only need to consider each item $\idxitemcontrib$ ranked between the two items $\idxitem$ and $\idxitem'$, in both the re-arranged ranking $\quantityinvestre$ and the true ranking $\quantityinvgt$. For each such item $\idxitemcontrib$, for any re-arranged ranking $\quantityinvestre$, we have the determinisitc equality
\begin{align}
\begin{split}
    & \ktcontribution(\quantityinvestre, \quantityinvgt,  \idxitemcontrib, \idxitem) +  \ktcontribution(\quantityinvestre, \quantityinvgt, \idxitemcontrib, \idxitem') =     2\ktcontribution(\quantityinvestre, \quantityinvgt, \idxitem, \idxitem')\\
    & \ktcontribution(\quantityinvcardourssortkt, \quantityinvgt,  \idxitemcontrib, \idxitem) +  \ktcontribution(\quantityinvcardourssortkt, \quantityinvgt, \idxitemcontrib, \idxitem') =     2\ktcontribution(\quantityinvcardourssortkt, \quantityinvgt, \idxitem, \idxitem')
\end{split}\label{eq:kt_kt_deterministic_equality_between}
\end{align}
Combining~\eqref{eq:kt_kt_deterministic_equality_between} and~\eqref{eq:kt_kt_case_one}, for each item $\idxitemcontrib$ ranked in between item $\idxitem$ and item $\idxitem'$ in both the re-arranged ranking $\quantityinvestre$ and the true ranking $\quantityinvgt$, we have
\begin{align}
    \Expect[\ktcontribution(\quantityinvcardourssortkt, \quantityinvgt,  \idxitemcontrib, \idxitem) +  \ktcontribution(\quantityinvcardourssortkt, \quantityinvgt, & \idxitemcontrib, \idxitem') \given  \infoordall = \infoordallobs, \eventexist]   \nonumber\\ &< \Expect[\ktcontribution(\quantityinvestre, \quantityinvgt,  \idxitemcontrib, \idxitem) +  \ktcontribution(\quantityinvestre, \quantityinvgt, \idxitemcontrib, \idxitem') \given \infoordall = \infoordallobs, \eventexist].\label{eq:kt_kt_case_two}
\end{align}
Combining the expression of Kendall-tau distance in~\eqref{eq:kt_rewrite} with the two cases in~\eqref{eq:kt_kt_case_one} and~\eqref{eq:kt_kt_case_two} of which the relative ordering of some pair $(\idxitemcontrib, \idxitemcontrib')$ is changed, we have
\begin{align*}
    \Expect[\losskt(\quantityinvcardourssortkt, \quantityinvgt) \given \infoordall = \infoordallobs, \eventexist] < \Expect[\losskt(\quantityinvestre, \quantityinvgt) \given \infoordall = \infoordallobs, \eventexist].
\end{align*}

Recall that $\Prob(\infoordall = \infoordallobs, \eventexist) > 0$ for some $\infoordallobs$. Taking an expectation over $\infoordall$ yields
\begin{align}
    \Expect[\losskt(\quantityinvcardourssortkt, \quantityinvgt) \given \eventexist] < \Expect[\losskt(\quantityinvestre, \quantityinvgt)\given  \eventexist].\label{eq:kt_kt_conditioned_twin}
\end{align}

Combining~\eqref{eq:kt_kt_conditioned_twin} and~\eqref{eq:kt_kt_conditioned_no_twin} yields
\begin{align}
    \Expect[\losskt(\quantityinvcardourssortkt, \quantityinvgt)] < \Expect[\losskt(\quantityinvestre, \quantityinvgt)].\label{eq:kt_kt_step_canonical}
\end{align}

Finally, combining~\eqref{eq:kt_kt_step_canonical} with inequality~\eqref{eq:kt_kt_step_rearranging} for the re-arranging step completes the proof for Kendall-tau distance.\\

\noindent\textbf{Spearman's footrule distance: } We condition on the event $\eventexist$ and any set of ordinal observations $\infoordallobs$ such that $\Prob(\infoordall = \infoordallobs, \eventexist) > 0$. Since only one pair $(\idxitem(\infoordallobs), \idxitem'(\infoordallobs))$ can be flipped by Algorithm~\ref{alg:sort_kendall_tau}, we only need to consider Spearman's footrule distance contributed by these two items. Consider any ranking $\quantityinv_{\idxitem\greateritem \idxitem'}\in \setposterior_{\idxitem\greateritem\idxitem'}$. Let $\quantityinv_{\idxitem\lessitem \idxitem'}$ be the ranking obtained by flipping items $(\idxitem, \idxitem')$ in $\quantityinv_{\idxitem \greateritem\idxitem'}$. Then we have $\quantityinv_{\idxitem\lessitem \idxitem'}\in \setposterior_{\idxitem\lessitem\idxitem'}$. For any such pair $\{\quantityinv_{\idxitem\greateritem\idxitem'}, \quantityinv_{\idxitem\lessitem\idxitem'}\}$, we have

\begin{align}
    \Prob(\quantityinvgt\in \{\quantityinv_{\idxitem\greateritem\idxitem'}, \quantityinv_{\idxitem\lessitem\idxitem'}\}, \infoordall = \infoordallobs, \eventexist) = & \Prob(\quantityinvgt\in \{\quantityinv_{\idxitem\greateritem\idxitem'}, \quantityinv_{\idxitem\lessitem\idxitem'}\} \given \infoordall = \infoordallobs, \eventexist)\cdot \Prob(\infoordall = \infoordallobs, \eventexist)\nonumber\nonumber\\
    = & \Prob(\quantityinvgt\in \{\quantityinv_{\idxitem\greateritem\idxitem'}, \quantityinv_{\idxitem\lessitem\idxitem'}\}\given \infoordall = \infoordallobs, \eventexist)\cdot \Prob(\infoordall = \infoordallobs, \eventexist)\nonumber\\
    = & \Prob(\quantityinvgt\in \{\quantityinv_{\idxitem\greateritem\idxitem'}, \quantityinv_{\idxitem\lessitem\idxitem'}\}\given \infoordall = \infoordallobs)\cdot \Prob(\infoordall = \infoordallobs, \eventexist)\label{eq:kt_sf_strict_nonzero_probability_intermediate}\\
    \stackrel{\stepone}{>} & 0\label{eq:kt_sf_strict_nonzero_probability},
\end{align}
where inequality \stepone is true, because the two terms in~\eqref{eq:kt_sf_strict_nonzero_probability_intermediate} are both non-zero. The first term in~\eqref{eq:kt_sf_strict_nonzero_probability_intermediate} is non-zero by the fact that $\quantityinv_{\idxitem\greateritem\idxitem'}, \quantityinv_{\idxitem\lessitem\idxitem'}$ are topological orderings, and by~\eqref{eq:posterior} in Lemma~\ref{lem:ranking}. The second term in~\eqref{eq:kt_sf_strict_nonzero_probability_intermediate} is non-zero, because by construction we find $\infoordallobs$ such that the second term $\Prob(\infoordall = \infoordallobs, \eventexist) > 0$.

Now we analyze the Spearman's footrule distance conditioned on the event $\quantityinvgt \in \{\quantityinv_{\idxitem\greateritem\idxitem'},  \quantityinv_{\idxitem\lessitem\idxitem'}\}$. Using the argument deriving~\eqref{eq:kt_kt_case_one}, we can further derive
\begin{align}
   \Expect[\ktcontribution(\quantityinvcardourssortkt, \quantityinvgt, \idxitem, \idxitem') \given \quantityinvgt \in \{\quantityinv_{\idxitem\greateritem\idxitem'},  \quantityinv_{\idxitem\lessitem\idxitem'}\} &, \infoordall = \infoordallobs,  \eventexist] \nonumber \\
   < & \Expect[\ktcontribution(\quantityinvestre, \quantityinvgt, \idxitem, \idxitem') \given \quantityinvgt \in \{\quantityinv_{\idxitem\greateritem\idxitem'}, \quantityinv_{\idxitem\lessitem\idxitem'}\}, \infoordall = \infoordallobs, \eventexist].\label{eq:kt_sf_pair}
\end{align}
By the rearrangement inequality~\eqref{eq:rearrangement}, if the relative ordering of the pair $(\idxitem, \idxitem')$ is correct, then Spearman's footrule distance does not increase compared to the ranking with the relative ordering of $(\idxitem, \idxitem')$ incorrect. Eq.~\eqref{eq:kt_sf_pair} implies that conditioned on $\infoordallobs$, the event $\eventexist$ and the event of $\quantityinvgt\in \{\quantityinv_{\idxitem\greateritem\idxitem'}, \quantityinv_{\idxitem\lessitem\idxitem'}\}$, the probability that the cardinal estimator $\quantityinvcardourssortkt$ gives the correct relative ordering of the pair $(\idxitem, \idxitem')$ is higher than the probability that $\quantityinvestre$ gives the correct relative ordering. Hence, for any set of ordinal observations $\infoordallobs$ and any pair $\{\quantityinv_{\idxitem\greateritem\idxitem'}, \quantityinv_{\idxitem\lessitem\idxitem'}\}$ of the true rankings, we have
\begin{align}
    \Expect[\losssf(\quantityinvcardourssortkt, \quantityinvgt)\given \quantityinvgt\in \{\quantityinv_{\idxitem\greateritem\idxitem'}, \quantityinv_{\idxitem\lessitem\idxitem'}\}, \infoordall= \infoordallobs, \eventexist] \le \Expect[\losssf(\quantityinvestre, \quantityinvgt) \given \quantityinvgt\in \{\quantityinv_{\idxitem\greateritem\idxitem'}, \quantityinv_{\idxitem\lessitem\idxitem'}\}, \infoordall = \infoordallobs, \eventexist].\label{eq:kt_sf_equal}
\end{align}

Note that directly applying the re-arrangement inequality does not translate the strict inequality from~\eqref{eq:kt_kt_case_one} to~\eqref{eq:kt_sf_equal}. The reason is that correctly ordering a \topoidentical pair does not guarantee strictly smaller Spearman's footrule distance. For example, if item $\idxitem$ and item $\idxitem'$ are the top-$2$ items in the true ranking, but are the bottom-$2$ items in $\quantityinvestre$. Then the relative ordering of the pair $(\idxitem, \idxitem')$ does not change the Spearman's footrule distance. In the rearrangement inequality~\eqref{eq:rearrangement}, strictly inequality holds if $\tmpa_1 \le \{\tmpb_1, \tmpb_2\} \le \tmpa_2$. Hence, we find one pair of true rankings $\{\quantityinv^*_{\idxitem\greateritem\idxitem'},  \quantityinv^*_{\idxitem\lessitem\idxitem'}\}$ such that one of the following is true:
\begin{align}\label{eq:kt_sf_condition_strict_inequality}
\begin{split}
    & \quantityinvgt_{\idxitem\greateritem\idxitem'}(\idxitem) \le \{\quantityinvestre(\idxitem), \quantityinvestre(\idxitem')\} \le \quantityinvgt_{\idxitem\greateritem\idxitem'}(\idxitem')\\
    \text{or }\qquad  & \quantityinvgt_{\idxitem\greateritem\idxitem'}(\idxitem') \le \{\quantityinvestre(\idxitem), \quantityinvestre(\idxitem')\} \le \quantityinvgt_{\idxitem\greateritem\idxitem'}(\idxitem).
\end{split}
\end{align} 
Then strictly inequality in~\eqref{eq:kt_sf_pair} holds on the pair $\{\quantityinv^*_{\idxitem\greateritem\idxitem'},  \quantityinv^*_{\idxitem\lessitem\idxitem'}\}$. Now we provide the construction of this pair $\{\quantityinv^*_{\idxitem\greateritem\idxitem'},  \quantityinv^*_{\idxitem\lessitem\idxitem'}\}$.

We start by constructing a topological ordering $\quantityinv(\idxitem, \idxitem', \infoordallobs)$ (or $\quantityinv$ in short) as follows. We topologically sort the items in $\setcomparelarger\defn \setcompare(\idxitem, \idxitem', \infoordallobs)$ and place them as the top $\abs{\setcomparelarger}$ items in $\quantityinv$. We topologically sort the items in $\setcomparesmaller\defn \setcomparesmaller(\idxitem, \idxitem, \infoordallobs)$ and place them as the bottom $\abs{\setcomparesmaller}$ items. Arbitrarily choose one item from $\{\idxitem, \idxitem'\}$ and place it at the position $(\abs{\setcomparelarger} + 1)$, and place the remaining item from the pair $\{\idxitem, \idxitem'\}$ at the position $(\numitems - \abs{\setcomparesmaller})$. Topologically sort the rest of the items, and place them in the remaining positions in $\quantityinv$.  

We now prove that the ranking $\quantityinv$ is a valid topological ordering. Assume for contradiction that $\quantityinv$ is not a valid topological ordering. Then there exists a pair $(\idxitemcontrib, \idxitemcontrib')$ that violates some pairwise comparison in $\infoordall$. Denote $\setcompare^c =[\numitems] \setminus (\setcomparelarger \union \setcomparesmaller \union \{\idxitem, \idxitem'\})$. Within each set $\setcomparelarger$, $\setcomparesmaller$ or $\setcompare^c$, the items are ordered by a topological ordering. Moreover, there is no direct comparison between item $\idxitem$ and item $\idxitem'$, so items $\{\idxitem, \idxitem'\}$ can be ranked with either $\idxitem\greateritem\idxitem'$ or $\idxitem\lessitem\idxitem'$. Hence, $\idxitemcontrib$ and $\idxitemcontrib'$ cannot belong to the same set of $\setcomparelarger, \setcompare, \setcompare^c$ or $\{\idxitem, \idxitem'\}$. By the definition of the sets $\setcomparelarger$ and $\setcomparesmaller$, in the true ranking $\setcomparelarger$ should be ranked higher than $\{\idxitem, \idxitem'\}$, and $\setcomparesmaller$ should be ranked lower than $\{\idxitem, \idxitem'\}$. In our ranking $\quantityinv$, we also rank $\setcomparelarger$ higher than $\{\idxitem, \idxitem'\}$, and $\setcomparesmaller$ lower than $\{\idxitem, \idxitem'\}$. Hence, if both item $\idxitemcontrib$ and item $\idxitemcontrib'$ are in $\setcomparelarger\union \setcomparesmaller\union \{\idxitem, \idxitem'\}$, the relative ordering between $(\idxitemcontrib, \idxitemcontrib')$ must be consistent with $\infoordall$. Then at least one item from the pair $(\idxitemcontrib, \idxitemcontrib')$ must be in $\setcompare^c$. Without loss of generality, assume $\idxitemcontrib'\in \setcompare^c$. Since $\idxitemcontrib$ and $\idxitemcontrib'$ cannot belong to the same set, we have $\idxitemcontrib\not\in \setcompare^c$. If $\idxitemcontrib\in \{\idxitem, \idxitem'\}$, since the pair $(\idxitemcontrib, \idxitemcontrib')$ violates some pairwise comparison, the items $(\idxitemcontrib, \idxitemcontrib')$ are compared in $\infoordall$, that is, $\idxitemcontrib'$ is compared to either $\idxitem$ or $\idxitem'$. By the definition of the sets $\setcomparelarger$ and $\setcomparesmaller$, it must be true that $\idxitemcontrib'\in \setcomparelarger$ or $\idxitemcontrib'\in \setcomparesmaller$, contradicting the assumption that $\idxitemcontrib'\in \setcompare^c$. If $\idxitemcontrib\in \setcomparelarger$, by construction the ranking $\quantityinv$ ranks $\idxitemcontrib$ higher than $\idxitemcontrib'$. Since the pair $(\idxitemcontrib, \idxitemcontrib')$ violates some pairwise comparison, the set $\infoordall$ must include the pairwise comparison $\idxitemcontrib' \greateritem\idxitemcontrib$. By the definition of $\setcomparelarger$, since $\idxitemcontrib\in \setcomparelarger$, there exists a path from $\idxitemcontrib$ to $\idxitem$. Concatenating the pairwise comparison $\idxitemcontrib' \greateritem \idxitemcontrib$ with the path from $\idxitemcontrib$ to $\idxitem$, we have a path from $\idxitemcontrib'$ to $\idxitem$. Hence, $\idxitemcontrib'\in \setcomparelarger$, contradicting the assumption that $\idxitemcontrib'\in \setcompare^c$. Similarly, $\idxitemcontrib\in \setcomparesmaller$ gives a contradiction. Hence, in the ranking $\quantityinv$ there exists no pair of items violating pairwise comparisons in $\infoordall$. By definition, the ranking $\quantityinv$ is a topological ordering. The ranking $\quantityinv$ places items $\{\idxitem, \idxitem'\}$ at positions $\{\abs{\setcomparelarger} + 1, \numitems-\abs{\setcomparesmaller}\}$. In Algorithm~\ref{alg:sort_kendall_tau}, the re-arranged ranking $\quantityinvestre$ places the set $\setcomparelarger$ before items $\{\idxitem, \idxitem'\}$, and the set $\setcomparesmaller$ after items $\{\idxitem, \idxitem'\}$. Hence, we have either $\quantityinv(\idxitem) \le \{\quantityinvestre(\idxitem), \quantityinvestre(\idxitem')\} \le \quantityinv(\idxitem')$ or $\quantityinv(\idxitem') \le \{\quantityinvestre(\idxitem), \quantityinvestre(\idxitem')\} \le \quantityinv(\idxitem)$.

Recall that when constructing $\quantityinv$, we arbitrarily place an item from the set $\{\idxitem, \idxitem'\}$ at position $(\abs{\setcomparelarger} + 1)$, and the remaining item from $\{\idxitem, \idxitem'\}$ at position $(\numitems-\abs{\setcomparesmaller})$. Denote $\quantityinv^*_{\idxitem\greateritem\idxitem'}$ as the topological ordering with item $\idxitem$ in position $(\abs{\setcomparelarger} + 1)$. Denote $\quantityinv^*_{\idxitem\lessitem\idxitem'}$ as the topological ordering with item $\idxitem'$ in position $(\abs{\setcomparelarger} + 1)$. For any possible $\quantityinvestre$, one of the conditions in~\eqref{eq:kt_sf_condition_strict_inequality} holds on the pair $\{\quantityinv^*_{\idxitem\greateritem\idxitem'},  \quantityinv^*_{\idxitem\lessitem\idxitem'}\}$, and hence strict inequality in~\eqref{eq:kt_sf_equal} holds for the pair $\{\quantityinv^*_{\idxitem\greateritem\idxitem'},  \quantityinv^*_{\idxitem\lessitem\idxitem'}\}$.

Eq.~\eqref{eq:kt_sf_strict_nonzero_probability} implies that the event $\quantityinvgt\in \{\quantityinv^*_{\idxitem\greateritem\idxitem'},  \quantityinv^*_{\idxitem\lessitem\idxitem'}\}$ arises with non-zero probability. Taking an expectation over all possible pairs $\{\quantityinv_{\idxitem\greateritem\idxitem'},  \quantityinv_{\idxitem\lessitem\idxitem'}\}$ in~\eqref{eq:kt_sf_equal}, and using the strict inequality for the pair $\{\quantityinv^*_{\idxitem\greateritem\idxitem'},  \quantityinv^*_{\idxitem\lessitem\idxitem'}\}$ yields
\begin{align*}
    & \Expect[\losssf(\quantityinvcardourssortkt, \quantityinvgt) \given \infoordall = \infoordallobs, \eventexist] < \Expect[\losssf(\quantityinvestre, \quantityinvgt) \given \infoordall = \infoordallobs, \eventexist].
\end{align*}

Taking an expectation over the set of ordinal observations $\infoordall$ yields
\begin{align}
        & \Expect[\losssf(\quantityinvcardourssortkt, \quantityinvgt) \given \eventexist] < \Expect[\losssf(\quantityinvestre, \quantityinvgt) \given \eventexist].\label{eq:kt_sf_conditioned_twin}
\end{align}

Combining~\eqref{eq:kt_sf_conditioned_twin} with inequality~\eqref{eq:kt_sf_conditioned_no_twin} for the re-arranging step yields
\begin{align}
    \Expect[\losssf(\quantityinvcardourssortkt, \quantityinvgt)] < \Expect[\losssf(\quantityinvestre, \quantityinvgt)].\label{eq:kt_sf_step_canonical}
\end{align}

Finally, combining~\eqref{eq:kt_sf_step_canonical} with inequality~\eqref{eq:kt_sf_step_rearranging} for the re-arranging step completes the proof for Spearman's footrule.\\

We make a comment about having multiple \topoidentical pairs. Notice that in Algorithm~\ref{alg:sort_kendall_tau}, we only find one \topoidentical pair, and then break out of the for-loops. Alternatively, we can identify and flip multiple disjoint \topoidentical pairs in a similar fashion as in Algorithm~\ref{alg:sort_cardinal}. This is still a valid algorithm, because each step of processing one \topoidentical pair does not increase Kendall-tau distance or Spearman's footrule distance.

It remains to prove Lemma~\ref{lemma_rearrange}.

\subsection{Proof of Lemma~\ref{lemma_rearrange}}
Consider any two items $\idxitemcontrib, \idxitemcontrib'\in [\numitems]$, such that $\idxitemcontrib\greateritem\idxitemcontrib'$ in the true ranking $\quantityinvgt$. Let $\quantityinvest_1$ be an arbitrary ranking. Let $\quantityinvest_2$ be a ranking where all items are ranked the same as in $\quantityinvest_1$, except that the positions of items $\idxitemcontrib$ and $\idxitemcontrib'$ are flipped as compared to $\quantityinvest_1$. The remainder of the proof is broken into two parts.\\

\noindent \emph{Part 1: If the relative ordering of a pair is inconsistent with the relative ordering indicated by the true ranking, then flipping this pair does not increase Kendall-tau distance or Spearman's footrule distance.} 

Specifically, we claim that if $\idxitemcontrib \lessitem\idxitemcontrib'$ in $\quantityest_1$, then $\quantityinvest_2$ has a smaller or equal loss than $\quantityinvest_1$, with respect to Kendall-tau distance and Spearman's footrule distance. We discuss the two distance metrics separately.\\

\noindent\textbf{Kendall-tau distance: }
First, consider Kendall-tau distance contributed by the pair $(\idxitemcontrib, \idxitemcontrib')$. We have $\idxitemcontrib\lessitem \idxitemcontrib'$ in $\quantityinvest_1$ and $\idxitemcontrib \greateritem\idxitemcontrib'$ in $\quantityinvest_2$. Since we have $\idxitemcontrib\greateritem\idxitemcontrib'$ in the true ranking, the relative ordering of this pair is correct in $\quantityinvest_2$, and incorrect in $\quantityinvest_1$. Hence, 
\begin{align}
    0 = \ktcontribution(\quantityinvest_2, \quantityinvgt, \idxitemcontrib, \idxitemcontrib') < \ktcontribution(\quantityinvest_1, \quantityinvgt, \idxitemcontrib, \idxitemcontrib') = 1.\label{eq:kt_lem_kt_one}
\end{align}

Denote $\idxitemcontribmid$ as any item ranked in between $\idxitemcontrib$ and $\idxitemcontrib'$ in $\quantityinvest_1$ (or equivalently, in $\quantityinvest_2$). In the rest of the pairs that are not $(\idxitemcontrib, \idxitemcontrib')$, the flip only changes the relative ordering of each pair of the form  $(\idxitemcontrib, \idxitemcontribmid)$ or $(\idxitemcontrib', \idxitemcontribmid)$. If in the true ranking $\quantityinvgt$, item $\idxitemcontribmid$ is ranked higher than both $(\idxitemcontrib, \idxitemcontrib')$, or ranked lower than both $(\idxitemcontrib, \idxitemcontrib')$, then the sum of the contributions to Kendall-tau distance by the pair $(\idxitemcontrib, \idxitemcontribmid)$ and the pair $(\idxitemcontrib', \idxitemcontribmid)$ is the same in $\quantityinvest_1$ and $\quantityinvest_2$:
\begin{align}
    \ktcontribution(\quantityinvest_2, \quantityinvgt, \idxitemcontrib, \idxitemcontribmid) +  \ktcontribution(\quantityinvest_2, \quantityinvgt, \idxitemcontrib', \idxitemcontribmid) = 1 = \ktcontribution(\quantityinvest_1, \quantityinvgt, \idxitemcontrib, \idxitemcontribmid) +  \ktcontribution(\quantityinvest_1, \quantityinvgt, \idxitemcontrib', \idxitemcontribmid).\label{eq:kt_lem_kt_two}
\end{align}

Otherwise $\idxitemcontribmid$ is ranked in between $\idxitemcontrib$ and $\idxitemcontrib'$ in the true ranking $\quantityinvgt$, then we have 
\begin{align}
    0 = \ktcontribution(\quantityinvest_2, \quantityinvgt, \idxitemcontrib, \idxitemcontribmid) +  \ktcontribution(\quantityinvest_2, \quantityinvgt, \idxitemcontrib', \idxitemcontribmid) < \ktcontribution(\quantityinvest_1, \quantityinvgt, \idxitemcontrib, \idxitemcontribmid) +  \ktcontribution(\quantityinvest_1, \quantityinvgt, \idxitemcontrib', \idxitemcontribmid) = 2.\label{eq:kt_lem_kt_three}
\end{align}

Combining the expression~\eqref{eq:kt_rewrite} of Kendall-tau distance with~\eqref{eq:kt_lem_kt_one},~\eqref{eq:kt_lem_kt_two} and~\eqref{eq:kt_lem_kt_three} yields
\begin{align*}
    \losskt(\quantityinvest_2, \quantityinvgt) < \losskt(\quantityinvest_1, \quantityinvgt).
\end{align*}

\noindent\textbf{Spearman's footrule distance: } By flipping the positions of the items $(\idxitemcontrib, \idxitemcontrib')$, only Spearman's footrule distance contributed by these two items has changed. Recall that the condition for flipping the pair $(\idxitemcontrib, \idxitemcontrib')$ requires $\idxitemcontrib \lessitem \idxitemcontrib'$ in $\quantityest_1$ and $\idxitemcontrib\greateritem \idxitemcontrib'$ in $\quantityinvgt$. Applying the rearrangement inequality~\eqref{eq:rearrangement} with $\tmpa_1 = \quantityinvest_1(\idxitemcontrib'), \tmpa_2 = \quantityinvest_1(\idxitemcontrib), \tmpb_1 = \quantityinvgt(\idxitemcontrib), \tmpb_2 = \quantityinvgt(\idxitemcontrib')$, we have
\begin{align}
    \abs{\quantityinvest_1(\idxitemcontrib') - \quantityinvgt(\idxitemcontrib') } + \abs{\quantityinvest_1(\idxitemcontrib) - \quantityinvgt(\idxitemcontrib)} \ge & \abs{\quantityinvest_1(\idxitemcontrib') - \quantityinvgt(\idxitemcontrib) } + \abs{\quantityinvest_1(\idxitemcontrib) - \quantityinvgt(\idxitemcontrib')}\nonumber\\
    = & \abs{\quantityinvest_2(\idxitemcontrib) - \quantityinvgt(\idxitemcontrib) } + \abs{\quantityinvest_2(\idxitemcontrib') - \quantityinvgt(\idxitemcontrib')}.\label{eq:kt_lem_sf}
\end{align}

Combining~\eqref{eq:kt_lem_sf} with the definition of Spearman's footrule distance yields
\begin{align*}
    \losssf(\quantityinvest_2, \quantityinvgt) \le \losssf(\quantityinvest_1, \quantityinvgt).
\end{align*}

This completes Part 1 of the proof.\\

\noindent \emph{Part 2: The re-arranging step in Algorithm~\ref{alg:sort_kendall_tau} is equivalent to a sequence of pair flips.} 

With Part 1 in place, we now explain the rest of the proof. For any arbitrary \topoidentical pair of items $(\idxitem, \idxitem')$ and any arbitrary set of ordinal observations $\infoordall$, denote the sets $\setcompare_1 \defn \setcomparelarger(\idxitem, \idxitem', \infoordall), \setcompare_2 \defn \{\idxitem, \idxitem'\}, \setcompare_3 \defn \setcomparesmaller(\idxitem, \idxitem', \infoordall)$. We consider the following procedure. We start by setting $\quantityinvest_1$  as the initial estimated ranking $\quantityinvinit$. We identify one pair $(\idxitemcontrib, \idxitemcontrib')$ (if any) such that the following three conditions are met. First, we have $\idxitemcontrib\in \setcompare_\idxgrader, \idxitemcontrib' \in \setcompare_{\idxgrader'}$ with $\idxgrader < \idxgrader'$. Second, we have $\idxitemcontrib\lessitem\idxitemcontrib'$ in $\quantityinvest_1$. Third, there is no item in $\setcompare_1 \union \setcompare_2 \union \setcompare_3$, whose position is in between $\idxitemcontrib$ and $\idxitemcontrib'$ in the ranking $\quantityinvest_1$. If such a pair is found, we flip the positions of $\idxitemcontrib$ and $\idxitemcontrib'$, and update $\quantityinvest_1$ to be this new ranking. Repeat this procedure until no such pair can be found.

Now we show that this procedure is equivalent to the re-arranging step in Line~\ref{line:kt_rearrange_start}-\ref{line:kt_rearrange_end} of Algorithm~\ref{alg:sort_kendall_tau}. This procedure properly terminates, because each pair of items $(\idxitemcontrib, \idxitemcontrib')$ can be swapped at most once, and there is a finite number of pairs. When the procedure terminates, the ranking is identical to the re-arranged ranking $\quantityinvest$ after Line~\ref{line:kt_rearrange_end} of Algorithm~\ref{alg:sort_kendall_tau}. To see this claim, we first note that this procedure has never moved items outside $\setcompare_1 \union \setcompare_2 \union \setcompare_3$, so we only need to concern about the items in $\setcompare_1 \union \setcompare_2 \union \setcompare_3$ and their positions. For each pair $(\idxitemcontrib, \idxitemcontrib')$ to be flipped, the procedure specifies that $\idxitemcontrib$ and $\idxitemcontrib'$ belong to two different sets from $\setcompare_1, \setcompare_2$ and $\setcompare_3$. Moreover, by the condition on the pair $(\idxitemcontrib, \idxitemcontrib')$, there cannot be any item in $\setcompare_1 \union \setcompare_2 \union \setcompare_3$ that is ranked in between $\idxitemcontrib$ and $\idxitemcontrib'$. Hence, the relative ordering of the items within each set of $\setcompare_1, \setcompare_2$ or $\setcompare_3$ is unchanged, consistent with the ranking specified in Line~\ref{line:kt_rearrange_placement_one} and Line~\ref{line:kt_rearrange_placement_two} of Algorithm~\ref{alg:sort_kendall_tau}. Moreover, the re-arranging step in Algorithm~\ref{alg:sort_kendall_tau} ranks all items in $\setcompare_1$ before all items in $\setcompare_2$, and all items in $\setcompare_2$ before all items in $\setcompare_3$. Assume that the final output of the procedure is a different ranking from the re-arranging step in Algorithm~\ref{alg:sort_kendall_tau}, then we can find a pair $(\idxitemcontrib, \idxitemcontrib')$ that can be flipped, contradicting the fact that no such pairs can be found at the termination of the procedure. Hence, the procedure and the re-arranging step in Algorithm~\ref{alg:sort_kendall_tau} are equivalent. Applying Part 1 to each flip in this procedure completes the proof of the lemma.

\section{Ranking under arbitrary true ranking}\label{app:ranking_arbit}

Theorem~\ref{thm:sort_ours_uniform_prior} in Section~\ref{sec:ranking} compared our cardinal estimator with arbitrary ordinal estimators under a uniform prior over the true ranking. In this section, we present a result for ranking under any arbitrary true ranking. This setting is more similar to our results on the canonical setting (Theorem~\ref{thm:canonical_ours}) and A/B testing (Theorem~\ref{thm:abtest_ours}) in the main text. When the true ranking is arbitrary, a minimax-optimal ordinal estimator outputs uniformly at random a topoglocial ordering consistent with the pairwise comparisons. We denote this optimal ordinal estimator as $\quantityordsortunif$.

Given this ordinal estimator, we then construct a cardinal estimator $\quantitycardourssortunif$ by simply setting the initial estimate $\quantityest = \quantityordsortunif(\infoordall)$ in Line 2 of Algorithm~\ref{alg:sort_cardinal} (instead of executing the current Line 2). The following theorem states the desired result for strict uniform dominance of this cardinal estimator over the optimal ordinal estimator $\quantityordsortunif$.

\begin{theorem}\label{cor:sort_ours}
When the true ranking is arbitrary, the cardinal estimator $\quantitycardourssortunif$ strictly uniformly dominates the minimax-optimal ordinal estimator $\quantityordsortunif$.
\end{theorem}

Importantly, we can think of this cardinal estimator as a post-processing step which builds on the output of the optimal ordinal estimator. This cardinal estimator takes polynomial time in the number of items $\numitems$, in addition to the time taken by one call to the ordinal estimator $\quantityordsortunif$.

We prove Theorem~\ref{cor:sort_ours} in the remainder of this section.

\subsection{Proof of Theorem~\ref{cor:sort_ours}}
The proof is a slight modification to the proof of Theorem~\ref{thm:sort_ours_uniform_prior}, so we only highlight the difference. First, we consider the probability of success of the optimal ordinal estimator $\quantityordsortunif$ that outputs one of the topological orderings uniformly at random:

\begin{align}
    \Prob(\quantityordsortunif(\infoordallobs) = \quantitygt \given \infoordall = \infoordallobs) = & \sum_{\quantity\in \topo(\infoordallobs)} \Prob(\quantity = \quantitygt \given \quantityordsortunif = \quantity, \infoordall = \infoordallobs)\Prob(\quantityordsortunif = \quantity \given  \infoordall = \infoordallobs)\nonumber\\
    \stackrel{\stepone}{=} &  \frac{1}{\numorders(\infoordallobs)}\sum_{\quantity\in \topo(\infoordallobs)} \Prob(\quantity = \quantitygt \given \quantityordsortunif = \quantity, \infoordall = \infoordallobs),\label{eq:cor_arbitrary_ordinal}
\end{align}
where equality \stepone is true because the ordinal estimator $\quantityordsortunif$ uniformly at random outputs one of the topological orderings consistent with $\infoordallobs$. 

Now we consider each term $\Prob(\quantity = \quantitygt \given \quantityordsortunif = \quantity, \infoordall = \infoordallobs )$ in~\eqref{eq:cor_arbitrary_ordinal}. The quantities $\quantitygt$ and $\quantity$ are both deterministic. Trivially, we have
\begin{align}
    \Prob(\quantity = \quantitygt \given \quantityordsort = \quantity, \infoordall = \infoordallobs)= 
    \begin{cases}
        1 & \text{if } \quantity = \quantitygt\\
        0 & \text{otherwise}.
    \end{cases}\label{eq:cor_arbitrary_ordinal_term}
\end{align}

Combining~\eqref{eq:cor_arbitrary_ordinal} and~\eqref{eq:cor_arbitrary_ordinal_term} with the fact that the true ranking $\quantitygt$ must be a topological ordering consistent with $\infoordallobs$, we have
\begin{align}
    \Prob(\quantityordsortunif(\infoordallobs) = \quantitygt \given \infoordall = \infoordallobs) = & \frac{1}{\numorders(\infoordallobs)}.\label{eq:ranking_ordinal_arbitrary}
\end{align}

Now consider the cardinal estimator $\quantitycardourssortunif$. When the number of flippable pairs is zero, the cardinal estimator behaves equivalently as the ordinal estimator $\quantityordsortunif$. Following a similar argument as Case 1 in the proof of Theorem~\ref{thm:sort_ours_uniform_prior}, for any set of ordinal observations $\infoordallobs$, we have (cf. Equation~\eqref{eq:ranking_cardinal_flippable_pair_zero} in the proof of Theorem~\ref{thm:sort_ours_uniform_prior}):
\begin{align}
    \Prob(\quantitycardourssortunif \given \infoordall = \infoordallobs, \numflips = 0) = \Prob(\quantityordsortunif = \quantitygt\given \infoordall = \infoordallobs).\label{eq:ranking_cardinal_arbitrary_flippable_pair_zero}
\end{align}

Denote $\quantityinit$ as the initial estimated ranking obtained by calling the ordinal estimator $\quantityordsortunif$. When the number of flippable pairs is $\numflips = \numflipsobs > 0$, the probability of success of the cardinal estimator is
\begin{align}
    \Prob(\quantitycardourssortunif  = \quantitygt & \given \infoordall = \infoord, \numflips = \numflipsobs) \nonumber \\
    = & \sum_{\quantity\in \topo(\infoordallobs)}\Prob(\quantitycardourssortunif = \quantitygt \given \infoordall = \infoordallobs, \numflips = \numflipsobs, \quantityinit = \quantity)\Prob(\quantityinit = \quantity \given \infoordall = \infoordallobs, \numflips = \numflipsobs)\nonumber\\
    \stackrel{\stepone}{=} & \frac{1}{\numorders(\infoordallobs)}\sum_{\quantity\in \topo(\infoordallobs)}\Prob(\quantitycardourssortunif = \quantitygt \given \infoordall = \infoordallobs, \numflips = \numflipsobs, \quantityinit = \quantity),\label{eq:ranking_cardinal_arbitrary_flippable_pair_at_least_one_intermediate}
\end{align}
where equality \stepone is true because the ordinal estimator $\quantityordsortunif$ outputs a topological ordering uniformly at random.

The remaining argument is similar to Case 2 in the proof of Theorem~\ref{thm:sort_ours_uniform_prior}, so we only outline the main steps. Consider all total rankings that are identical to the true ranking $\quantitygt$, except for (possibly) the relative ordering of the $\numflipsobs$ flippable pairs. There are $2^\numflipsobs$ such total rankings, and all these $2^\numflipsobs$ total rankings are topological orderings on the graph $\graph(\infoordall)$. In~\eqref{eq:ranking_cardinal_arbitrary_flippable_pair_at_least_one_intermediate}, the summation of $\quantity$ is over all topological orderings. In particular, this summation includes these $2^\numflipsobs$ total rankings. Recall that the cardinal estimator $\quantitycardourssortunif$ is obtained by replacing Line~\ref{line_alt:initial_guess} of Algorithm~\ref{alg:sort_cardinal} by calling the ordinal estimator $\quantityordsortunif$. To be able to apply Theorem~\ref{thm:canonical_ours}, we obtain a cardinal estimator $\quantitycardourssortunifequiv$ by replacing Line~\ref{line_alt:initial_guess} of Algorithm~\ref{alg:sort_cardinal_alternative} by calling the ordinal estimator $\quantityordsortunif$. This estimator $\quantitycardourssortunifequiv$ is equivalent to the original estimator $\quantitycardourssortunif$. When the initial estimated ranking $\quantityinit$ is any of the $2^\numflipsobs$ total rankings, the probability that the cardinal estimator $\quantitycardourssortunifequiv$ gives the correct output is strictly greater than $\frac{1}{2^\numflipsobs}$. Hence, we bound~\eqref{eq:ranking_cardinal_arbitrary_flippable_pair_at_least_one_intermediate} as (cf. Equation~\eqref{eq:ranking_cardinal_flippable_pair_at_least_one} in the proof of Theorem~\ref{thm:sort_ours_uniform_prior}):
\begin{align}
    \Prob(\quantitycardourssortunifequiv = \quantitygt \given \infoordall = \infoord, \numflips = \numflipsobs) > \frac{1}{\numorders(\infoordallobs)}\cdot 2^\numflipsobs \cdot \frac{1}{2^\numflipsobs} = \frac{1}{\numorders(\infoordallobs)} \stackrel{\stepone}{=} \Prob(\quantityordsortunif = \quantitygt \given \infoordall = \infoordallobs).\label{eq:ranking_cardinal_arbitrary_flippable_pair_at_least_one}
\end{align}
where equality \stepone is true from~\eqref{eq:ranking_ordinal_arbitrary}.

Having established~\eqref{eq:ranking_cardinal_arbitrary_flippable_pair_zero} and~\eqref{eq:ranking_cardinal_arbitrary_flippable_pair_at_least_one}, the rest of the argument follows the proof of Theorem~\ref{thm:sort_ours_uniform_prior}.

\end{document}